\documentclass[11pt]{article}

\usepackage[final]{acl}

\usepackage{times}
\usepackage{latexsym}

\usepackage{pdflscape}
\usepackage{subcaption}

\usepackage{fontawesome}

\usepackage[T1]{fontenc}

\usepackage[utf8]{inputenc}

\usepackage{microtype}

\usepackage{inconsolata}

\usepackage{graphicx}
\usepackage{amsmath}
\usepackage{amssymb}
\usepackage{mathtools}
\usepackage{amsthm}

\usepackage{fontawesome}

\usepackage{multirow}
\usepackage{array}
\usepackage{booktabs}

\usepackage[disable,textsize=tiny]{todonotes}
\renewcommand{\textcolor}[2]{}

\title{No Single Tokenizer Feature Reliably Predicts \\ Downstream Language Model Performance}

\author{
 \textbf{Antonio Lopardo\textsuperscript{1,5}},
 \textbf{Luis Frentzen Salim\textsuperscript{1,2,3}},
 \textbf{Shikhar Srivastava\textsuperscript{1,4}}, \\
 \textbf{Kevin Jiang\textsuperscript{1}}, 
 \textbf{Mridul Sharma\textsuperscript{1}}, 
 \textbf{Catherine Arnett\textsuperscript{1}}
\\
 \textsuperscript{1}EleutherAI,
 \textsuperscript{2}National Taiwan University of Science and Technology,
\\
 \textsuperscript{3}Academia Sinica,
 \textsuperscript{4}{University of Rochester},
 \textsuperscript{5}University of Basel
\\
 \small{
   \textbf{Correspondence:} \href{mailto:catherine@eleuther.ai}{catherine@eleuther.ai}
 }
}

\begin{document}

\theoremstyle{plain}
\newtheorem{theorem}{Theorem}[section]
\newtheorem{proposition}[theorem]{Proposition}
\newtheorem{lemma}[theorem]{Lemma}
\newtheorem{corollary}[theorem]{Corollary}
\theoremstyle{definition}
\newtheorem{definition}[theorem]{Definition}
\newtheorem{assumption}[theorem]{Assumption}
\theoremstyle{remark}
\newtheorem{remark}[theorem]{Remark}

\maketitle
\begin{abstract}
Tokenization is a critical component of the language modeling pipeline, but it is poorly understood. Tokenizers can impact downstream performance, but there is currently mixed evidence about which tokenizer features predict a better language model without requiring a full training run. In this paper, we analyze 159 language models in over 200 model-language evaluations across nine languages. We evaluate both models and their tokenizers and test whether there is a relationship between any proposed intrinsic tokenizer evaluations and model performance, as measured by a wide variety of benchmarks. Despite the quantity and variety of proposed intrinsic tokenizer metrics, we find that none consistently predicts downstream performance across model families, tasks, and languages. This motivates the need for novel tokenizer evaluation metrics and approaches. A new metric introduced in this work, based on unsupervised word-boundary statistics, correlates more strongly with English and Turkish performance than any existing metric in one model family, though we find an effect in the opposite direction than predicted.

\end{abstract}

\begin{center}
       \raisebox{-2.2pt}{\includegraphics[scale=0.09]{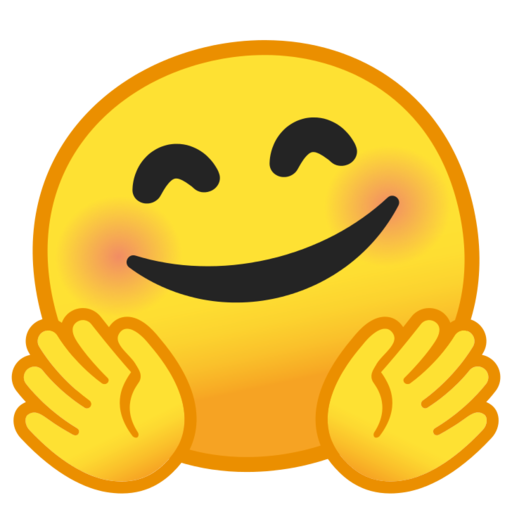}} \href{https://huggingface.co/collections/shikhar-srivastava/vista-130m}{~\texttt{VISTA 130M}} ~~ \href{https://huggingface.co/collections/shikhar-srivastava/vista-350m}{~\texttt{VISTA 350M}}~~~  \\ 
       \faGithub~ \href{https://github.com/AntonioLopardo/tokenizer-metric-multilingual-correlations}{\texttt{Code and Data}}

\end{center}

\section{Introduction} \label{sec:Introduction}

Tokenization is one of the earliest steps in the language modeling pipeline and it affects many aspects of the language model, including performance. 
Several studies have shown performance differences between models trained otherwise identically, varying only the tokenizer \citep{ali-etal-2024-tokenizer,goldman-etal-2024-unpacking,schmidt-etal-2024-tokenization,lotz2025beyond}. However, it is not currently well understood which properties of a tokenizer generally lead to better performance. 
This prevents language model developers from making principled design decisions for selecting tokenization algorithm, pretokenizer, vocabulary size, etc.

Reliable \textit{intrinsic tokenizer evaluations}, i.e. metrics derived directly from the tokenizer, would be valuable if they allow researchers to identify more optimal tokenizers before training any language models. Many intrinsic metrics have been proposed, ranging from statistical (e.g. Zipfianness; \citealp{lotz2025beyond}), compression (e.g. fertility; \citealp[]{rust-etal-2021-good}), information theoretic (e.g. Rényi entropy; \citealp[]{zouhar-etal-2023-tokenization}), and linguistic (e.g. MorphScore; \citealp{arnett-bergen-2025-language,arnett2025morphscore}) metrics. However, some of these have later been shown to correlate poorly with downstream performance \citep{schmidt-etal-2024-tokenization, cognetta-etal-2024-two, arnett-bergen-2025-language, arnett2025morphscore}.

In the current work, we conduct a comprehensive analysis of proposed intrinsic tokenizer metrics across nine languages and a wide variety of tasks. We use three model suites: two existing suites and a new suite of monolingual models we trained from scratch, which differ only in their tokenizers. We address three related high-level questions:

\paragraph{RQ1: Are there any intrinsic tokenizer metrics that reliably predict downstream model performance?} Given the mixed evidence about which tokenizer metrics correlate with better downstream performance, we calculate the correlation between most proposed tokenizer metrics and downstream language model performance for three different model suites. Many of these tokenizer metrics have been proposed based on empirical evidence from a single language or a small number of languages. We ask whether any of these metrics reliably correlate with downstream performance across tasks, languages, and model suites (Section \ref{sec:individual_metric_corr}). We also propose and test a novel intrinsic tokenizer metric, which we adapt from prior work on entropy-guided segmentation (Section \ref{sec:ustat_deep_dive}).

\paragraph{RQ2: What is the relationship between different tokenizer metrics?} 
Several proposed intrinsic tokenizer metrics are in apparent conflict with one another. For example, MorphScore measures the extent to which a tokenizer preserves meaningful linguistic boundaries in a word. One way tokenizers can achieve higher MorphScore is by segmenting words into more tokens, so that token boundaries are more likely to overlap with linguistic boundaries. This would affect compression, and it is thought segmenting a word into more tokens (i.e. having worse compression) is associated with \textit{worse performance}. We ask: are there tradeoffs between features like MorphScore and compression that an optimal tokenizer should balance? 
In Section \ref{sec:multiple_metric_corr}, we characterize the relationships between different intrinsic evaluations. We can then ask whether identifying these tradeoffs helps us understand which combinations of metrics might be more useful for predicting downstream performance. We then test whether any combination of intrinsic metrics may more reliably predict downstream performance. 

\paragraph{RQ3: Do orthographic tasks correlate with better performance?}
It has been asserted that there are certain undesirable language model behaviors that are caused by tokenization. One of these is the lack of orthographic awareness, including the so-called ``strawberry problem'' \citep{silberling2024why,cosma-etal-2025-strawberry}.
This refers to the fact that even highly performant language models may not have orthographic awareness of their tokens and, as a result, cannot count the number of \textit{r}s in `strawberry.' Partially in response to this observed behavior, there have been a number of proposed orthographic knowledge and string manipulation tasks, e.g. LMEntry \citep{efrat2023lmentry} and CUTE \citep{edman-etal-2024-cute}, which aim to measure how much awareness language models have of the characters in their tokens and their ability to manipulate strings based on that awareness. However, there is minimal theoretical motivation behind these tasks. Do these tasks measure desirable properties of a tokenizer? To address this question, we ask whether performance on orthographic tasks correlates with better performance on other tasks.

\smallskip

Our main findings are:
\begin{itemize}
    \item \textbf{RQ1:} No intrinsic tokenizer metric reliably predicts downstream performance across model suites, languages, and tasks. While idiosyncratic correlations exist, they do not generalize.
    \item \textbf{RQ2:} Intrinsic metrics are highly redundant, metrics within the same category are strongly correlated, and combining metrics does not improve prediction.
    \item \textbf{RQ3:} Orthographic tasks are only weakly correlated with general downstream performance, raising questions about their role in guiding tokenizer design.
\end{itemize}
Tokenizers are a critical aspect of language model design \citep{alqahtani2026stop}, but it is still not known how to measure whether a tokenizer is suitable for a particular model. Our results motivate new approaches to tokenizer design and evaluation.

\section{Related Work} \label{sec:related_work}

Most researchers evaluate tokenizers through intrinsic evaluation: evaluating the tokenizer itself, independent of any model.
Reliable intrinsic evaluation would be valuable because experiments involving training models while varying only the tokenizer can be prohibitively expensive. Here we introduce the most widely used intrinsic tokenizer metrics that have been proposed.

\paragraph{Compression metrics.} 
One of the most commonly studied dimensions of tokenizer quality is compression. 
Compression is measured in a variety of ways, most often fertility (tokens per word; \citealp{rust-etal-2021-good}) or bytes per token \citep{liu2025superbpe}. 
Higher compression on the test set is associated with better performance in machine translation \citep{galle-2019-investigating}.
\citet{deletang2024language} show that minimizing a language model's cross-entropy loss is equivalent to minimizing the number of bits required to compress the data, establishing a direct connection between compression and modeling quality. 
But in controlled studies where comparable models are trained with different tokenizers, results generally have not supported this link between compression and performance \citep{ali-etal-2024-tokenizer,schmidt-etal-2024-tokenization}.

\paragraph{Information-theoretic metrics.} 
Information-theoretic measures have been proposed to characterize the distributional properties of tokenizer vocabularies. R\'enyi entropy \citep{zouhar-etal-2023-tokenization} generalizes Shannon entropy and penalizes unbalanced token distributions, particularly those dominated by a few high-frequency tokens. 
Other related metrics include unigram and bigram entropy, which measure the diversity of token usage and contextual variety where, similarly, higher entropy may be more desirable. However, \citet{cognetta-etal-2024-two} showed that while R\'enyi entropy captures some desirable properties of a tokenizer, it is not a holistic measure of tokenizer quality. Furthermore, \citet{dagan2024getting} find that higher R\'enyi entropy in fact is correlated with \textit{lower} performance in code generation. Beyond vocabulary characterization, entropy at character boundaries can identify meaningful linguistic units, a principle used by the Byte Latent Transformer to dynamically segment byte sequences \citep{pagnoni-etal-2025-byte}, and may help preserve multi-word expressions, an important aspect of tokenizer quality (\citealp[][\textit{inter alia}]{gow-smith-etal-2022-improving}).

\paragraph{Linguistic metrics.} 
Tokenizers that preserve the internal structure of words and segment along morphological boundaries are argued to lead to better performance \citep{klein-tsarfaty-2020-getting,park-etal-2020-empirical,vasiu2020romanian,hofmann-etal-2020-dagobert,nzeyimana-niyongabo-rubungo-2022-kinyabert,libovicky-helcl-2024-lexically,bauwens-delobelle-2024-bpe}.
For example, \citet{bostrom-durrett-2020-byte} proposed evaluating tokenizers based on their alignment with morphological structure. Subsequent work has further developed morphology-based evaluation criteria, showing, for instance, that derivational segmentation improves model interpretations of complex words \citep{hofmann-etal-2021-superbizarre}.
But, there are also a number of studies which have failed to find benefits of morphologically aligned tokenization \citep{machavcek2018morphological,saleva-lignos-2021-effectiveness,choo2023study,toraman2023turkish,arnett-etal-2024-different,arnett-bergen-2025-language, arnett2025morphscore}.

\paragraph{Statistical metrics.} 
Statistical approaches evaluate how effectively tokenizers use their vocabularies and how token usage is distributed.
Measures like vocabulary utilization (the proportion of vocabulary items actually used when encoding a corpus) and type-token ratio (the ratio of unique tokens to total tokens) capture basic aspects of vocabulary coverage and lexical diversity.
Other metrics capture how token usage is distributed across the vocabulary: average (token) rank is the expected rank of a token sampled from a corpus when vocabulary items are sorted by frequency. 
\citet{limisiewicz-etal-2023-tokenization} show that vocabulary allocation, which they measure via average token rank and characters per token, has nuanced effects on multilingual performance: higher allocation benefits word-level tasks but harms masked language modeling scores.
Also examining vocabulary frequency distributions, \citet{lotz2025beyond} propose metrics based on Zipf's law, which assess whether token frequencies follow the power-law patterns observed in natural language.
In these studies, these metrics correlate more strongly with downstream performance than compression-based metrics, particularly in multilingual settings.

\section{U-Statistic} \label{sec:ustat_deep_dive}

Here we introduce a novel intrinsic tokenizer metric.
We adopt the U-statistic of \citet{hu2025entropydriven}, a per-token score that formalizes ideas from unsupervised word segmentation \citep{harris1955phoneme,jin-tanaka-ishii-2006-unsupervised}, and aggregate it over a tokenizer's vocabulary as an intrinsic evaluation asking: \textit{does this token correspond to a natural linguistic unit?} The score, for each token $w$, combines two signals from corpus character statistics:
\begin{multline} \label{eq:ustat}
U_{\text{stat}}(w) = \underbrace{\text{PMI}_{\min}(w)}_{\text{internal cohesion}} \\
+ \; \lambda \cdot \underbrace{\min\bigl(H_{\text{left}}(w),\, H_{\text{right}}(w)\bigr)}_{\text{contextual separability}}
\end{multline}

The first term takes the minimum pointwise mutual information (PMI) over adjacent character pairs inside the token, so a single weak pair flags the token as crossing a natural boundary. The second takes the minimum branching entropy at the token's edges; high entropy signals a natural break point \citep{harris1955phoneme}. We report mean (\texttt{u\_mean}), median (\texttt{u\_median}), frequency-weighted mean (\texttt{u\_mean\_weighted}), and log-frequency-weighted mean (\texttt{u\_mean\_log\_weighted}) of $U_{\text{stat}}$; full definitions and illustrative examples are in Appendix~\ref{app:tokenizer_metrics}.

\section{Materials}

\subsection{Models}

We use three suites of models: the models from TokSuite \citep{toksuite2025}; the models in the paper ``Tokenization Is More Than Compression'' (\citealp{schmidt-etal-2024-tokenization}; henceforth TIMTC); and a suite of models we train ourselves, VISTA.

\paragraph{TokSuite models.}

The TokSuite models \citep{toksuite2025} consist of 14 1B-parameter Llama architecture models with identical training budget and initialization, which differ only in their tokenizers. These models enable us to attribute performance differences to tokenizer characteristics. 
The tokenizers are drawn from existing models and feature various vocabulary construction algorithms (BPE, Unigram, WordPiece), as well as both byte-level and subword tokenizers. 
Among these models, vocabulary sizes vary substantially, from a byte-level vocabulary of 259 tokens to approximately 250k (most are between 30k--130k).  

The models were trained on a multilingual corpus  consisting of 40B tokens of English data from FineWeb-Edu  \citep{penedo2024fineweb} and 15B tokens each of Turkish, Chinese, Italian, and Farsi data from FineWeb-2 \citep{penedo2025fineweb2}. 

\paragraph{TIMTC models.}
In \citet{schmidt-etal-2024-tokenization}, the authors investigate the hypothesis that the effectiveness of BPE arises primarily from its text compression capabilities. 
They release a suite of 54\footnote{We used 45 of the available models due to persistent out-of-vocabulary errors that we could not resolve.} 350M-parameter models that differ only in their tokenizers, and which vary systematically in their:  
(1) \emph{vocabulary construction}, how the vocabulary is built; 
(2) \emph{initial vocabulary}, the token set used to seed vocabulary construction; 
(3) \emph{pre-tokenization}, rules that restrict or enforce token boundaries; and
(4) \emph{segmentation}, the algorithm that assigns segmented text into tokens given a vocabulary 
(see Appendix~\ref{app:timtc_details} for details). The models are trained only on English data.

This carefully controlled suite—where models differ only in their tokenization choices while sharing identical architecture, training data, and hyperparameters—allows us to isolate the effects of tokenization on downstream performance.

\paragraph{VISTA Models.} \label{sec:our_models}
We introduce VISTA (\underline{V}ocabulary-size \underline{I}nvestigation \underline{S}uite for \underline{T}okenizer \underline{A}nalysis), a suite of monolingual models for five languages (English, Thai, Urdu, Amharic, and Vietnamese), which systematically varies vocabulary size using both BPE and Unigram tokenizers. 
For each language and vocabulary size, we pre-train a language model based on the LLAMA 1 architecture~\cite{touvron2023llama1}.
The language models are pre-trained using a standard auto-regressive cross-entropy loss. 
We train monolingual models with corresponding language-specific tokenizers, with varying vocabulary sizes. 
Varying vocabulary size is one way to systematically manipulate tokenizer characteristics like compression \citep{arnett2025crosslingual} and has been shown to impact model performance \citep{toraman2023turkish}. 

We train two model sizes: 130M and 350M parameters. 
We detail the hyperparameters and training design decisions in Appendix~\ref{app:vista_models}. 
We release these models on Hugging Face\footnote{\url{https://huggingface.co/collections/shikhar-srivastava/vista-350m}}\footnote{\url{https://huggingface.co/collections/shikhar-srivastava/vista-130m}} along with their training pipeline\footnote{\url{https://github.com/shikhar-srivastava/vista-training/}}. \textcolor{red}{[TODO when repos are re-named]} \textcolor{green}{Antonio: I think we renamed the links correctly, the collections and the training repo are public and everything seems fine with the links as far as I can tell} \textcolor{green}{. Shikhar: Yep -- this is done.}

\paragraph{Training Data}

For the 130M parameter models, we used data from \texttt{fish-food}\footnote{\url{https://huggingface.co/datasets/goldfish-models/fish-food}} \citep{chang2024goldfish}, with each language's training corpus sized to approximately 1\,GB $\times$ its byte premium (or `BP', as discussed below).
Each tokenizer is trained on the equivalent of 300\,MB of English data in the corresponding language~\citep{arnett2025crosslingual}.
For the larger 350M models, we use FineWeb and FineWeb-2~\citep{penedo2024fineweb,penedo2025fineweb2}, with a Chinchilla optimal~\citep{hoffmann2022training} 7 billion tokens (${\sim}$31\,GiB tokenized) for English.  
We control for character-encoding efficiency, following \citet{chang2024goldfish}, by normalizing the corpus size according to the \emph{byte premium} (BP), i.e.\ the ratio of UTF-8 bytes required to encode content-matched text in a language $L$ vs.\ English~\citep{arnett2024bit}.
Each non-English language's training data is further byte premium (BP)-scaled to match: $\text{target\_bytes} = \text{BP} \times \text{(English ref. bytes)}$, yielding 42--85\,GiB of tokenized data depending on the language and vocabulary size (see Table~\ref{tab:fw2-350m-data}).
The 130M models are trained for 10 epochs over their BP-scaled sets, matching the multi-epoch regime advocated for smaller monolingual budgets~\citep{muennighoff2023scaling,chang2024goldfish}. The 350M models are trained for 1 epoch on their BP-scaled Chinchilla-optimal sets, except Urdu which uses 3--4 epochs to compensate for limited FineWeb-2 availability (${\sim}$24\% of the BP-scaled target). We do not train an Amharic 350M model, because FineWeb-2 contains less than 5\% of the data necessary to optimally train a model of that scale. VISTA architecture and training details are discussed in detail in Appendix~\ref{app:vista_models}.

\paragraph{Tokenizer Manipulations}

For each language, we train tokenizers from two families: (i) BPE and (ii) Unigram. For each language and tokenizer family, we train tokenizers with vocabulary sizes $v\!\in\!\{8192,\allowbreak\ 16384,\allowbreak\ 32768,\allowbreak\ 49152,\allowbreak\ 65536,\allowbreak\ 81920,\allowbreak\ 98304,\allowbreak\ 114688\}$, following \citet{arnett2025crosslingual}.  The language modeling head dimensions are adjusted to $v$, with all other architectural and optimization hyperparameters held fixed (see Appendix Table~\ref{tab:hyperparameters}). This design isolates the effect of language and vocabulary size on compression while controlling for compute and model capacity. 
For each language, we train a model for each combination of tokenizer type and vocabulary size. Due to computational resource constraints, at the larger scale, we train only on two of the vocabulary sizes, 8192 and 98304, plus 32768 and 65536 for English.

\subsection{Intrinsic Tokenizer Metrics}

\textcolor{red}{i think gini coefficient is missing from table 2} \textcolor{green}{Antonio: added gini\_coefficient to Table~\ref{tab:metrics_summary} with a description in Appendix~\ref{app:tokenizer_metrics}; it is implemented in the eval suite, so no dagger. It is the \textit{fairness} Gini over per-language token costs, and it is only meaningful for TokSuite}

We use 27 previously proposed intrinsic tokenizer metrics from the different categories discussed in Section \ref{sec:related_work} and the two U-statistic metrics we introduced in Section \ref{sec:ustat_deep_dive}. We list the metrics used in Appendix \ref{app:tokenizer_metrics} and provide additional description of each.
To compute these metrics, we use the \texttt{tokenizer-intrinsic-evals} (TokEval)\footnote{\url{https://github.com/cimeister/tokenizer-intrinsic-evals}} suite \citep{meister2026tokeval}, a unified framework for tokenizer evaluation that provides implementations of information-theoretic, morphological, and statistical metrics, among others. For metrics not in the evaluation suite, we implemented them in our own fork (marked with $\dagger$ in Table~\ref{tab:metrics_summary}, Appendix~\ref{app:tokenizer_metrics}). The fork is available online, linked from our main repository.\footnote{\url{https://github.com/AntonioLopardo/tokenizer-metric-multilingual-correlations}}
\textcolor{red}{Are we going to merge this? or otherwise make it public? and do we want to link to the swiss-ai repo?} \textcolor{green}{Antonio: the fork will be public and linked and referenced in the main repo}
For test data we use samples from the OSCAR corpus \citep{abadji2022towards}.

\subsection{Model Evaluations}

To quantify the performance impact of different tokenization processes, we carefully selected a group of downstream task benchmarks. In total, we use 48 distinct benchmarks spanning linguistic acceptability, natural language understanding, commonsense reasoning, world knowledge and question answering, multi-task suites, token-level (orthographic) tasks, and perplexity, across nine languages; the per-suite lists are given in Tables~\ref{tab:toksuite_eval_list}, \ref{tab:timtc_evals}, and \ref{tab:vista_list}. A model's aggregate performance in a language is the unweighted mean over that language's benchmarks outside the token-level and perplexity categories, with the tasks of a benchmark averaged first; it is the Avg.\ of the evaluation tables in the Appendix. We use the LM Evaluation Harness  \cite{eval-harness,biderman2024lessons} for evaluation. For the tasks not implemented in the harness, we implemented them ourselves. \textcolor{red}{Can we indicate these in the tables as well? And did we try to merge them?} \textcolor{green}{Antonio: done, the tasks we implemented are now marked with $\dagger$ in all three eval tables, verified by diffing our fork against the upstream harness} This also enabled us to adapt the harness to support custom models, especially the TIMTC models which use non-standardized tokenizers. We selected the benchmarks using a few criteria: first, we chose tasks that were appropriate for the scale of models we evaluate in these experiments (100M-1B parameters). Second, we selected tasks that we expect to show sensitivity to tokenization variation, particularly linguistic benchmarks like BLiMP and other NLU tasks. Finally, we select an additional set of common tasks such as commonsense reasoning and question answering. The full list of benchmark tasks is provided in Appendix \ref{app:eval-data-details}.

\begin{figure*}[ht]
    \centering
    \includegraphics[width=0.95\linewidth]{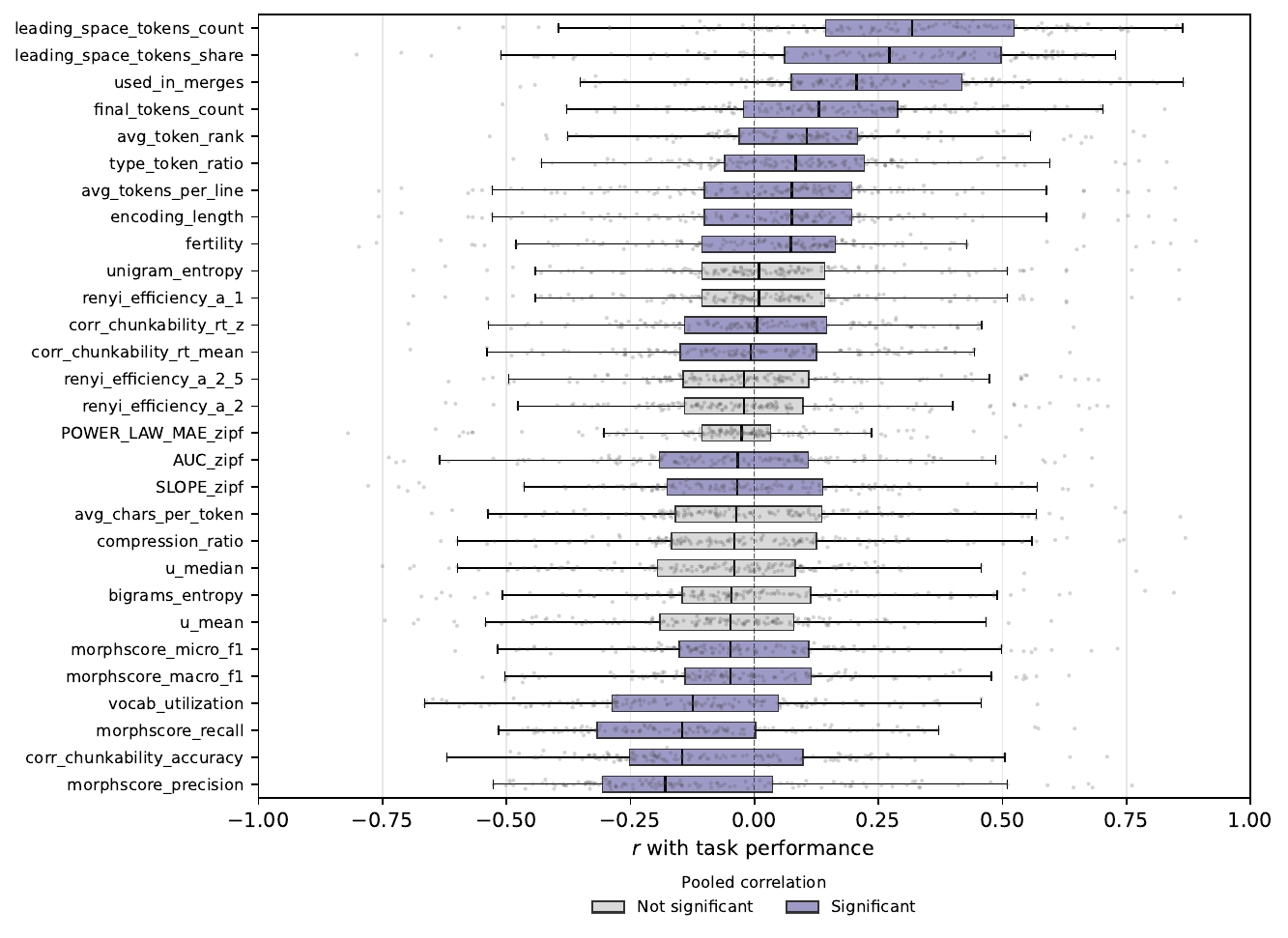}
    \caption{Each row represents the distribution of correlation coefficients (x-axis) between a tokenizer metric (y-axis) and individual tasks. The color of the box plot represents whether the overall correlation between the tokenizer metric and aggregate performance is significant. }
    \label{fig:by_metric_corr_dist}
\end{figure*}

\section{Tokenizers \textit{Do} Affect Model Performance} \label{sec:tokenizers_affect_performance}

Before addressing our research questions, we describe the results from the tokenizer and model evaluation. We release the raw tokenizer evaluation metrics and language model benchmark scores on GitHub\footnote{\url{https://github.com/AntonioLopardo/tokenizer-metric-multilingual-correlations}} and the full model evaluation results are presented in Appendix \ref{app:full_eval_results}.

\paragraph{TokSuite models.}
Consistent with previous work, we find that across multiple model suites, many evaluations, and several languages, the tokenizer used to train a model significantly impacts performance. 
The range of aggregate performance was 4.5 percentage points averaged over the five languages (3.4 to 8.0 points within each language), with performance on individual benchmarks differing by up to 46 points (for ANALOGICAL, \citealp{wijesiriwardene-etal-2023-analogical}; the lowest performance was 36\% and the highest was 82\%). 

Some of these differences may be driven by the vocabulary size, which also impacts the number of embedding parameters, and therefore model size. We find a positive correlation between vocabulary size and average performance for these models, which, after correcting for the five languages tested, is significant only for Italian ($\rho=0.81$, $p=0.002$). It is not significant for English ($\rho=0.65$, $p=0.06$), for the other three languages, or for the average over languages ($\rho=0.48$). All 14 models share the same non-embedding architecture, so vocabulary size alone moves their total size from 1.3B to 2.3B parameters. 
But vocab size doesn't account for all differences. XGLM, BLOOM, and Gemma all have vocab sizes of around 256,000, yet they are spread across the ranking of the 14 models in aggregate performance. While these models differ in many aspects, this suggests that other tokenize features besides vocabulary size may drive model performance more.

\paragraph{TIMTC models.}

After evaluating TIMTC on our own set of evaluations, we find that there are differences in performance between the models in the suite, as found in \citet{schmidt-etal-2024-tokenization}. Across the different tokenizer conditions, average performance differed by up to 5.0 percentage points across models. The maximum range in performance per benchmark was 28 points.

\paragraph{VISTA models.} 
The VISTA models also exhibit differences in performance depending on tokenizer, with differences in aggregate performance of 1.7 to 3.8 percentage points within each model size, depending on the language. As with the TokSuite models, we do not find a significant relationship between vocabulary size and model performance overall ($\rho=0.13$, $p=0.21$), although it is significant for English ($\rho=0.56$, $p=0.02$ after correcting for the five languages tested), driven by the eight 350M models ($\rho=0.93$), where the largest vocabulary also grows the embedding table from 17M to 201M parameters (results in Appendix \ref{app:vista_vocab_size_perf_corr}).

\section{No Intrinsic Tokenizer Metric Reliably Predicts Downstream Performance} \label{sec:individual_metric_corr}

\begin{figure*}[ht!]
    \centering
    \begin{subfigure}{0.48\textwidth}
        \centering
        \includegraphics[height=9.5cm]{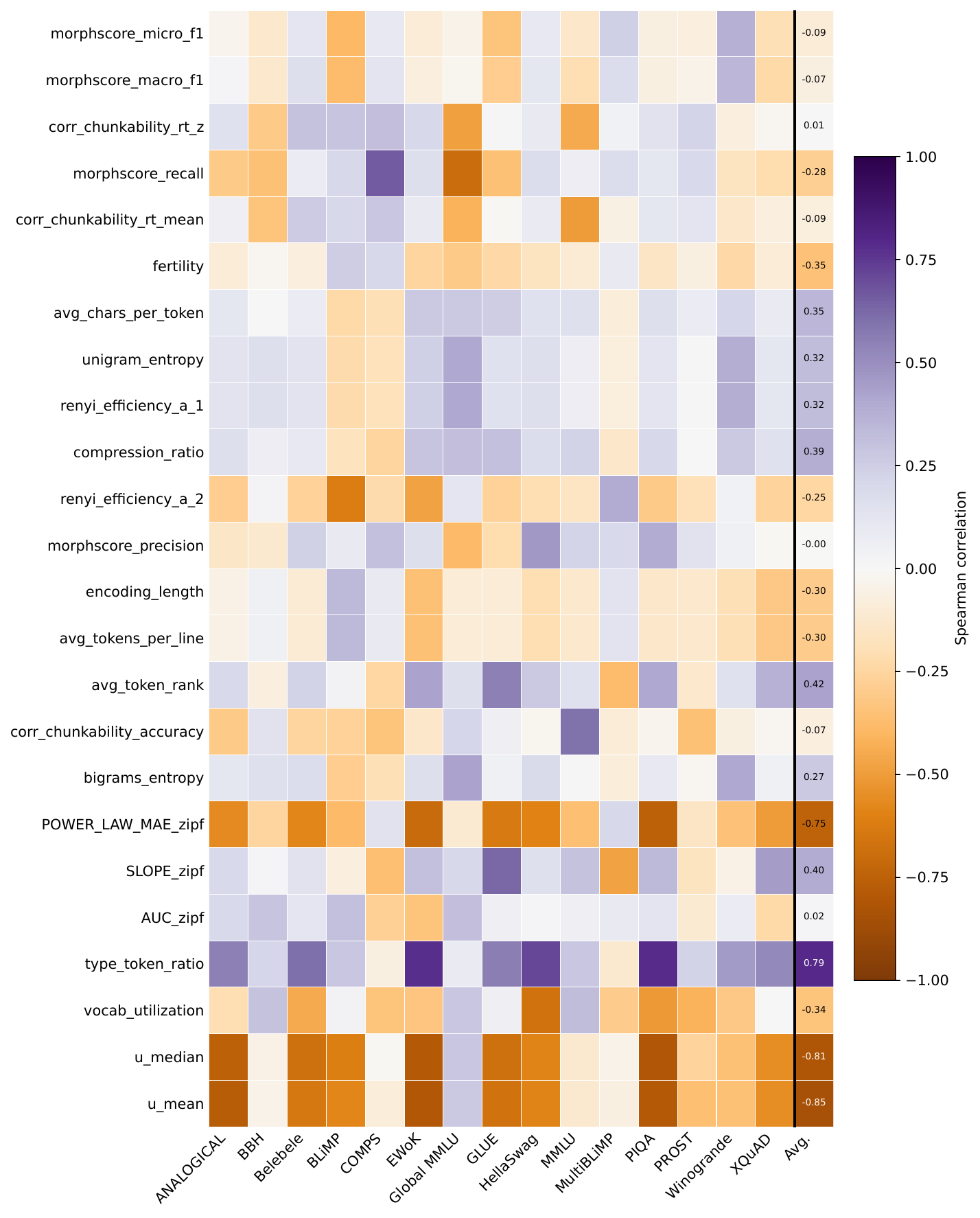}
        \caption{TokSuite}
        \label{fig:toksuite_english_results}
    \end{subfigure}
    \begin{subfigure}{0.48\textwidth}
        \centering
        \includegraphics[height=9.5cm]{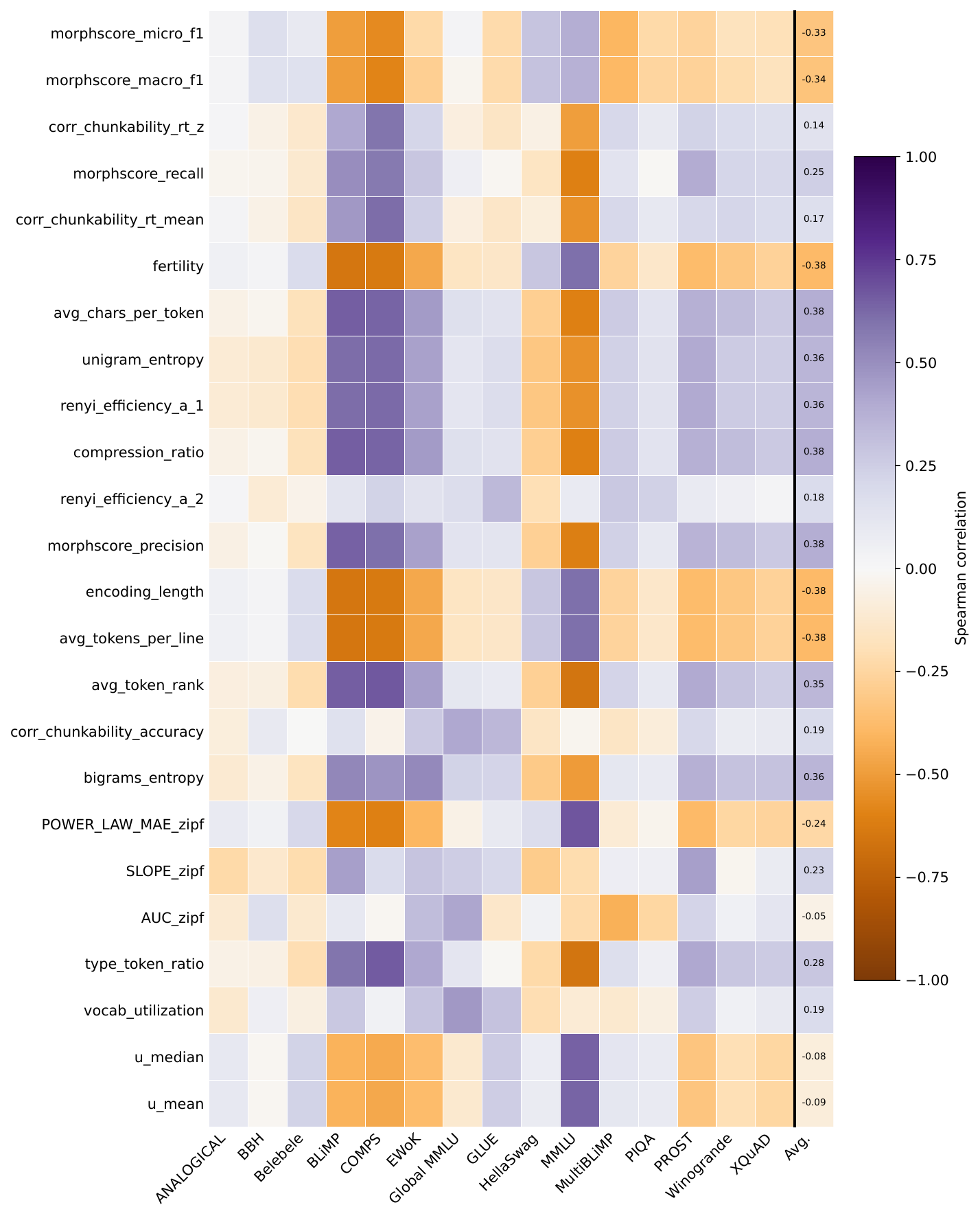}
        \caption{VISTA}
        \label{fig:vista_english_results}
    \end{subfigure}
    \caption{Correlations between intrinsic metric and English model performance for two model suites, including aggregated performance (Avg.). The color of each cell corresponds to direction of correlation and the shade corresponds to the strength. Results for other languages are reported in Appendix \ref{app:full_corr_results}.}
    \label{fig:eng_corr_comparison}
\end{figure*}

\paragraph{Analysis method.} To answer this question, we test for a correlation between model performance and individual intrinsic tokenizer metrics, following \citet{ali-etal-2024-tokenizer}. In the main analysis, we test correlation with task performance, which we normalized by first z-scoring the accuracy scores within each language-task combination (e.g. the Turkish subset of Global MMLU). We test for a correlation across all task scores. We report all correlations between individual metrics and tasks in Appendix \ref{app:full_corr_results}.

We find that for several of the tasks that we evaluate our models on, a large number of models perform at- or near-chance, especially the smaller models. We conduct an additional analysis after excluding these tasks, which we describe in Appendix \ref{app:exclusion_analysis}. The results are qualitatively the same.

\paragraph{Main results.}
We find that 19 of the 29 tokenizer metrics are significantly correlated with z-scored model performance, across all tasks, languages, and model suites. The metrics with the strongest correlations are \texttt{leading\_space\_tokens\_count} ($r=0.27$, $p<0.001$)\footnote{All p-values in this paper have a Bonferroni correction for multiple comparisons, unless otherwise noted. We report corrected and uncorrected p-values in Appendix \ref{app:full_corr_results}.}, \texttt{leading\_space\_tokens\_share} ($r=0.25$, $p<0.001$), and \texttt{used\_in\_merges} ($r=0.22$, $p<0.001$). These are relatively weak correlations.

We found that each metric had a wide range of correlation strengths and directions across all tasks and languages. 
While there are instances where there are strong correlations between a metric and an individual task, no patterns hold across all languages, tasks, and model suites. 
In Figure \ref{fig:by_metric_corr_dist}, we plot the distribution of correlations between each tokenizer metric and individual tasks. We note that almost all tokenizer metrics (including the two metrics with the strongest overall correlations) have at least one task for which there is a moderate ($r > 0.5$) positive correlation and another task for which there is a moderate negative correlation. Not only do the magnitude of the correlations vary by task, but also the direction.

We test how much of the variance can be accounted for simply by model size (in number of total parameters), and find that while for 19 of the metrics, it significantly improves model fit for a linear regression above and beyond model size, the maximum increased variance explained is negligible ($R^2$ gain of $0.015$, $p<0.001$). The full results of this analysis can be found in Table \ref{tab:model_size}.

\begin{figure}[ht!]
    \centering
    \includegraphics[width=0.85\linewidth]{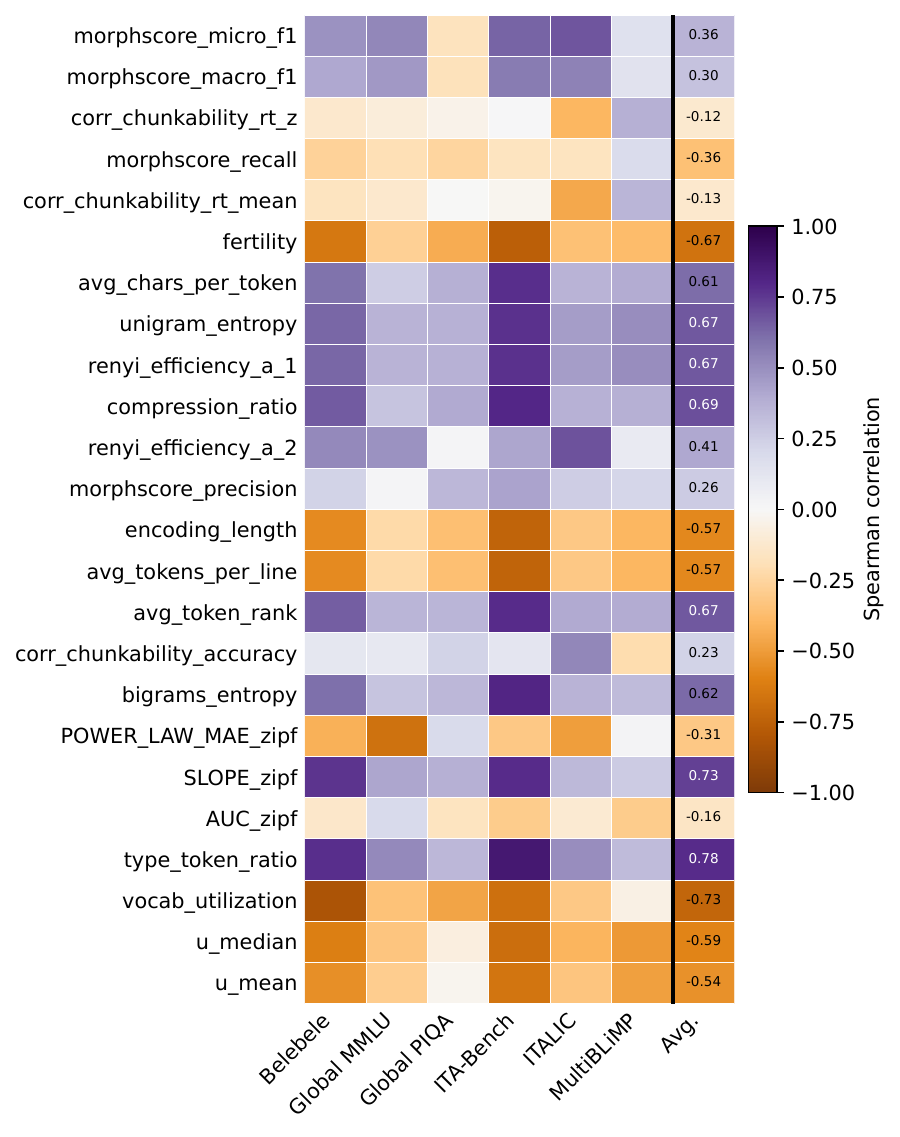}
    \caption{Correlation between each tokenizer metric and the Italian downstream evaluations, individually and in aggregate (Avg.) for the TokSuite models.}
    \label{fig:by_language_italian}
\end{figure}

If we look at the correlations between metrics and tasks within a single language and a single model suite, there are some trends.
Figure \ref{fig:eng_corr_comparison} shows the correlations between each tokenizer metric and both individual and aggregated English tasks for the TokSuite and English VISTA models.
For example, U-statistic metrics are strongly negatively correlated with downstream performance in English for the TokSuite models across most tasks. 

Comparing within model suite but across languages (cf. Figure \ref{fig:eng_corr_comparison}, left, and Figure \ref{fig:by_language_italian}), the strength of the correlations varies greatly but the direction of the correlation largely holds in aggregate. Considering individual tasks, the correlations are not consitent. For example \texttt{SLOPE\_zipf} is negatively correlated with MultiBLiMP in English ($\rho=-0.48$), but positively correlated in Italian ($\rho=0.26$) for the TokSuite models. Comparing across model suites, the correlations can change again, e.g. \texttt{SLOPE\_zipf} shows almost no correlation with MultiBLiMP in English with the VISTA models ($\rho=0.07$; Figure \ref{fig:eng_corr_comparison}, right). This supports the aggregate analysis that the correlations are unstable across models, tasks, and languages. 

We report the by-language, by-task correlations for each model suite in Figures \ref{fig:toksuite_correlations_full}, \ref{fig:timtc_correlations_full}, and \ref{fig:vista_correlations_full} in the Appendix.

\subsection{U-statistic}

In TokSuite, the U-statistic (\texttt{u\_mean}) correlates more strongly with aggregate English and Turkish performance than any previously proposed metric ($\rho=-0.85$ and $\rho=-0.58$; Figure \ref{fig:eng_corr_comparison} and Figure \ref{fig:toksuite_correlations_full}c), though in the opposite direction than predicted, with lower values going with better models. The English correlation survives controls for segmentation rate, whole-word vocabulary allocation and entry length, but does not generalize across languages or suites. For TIMTC, boundary statistics remain among the more strongly correlated metrics, though with weaker effect sizes. However, for VISTA, U-statistics show no consistent relationship with performance, with correlations that are not significant in any language and change sign across languages. All three suites control for architecture, yet the U-statistic's predictive power varies substantially across them, suggesting that the relationship between linguistic alignment and downstream performance depends on other properties of training.

The metric is also language-dependent (Table~\ref{tab:ustat_components}). Alphabetic languages cluster around $\bar{u} \approx 2.4$, while Chinese and Farsi are ${\sim}2.7\times$ higher ($\bar{u} \approx 6.4$) and show no correlation with performance. Decomposing the U-statistic into its components explains this gap. The $\text{PMI}_{\min}$ component varies more across languages than across tokenizers: cross-tokenizer standard deviations are small across all languages (0.09--0.22), but the mean values diverge sharply, alphabetic languages have negative $\text{PMI}_{\min}$ ($-1.06$ to $-1.70$), while Farsi and Chinese have positive $\text{PMI}_{\min}$ ($3.02$ and $1.93$). In alphabetic scripts, most adjacent character pairs within tokens have moderate cohesion, so the $\text{PMI}_{\min}$ bottleneck is sensitive to tokenization quality: a bad split produces a low-PMI pair that drags the score down. In Chinese, characters are meaning-bearing units that combine productively, yielding uniformly high character-pair PMI; the bottleneck rarely binds, so all tokenizers score similarly. Boundary entropy (BE) varies no more across tokenizers (std 0.13--0.23). \citet{hu2025entropydriven} originally developed these signals for Chinese pre-tokenization, where local contrasts in PMI and entropy effectively identify word boundaries, but aggregated across a tokenizer's full vocabulary, these contrasts wash out. Token frequencies-weighted variants (\texttt{u\_mean\_weighted}, \texttt{u\_mean\_log\_weighted}) do not resolve this, as the issue lies in the underlying character statistics rather than the aggregation; language-specific tuning of $\lambda$ could help if the two components degrade unevenly across scripts, but we leave this to future work.

\begin{table}[h]
\centering
\caption{U-statistic components by language for the TokSuite models (mean and std across the 13 TokSuite tokenizers, excluding ByT5, whose byte-level vocabulary has no character statistics). $\text{PMI}_{\min}$ shows minimal cross-tokenizer variation, especially for Chinese (std 0.09), confirming it reflects script properties rather than tokenizer quality. \textcolor{red}{can you clarify which model suites these are from?} \textcolor{green}{Antonio: TokSuite, now stated in the caption}}
\label{tab:ustat_components}
\small
\begin{tabular}{@{}lrrr@{}}
\toprule
\textbf{Language} & \textbf{$\text{PMI}_{\min}$ (std)} & \textbf{BE (std)} & $\bar{u}$ \\
\midrule
English & $-$1.06 (0.22) & 0.70 (0.16) & 2.44 \\
Italian & $-$1.42 (0.20) & 0.75 (0.14) & 2.42 \\
Turkish & $-$1.70 (0.21) & 0.77 (0.14) & 2.35 \\
\midrule
Farsi   & 3.02 (0.21) & 0.66 (0.13) & 6.31 \\
Chinese & 1.93 (0.09) & 1.00 (0.23) & 6.51 \\
\bottomrule
\end{tabular}
\end{table}

The U-statistic targets a complementary dimension of morphological alignment, whether tokens correspond to coherent character sequences with natural boundaries, using only unsupervised character statistics, without requiring morphological annotations. Where it has discriminative power, in TokSuite English and Turkish, it correlates more strongly with performance than previously proposed metrics, but with lower values going with better models, the opposite of its design. Its predictive power also varies across model suites and breaks down for writing systems where character-level co-occurrence statistics lack the structure the metric relies on. These results suggest that boundary statistics pick up a signal that existing metrics miss, but what that signal measures, and whether it transfers across model families and languages, remains open.

\subsection{Summary}
Together these results show the lack of consistency of the tokenizer metrics to reliably predict model performance across different model suites, tasks, and languages.
Because the results are not only different between languages but also between models, we argue that none of the intrinsic tokenizer evaluations is able to reliably predict model performance.

\section{Combining Metrics Does Not Improve Prediction} \label{sec:multiple_metric_corr}

Here we ask whether combining multiple metrics can better predict downstream performance.
We find that many of the metrics are correlated with each other, which suggests that many of the metrics may be identifying similar or related tokenizer features.
We test whether any combination of metrics better predicts performance across all languages. We fit linear regressions with all combinations of two metrics. We find that although combinations significantly improve model fit, the joint adjusted R$^2$ never exceeds $0.09$, and no pair explains more than $0.04$ of variance beyond its stronger metric alone. Therefore, combining two intrinsic tokenizer evaluations does not seem to meaningfully help predict model performance. 
Additional results and statistical tests are reported in Appendix \ref{app:multiple_predictors}.

\section{Orthographic Tasks Are Weak Predictors}

Here we analyze whether there is a relationship between orthographic tasks and general model performance. In this analysis, we focus on the TokSuite models, as the other models achieve near-zero performance at most of these tasks. We test whether the orthographic tasks are correlated with any of the downstream tasks or aggregated model performance. 
After correcting for the seven task groups tested, only the Farsi TokSuite task correlates significantly with aggregate performance ($\rho=0.86$, $p<0.001$), and the other correlations are weak or not significant. The Farsi result rests on a single 40-item task. We report the results in Appendix \ref{app:orthographic_corr}.
These results call into question whether these tasks should be used to shape tokenizer selection and design. 

\section{Discussion}

\paragraph{Lack of Robust Correlations.} 
While some tokenizer metrics correlate with individual tasks for some models, there is no clear pattern. We demonstrated that tokenizers affect model performance (Section \ref{sec:tokenizers_affect_performance}), but none of the proposed tokenizer evaluations correlates reliably with performance. We find that using two metrics instead of one does not improve predictive power. Based on these results, we would predict that testing for a correlation between a tokenizer metric and performance on a small number of tasks---especially in a single language or in one family of models---may often result in a significant result. However, our analysis suggests that those results are unlikely to generalize to additional tasks, languages, and model families.

The current approach to intrinsic tokenizer evaluation currently does not allow us to determine which tokenizer will result in a better model. More work is needed to be able to determine what an optimal tokenizer looks like and how to identify it.

\paragraph{Different Approaches To Tokenizer Evaluation.}
We have focused on the impact of tokenization on model performance. There are other ways in which tokenizers affect language models. 
Tokenizers in part determine the speed and cost of model training \citep{ahia-etal-2023-languages, petrov2023language,dagan2024getting, ali-etal-2024-tokenizer, asgari2025morphbpe}, because they determine how many computations are needed for a model to process a given text. Therefore compression may be a useful metric, even if it does not correlate with downstream performance. 
Relatedly, tokenizer evaluation could help identify unused or under-used tokens in the vocabulary that may lead to undertrained tokens and subsequently undesirable model behavior \citep{land-bartolo-2024-fishing}. 

Another dimension is multilingual parity. Ideally, tokenizers can contribute to more balanced multilingual performance \citep{petrov2023language, ahia-etal-2023-languages}. By better understanding how tokenization impacts languages differently (\citealp[e.g.][]{arnett2025crosslingual}), we can find ways to develop more balanced tokenizers (\citealp[e.g.][]{limisiewicz-etal-2024-myte,foroutan2025parity}).

\paragraph{The Impact of Tokenizer Design Decisions.}
We do not exhaustively analyze the effect of different tokenizer design decisions, but we do discuss the effect of vocabulary size in Section \ref{sec:tokenizers_affect_performance}. 
Vocabulary size is one of the parameters that can be set for a tokenizer, but which is usually chosen with little rationale \citep{alqahtani2026stop}. 
\citet{tao2024vocab_scaling} find that most model scales benefit from increased vocabulary, suggesting a bigger-is-better principle for vocabulary size. 
However, for the small models used in this study, at a scale where changes in vocabulary size have a large proportional impact on the number of trainable parameters, we do not find a robust effect across our model suites. 
Alongside developing better tokenizer evaluations, we need to develop a better understanding of how different design decisions impact model performance.

\section{Conclusion}

We have conducted, to the best of our knowledge, the most comprehensive study about the relationship between intrinsic tokenizer metrics and downstream model performance. We find that none of these intrinsic measures reliably predicts model performance and we call for a novel approach to tokenizer evaluation. 

\section*{Limitations}

Each of the model suites we use has its own limitations. TokSuite was trained on tokenizers from existing models, which vary in a number of dimensions: each was trained on a different mix of data, the tokenizers vary in vocabulary size, and many aspects of the tokenization algorithm are different including pretokenization and normalization. Therefore in our analyses, we are unable to completely account for each of these variables which may influence the resulting model's performance. 
The TIMTC models include only models trained on English, therefore we are unable to determine to what extent any patterns may generalize to other languages. 
Finally, the VISTA models are trained in a controlled way and are trained for multiple languages. However, they are much smaller than the models in the other suites. 
None of these model suites is able to definitively address the research questions because of these limitations. However, putting these models together and comparing across suites helps us to address some of the weaknesses of the models. 

In our study, we studied models of a variety of sizes, including models up to 1B parameters. That said, our smaller models are generally not very strong. They achieved near-chance performance at several tasks. Analyzing models with poor performance may not provide the same signal as larger models trained on more tokens may provide. We address the effects of poor task performance on our results in Appendix \ref{app:exclusion_analysis}. Additionally, we highlight that small differences in performance at smaller scales still represent differences in performance that intrinsic tokenizer evaluation should seek to predict and explain.

The benchmarks we used to evaluate model performance varied across languages. Some languages had more available benchmarks than others, and these benchmarks vary in size and quality. Additionally, some of the intrinsic tokenizer evaluations relied on annotated data. The chunkability data from \citet{beinborn-etal-2023-cognitive} is only available in English. There is also not sufficient labeled data to calculate MorphScore for Vietnamese. The lack of evaluations in languages other than English is a widespread issue in the field and limits our analyses.

\section*{Acknowledgements}
We thank the Center for Integrated Research Computing (CIRC) at the University of Rochester and CoreWeave for providing the computational resources used in this work. We would like to thank Craig Schmidt for making the TIMTC model checkpoints freely available and for his assistance in setting up the PathPiece library.

\section*{Ethical Considerations}

This work aims to improve the understanding of the impact of tokenization across multiple settings, including multiple languages. We are not aware of any potential risks or negative impacts of this work. In fact, we hope this work improves the understanding of how design decisions may not affect all languages equally.
\bibliography{custom}

@inproceedings{abadji2022towards,
    title = "Towards a Cleaner Document-Oriented Multilingual Crawled Corpus",
    author = "Abadji, Julien  and
      Ortiz Suarez, Pedro  and
      Romary, Laurent  and
      Sagot, Beno{\^\i}t",
    booktitle = "Proceedings of the Thirteenth Language Resources and Evaluation Conference",
    month = jun,
    year = "2022",
    address = "Marseille, France",
    publisher = "European Language Resources Association",
    url = "https://aclanthology.org/2022.lrec-1.463",
    pages = "4344--4355",
}

@misc{eval-harness,
  author       = {Gao, Leo and Tow, Jonathan and Abbasi, Baber and Biderman, Stella and Black, Sid and DiPofi, Anthony and Foster, Charles and Golding, Laurence and Hsu, Jeffrey and Le Noac'h, Alain and Li, Haonan and McDonell, Kyle and Muennighoff, Niklas and Ociepa, Chris and Phang, Jason and Reynolds, Laria and Schoelkopf, Hailey and Skowron, Aviya and Sutawika, Lintang and Tang, Eric and Thite, Anish and Wang, Ben and Wang, Kevin and Zou, Andy},
  title        = {{The Language Model Evaluation Harness}},
  month        = 07,
  year         = 2024,
  publisher    = {Zenodo},
  version      = {v0.4.3},
  doi          = {10.5281/zenodo.12608602},
  url          = {https://zenodo.org/records/12608602}
}

@article{biderman2024lessons,
  title={Lessons from the trenches on reproducible evaluation of language models},
  author={Biderman, Stella and Schoelkopf, Hailey and Sutawika, Lintang and Gao, Leo and Tow, Jonathan and Abbasi, Baber and Aji, Alham Fikri and Ammanamanchi, Pawan Sasanka and Black, Sidney and Clive, Jordan and others},
  journal={arXiv preprint arXiv:2405.14782},
  year={2024}
}

@article{warstadt2020blimp,
  author = {Warstadt, Alex and Parrish, Alicia and Liu, Haokun and Mohananey, Anhad and Peng, Wei and Wang, Sheng-Fu and Bowman, Samuel R.},
  title = {{{BLiMP}: The Benchmark of Linguistic Minimal Pairs for {E}nglish}},
  journal = {Transactions of the Association for Computational Linguistics},
  volume = {8},
  pages = {377--392},
  year = {2020},
  doi = {10.1162/tacl_a_00321},
  url = {https://aclanthology.org/2020.tacl-1.25/}
}

@inproceedings{conneau-etal-2018-xnli,
    title = "{{XNLI}: Evaluating Cross-lingual Sentence Representations}",
    author = "Conneau, Alexis  and
      Rinott, Ruty  and
      Lample, Guillaume  and
      Williams, Adina  and
      Bowman, Samuel  and
      Schwenk, Holger  and
      Stoyanov, Veselin",
    editor = "Riloff, Ellen  and
      Chiang, David  and
      Hockenmaier, Julia  and
      Tsujii, Jun{'}ichi",
    booktitle = "Proceedings of the 2018 Conference on Empirical Methods in Natural Language Processing",
    month = oct # "-" # nov,
    year = "2018",
    address = "Brussels, Belgium",
    publisher = "Association for Computational Linguistics",
    url = "https://aclanthology.org/D18-1269/",
    doi = "10.18653/v1/D18-1269",
    pages = "2475--2485"
}

@inproceedings{zellers2019hellaswag,
  title={{{HellaSwag}: Can a Machine Really Finish Your Sentence?}},
  author={Zellers, Rowan and Holtzman, Ari and Bisk, Yonatan and Farhadi, Ali and Choi, Yejin},
  booktitle={Proceedings of the 57th Annual Meeting of the Association for Computational Linguistics},
  pages={4791--4800},
  year={2019},
  month={jul},
  address={Florence, Italy},
  publisher={Association for Computational Linguistics},
  url={https://aclanthology.org/P19-1472/},
  doi={10.18653/v1/P19-1472}
}

@inproceedings{sakaguchi2020winogrande,
  title={{{WinoGrande}: An Adversarial {W}inograd Schema Challenge at Scale}},
  author={Sakaguchi, Keisuke and Le Bras, Ronan and Bhagavatula, Chandra and Choi, Yejin},
  booktitle={Proceedings of the AAAI Conference on Artificial Intelligence},
  volume={34},
  pages={8732--8740},
  year={2020},
  publisher={AAAI Press},
  url={https://ojs.aaai.org/index.php/AAAI/article/view/6399},
  doi={10.1609/aaai.v34i05.6399}
}

@inproceedings{bandarkar-etal-2024-belebele,
    title = "{The Belebele Benchmark: a Parallel Reading Comprehension Dataset in 122 Language Variants}",
    author = "Bandarkar, Lucas  and
      Liang, Davis  and
      Muller, Benjamin  and
      Artetxe, Mikel  and
      Shukla, Satya Narayan  and
      Husa, Donald  and
      Goyal, Naman  and
      Krishnan, Abhinandan  and
      Zettlemoyer, Luke  and
      Khabsa, Madian",
    editor = "Ku, Lun-Wei  and
      Martins, Andre  and
      Srikumar, Vivek",
    booktitle = "Proceedings of the 62nd Annual Meeting of the Association for Computational Linguistics (Volume 1: Long Papers)",
    month = aug,
    year = "2024",
    address = "Bangkok, Thailand",
    publisher = "Association for Computational Linguistics",
    url = "https://aclanthology.org/2024.acl-long.44/",
    doi = "10.18653/v1/2024.acl-long.44",
    pages = "749--775",
}

@inproceedings{edman-etal-2024-cute,
    title = "{{CUTE}: Measuring {LLM}s' Understanding of Their Tokens}",
    author = "Edman, Lukas  and
      Schmid, Helmut  and
      Fraser, Alexander",
    editor = "Al-Onaizan, Yaser  and
      Bansal, Mohit  and
      Chen, Yun-Nung",
    booktitle = "Proceedings of the 2024 Conference on Empirical Methods in Natural Language Processing",
    month = nov,
    year = "2024",
    address = "Miami, Florida, USA",
    publisher = "Association for Computational Linguistics",
    url = "https://aclanthology.org/2024.emnlp-main.177/",
    doi = "10.18653/v1/2024.emnlp-main.177",
    pages = "3017--3026"
}

@inproceedings{edman2025execute,
    title = "{{EXECUTE}: A Multilingual Benchmark for {LLM} Token Understanding}",
    author = "Edman, Lukas  and
      Schmid, Helmut  and
      Fraser, Alexander",
    editor = "Che, Wanxiang  and
      Nabende, Joyce  and
      Shutova, Ekaterina  and
      Pilehvar, Mohammad Taher",
    booktitle = "Findings of the Association for Computational Linguistics: ACL 2025",
    month = jul,
    year = "2025",
    address = "Vienna, Austria",
    publisher = "Association for Computational Linguistics",
    url = "https://aclanthology.org/2025.findings-acl.95/",
    doi = "10.18653/v1/2025.findings-acl.95",
    pages = "1878--1887",
    ISBN = "979-8-89176-256-5"
}

@inproceedings{he2025xcompsmultilingualbenchmarkconceptual,
  title={{{XCOMPS}: A Multilingual Benchmark of Conceptual Minimal Pairs}},
  author={He, Linyang and Nie, Ercong and Dindar, Sukru Samet and Firoozi, Arsalan and Florea, Adrian and Nguyen, Van and Puffay, Corentin and Shimizu, Riki and Ye, Haotian and Brennan, Jonathan and Schmid, Helmut and Sch{\"u}tze, Hinrich and Mesgarani, Nima},
  booktitle={Proceedings of the 7th Workshop on Research in Computational Linguistic Typology and Multilingual NLP (SIGTYP 2025)},
  year={2025},
  month={aug},
  address={Vienna, Austria},
  publisher={Association for Computational Linguistics},
  url={https://aclanthology.org/2025.sigtyp-1.9/},
  doi={10.18653/v1/2025.sigtyp-1.9},
  pages={75--81}
}

@article{ivanova2024elements,
  title={{Elements of World Knowledge ({EWoK}): A Cognition-Inspired Framework for Evaluating Basic World Knowledge in Language Models}},
  author={Ivanova, Anna A. and Sathe, Aalok and Lipkin, Benjamin and Kumar, Unnathi and Radkani, Setayesh and Clark, Thomas H. and Kauf, Carina and Hu, Jennifer and Pramod, R.T. and Grand, Gabriel and Paulun, Vivian and Ryskina, Maria and Aky{\"u}rek, Ekin and Wilcox, Ethan and Rashid, Nafisa and Choshen, Leshem and Levy, Roger and Fedorenko, Evelina and Tenenbaum, Joshua and Andreas, Jacob},
  journal={Transactions of the Association for Computational Linguistics},
  volume={13},
  pages={1245--1270},
  year={2025},
  publisher={MIT Press},
  url={https://aclanthology.org/2025.tacl-1.57/},
  doi={10.1162/tacl.a.38}
}

@inproceedings{bisk2020piqa,
  title={{PIQA}: Reasoning about Physical Commonsense in Natural Language},
  author={Bisk, Yonatan and Zellers, Rowan and Le Bras, Ronan and Gao, Jianfeng and Choi, Yejin},
  booktitle={Proceedings of the AAAI Conference on Artificial Intelligence},
  volume={34},
  pages={7432--7439},
  year={2020},
  publisher={AAAI Press},
  url={https://ojs.aaai.org/index.php/AAAI/article/view/6239},
  doi={10.1609/aaai.v34i05.6239}
}

@inproceedings{chang2024goldfish,
  title = {Goldfish: Monolingual Language Models for 350 Languages},
  author = {Chang, Tyler A. and Arnett, Catherine and Tu, Zhuowen and Bergen, Benjamin K.},
  booktitle = {Proceedings of the 15th Language Resources and Evaluation Conference (LREC)},
  year = {2026},
  url = {https://www.arxiv.org/abs/2408.10441}
}

@misc{nllb2022flores200,
  title={{No Language Left Behind: Scaling Human-Centered Machine Translation}},
  author       = {{NLLB Team} and
                  Costa-juss{\`a}, Marta R. and
                  Cross, James and
                  {\c{C}}elebi, Onur and
                  Elbayad, Maha and
                  Heafield, Kenneth and
                  Heffernan, Kevin and
                  Kalbassi, Elahe and
                  Lam, Janice and
                  Licht, Daniel and
                  Maillard, Jean and
                  Sun, Anna and
                  Wang, Skyler and
                  Wenzek, Guillaume and
                  Youngblood, Al and
                  Akula, Bapi and
                  Barrault, Lo{\"i}c and
                  Mejia Gonzalez, Gabriel and
                  Hansanti, Prangthip and
                  Hoffman, John and
                  Jarrett, Semarley and
                  Ram Sadagopan, Kaushik and
                  Rowe, Dirk and
                  Spruit, Shannon and
                  Tran, Chau and
                  Andrews, Pierre and
                  Ayan, Necip Fazil and
                  Bhosale, Shruti and
                  Edunov, Sergey and
                  Fan, Angela and
                  Gao, Cynthia and
                  Goswami, Vedanuj and
                  Guzm{\'a}n, Francisco and
                  Koehn, Philipp and
                  Mourachko, Alexandre and
                  Ropers, Christophe and
                  Saleem, Safiyyah and
                  Schwenk, Holger and
                  Wang, Jeff},
  year={2022},
  eprint={2207.04672},
  archivePrefix={arXiv},
  primaryClass={cs.CL},
  url={https://arxiv.org/abs/2207.04672}}

@inproceedings{ahia-etal-2023-languages,
    title = "{Do All Languages Cost the Same? Tokenization in the Era of Commercial Language Models}",
    author = "Ahia, Orevaoghene  and
      Kumar, Sachin  and
      Gonen, Hila  and
      Kasai, Jungo  and
      Mortensen, David  and
      Smith, Noah  and
      Tsvetkov, Yulia",
    editor = "Bouamor, Houda  and
      Pino, Juan  and
      Bali, Kalika",
    booktitle = "Proceedings of the 2023 Conference on Empirical Methods in Natural Language Processing",
    month = dec,
    year = "2023",
    address = "Singapore",
    publisher = "Association for Computational Linguistics",
    url = "https://aclanthology.org/2023.emnlp-main.614/",
    doi = "10.18653/v1/2023.emnlp-main.614",
    pages = "9904--9923"
}

@inproceedings{petrov2023language,
  title={{Language Model Tokenizers Introduce Unfairness Between Languages}},
  author={Petrov, Aleksandar and La Malfa, Emanuele and Torr, Philip H.S. and Bibi, Adel},
  booktitle={Advances in Neural Information Processing Systems},
  volume={36},
  pages={36963--36990},
  year={2023},
  url={https://neurips.cc/virtual/2023/poster/72721}
}

@inproceedings{schmidt-etal-2024-tokenization,
    title = "{Tokenization Is More Than Compression}",
    author = "Schmidt, Craig W  and
      Reddy, Varshini  and
      Zhang, Haoran  and
      Alameddine, Alec  and
      Uzan, Omri  and
      Pinter, Yuval  and
      Tanner, Chris",
    editor = "Al-Onaizan, Yaser  and
      Bansal, Mohit  and
      Chen, Yun-Nung",
    booktitle = "Proceedings of the 2024 Conference on Empirical Methods in Natural Language Processing",
    month = nov,
    year = "2024",
    address = "Miami, Florida, USA",
    publisher = "Association for Computational Linguistics",
    url = "https://aclanthology.org/2024.emnlp-main.40/",
    doi = "10.18653/v1/2024.emnlp-main.40",
    pages = "678--702"
}

@inproceedings{deletang2024language,
  title={{Language Modeling Is Compression}},
  author={Del{\'e}tang, Gr{\'e}goire and Ruoss, Anian and Duquenne, Paul-Ambroise and Catt, Elliot and Genewein, Tim and Mattern, Christopher and Grau-Moya, Jordi and Wenliang, Li Kevin and Aitchison, Matthew and Orseau, Laurent and Hutter, Marcus and Veness, Joel},
  booktitle={The Twelfth International Conference on Learning Representations (ICLR)},
  year={2024},
  url={https://iclr.cc/virtual/2024/poster/17997}
}

@inproceedings{goldman-etal-2024-unpacking,
    title = "{Unpacking Tokenization: Evaluating Text Compression and its Correlation with Model Performance}",
    author = "Goldman, Omer  and
      Caciularu, Avi  and
      Eyal, Matan  and
      Cao, Kris  and
      Szpektor, Idan  and
      Tsarfaty, Reut",
    editor = "Ku, Lun-Wei  and
      Martins, Andre  and
      Srikumar, Vivek",
    booktitle = "Findings of the Association for Computational Linguistics: ACL 2024",
    month = aug,
    year = "2024",
    address = "Bangkok, Thailand",
    publisher = "Association for Computational Linguistics",
    url = "https://aclanthology.org/2024.findings-acl.134/",
    doi = "10.18653/v1/2024.findings-acl.134",
    pages = "2274--2286"
}

@inproceedings{arnett-bergen-2025-language,
    title = "Why do language models perform worse for morphologically complex languages?",
    author = "Arnett, Catherine  and
      Bergen, Benjamin",
    editor = "Rambow, Owen  and
      Wanner, Leo  and
      Apidianaki, Marianna  and
      Al-Khalifa, Hend  and
      Eugenio, Barbara Di  and
      Schockaert, Steven",
    booktitle = "Proceedings of the 31st International Conference on Computational Linguistics",
    month = jan,
    year = "2025",
    address = "Abu Dhabi, UAE",
    publisher = "Association for Computational Linguistics",
    url = "https://aclanthology.org/2025.coling-main.441/",
    pages = "6607--6623"
}

@inproceedings{rust-etal-2021-good,
    title = "How Good is Your Tokenizer? On the Monolingual Performance of Multilingual Language Models",
    author = "Rust, Phillip  and
      Pfeiffer, Jonas  and
      Vuli{\'c}, Ivan  and
      Ruder, Sebastian  and
      Gurevych, Iryna",
    editor = "Zong, Chengqing  and
      Xia, Fei  and
      Li, Wenjie  and
      Navigli, Roberto",
    booktitle = "Proceedings of the 59th Annual Meeting of the Association for Computational Linguistics and the 11th International Joint Conference on Natural Language Processing (Volume 1: Long Papers)",
    month = aug,
    year = "2021",
    address = "Online",
    publisher = "Association for Computational Linguistics",
    url = "https://aclanthology.org/2021.acl-long.243/",
    doi = "10.18653/v1/2021.acl-long.243",
    pages = "3118--3135"
}

@inproceedings{zouhar-etal-2023-tokenization,
    title = "{Tokenization and the Noiseless Channel}",
    author = "Zouhar, Vil{\'e}m  and
      Meister, Clara  and
      Gastaldi, Juan  and
      Du, Li  and
      Sachan, Mrinmaya  and
      Cotterell, Ryan",
    editor = "Rogers, Anna  and
      Boyd-Graber, Jordan  and
      Okazaki, Naoaki",
    booktitle = "Proceedings of the 61st Annual Meeting of the Association for Computational Linguistics (Volume 1: Long Papers)",
    month = jul,
    year = "2023",
    address = "Toronto, Canada",
    publisher = "Association for Computational Linguistics",
    url = "https://aclanthology.org/2023.acl-long.284/",
    doi = "10.18653/v1/2023.acl-long.284",
    pages = "5184--5207"
}

@inproceedings{cognetta-etal-2024-two,
    title = "{Two Counterexamples to Tokenization and the Noiseless Channel}",
    author = "Cognetta, Marco  and
      Zouhar, Vil{\'e}m  and
      Moon, Sangwhan  and
      Okazaki, Naoaki",
    editor = "Calzolari, Nicoletta  and
      Kan, Min-Yen  and
      Hoste, Veronique  and
      Lenci, Alessandro  and
      Sakti, Sakriani  and
      Xue, Nianwen",
    booktitle = "Proceedings of the 2024 Joint International Conference on Computational Linguistics, Language Resources and Evaluation (LREC-COLING 2024)",
    month = may,
    year = "2024",
    address = "Torino, Italia",
    publisher = "ELRA and ICCL",
    url = "https://aclanthology.org/2024.lrec-main.1469/",
    pages = "16897--16906"
}

@inproceedings{hofmann-etal-2021-superbizarre,
    title = "{{Superbizarre Is Not Superb: Derivational Morphology Improves {BERT}{'}s Interpretation of Complex Words}}",
    author = {Hofmann, Valentin  and
      Pierrehumbert, Janet  and
      Sch{\"u}tze, Hinrich},
    editor = "Zong, Chengqing  and
      Xia, Fei  and
      Li, Wenjie  and
      Navigli, Roberto",
    booktitle = "Proceedings of the 59th Annual Meeting of the Association for Computational Linguistics and the 11th International Joint Conference on Natural Language Processing (Volume 1: Long Papers)",
    month = aug,
    year = "2021",
    address = "Online",
    publisher = "Association for Computational Linguistics",
    url = "https://aclanthology.org/2021.acl-long.279/",
    doi = "10.18653/v1/2021.acl-long.279",
    pages = "3594--3608"
}

@inproceedings{liu2025superbpe,
  title={{{SuperBPE}: Space Travel for Language Models}},
  author={Liu, Alisa and Hayase, Jonathan and Hofmann, Valentin and Oh, Sewoong and Smith, Noah A. and Choi, Yejin},
  booktitle={Proceedings of the Conference on Language Modeling (COLM)},
  year={2025},
  url={https://openreview.net/forum?id=lcDRvffeNP}
}

@inproceedings{lotz2025beyond,
  author = {Lotz, Jonas F. and Lopes, Ant{\'o}nio V. and Peitz, Stephan and Setiawan, Hendra and Emili, Leonardo},
  title = {{Beyond Text Compression: Evaluating Tokenizers Across Scales}},
  booktitle = {Proceedings of the 63rd Annual Meeting of the Association for Computational Linguistics (Volume 1: Long Papers)},
  year = {2025},
  month = jul,
  address = {Vienna, Austria},
  publisher = {Association for Computational Linguistics},
  url = {https://aclanthology.org/2025.acl-long.1546/},
  doi = {10.18653/v1/2025.acl-long.1546},
  pages = {32155--32173}
}

@article{jumelet2025multiblimp,
  title={{MultiBLiMP} 1.0: A Massively Multilingual Benchmark of Linguistic Minimal Pairs},
  author={Jumelet, Jaap and Weissweiler, Leonie and Nivre, Joakim and Bisazza, Arianna},
  journal={Transactions of the Association for Computational Linguistics},
  volume={14},
  pages={193--216},
  year={2026},
  publisher={MIT Press},
  doi={10.1162/tacl.a.600},
}

@inproceedings{ponti-etal-2020-xcopa,
    title = "{{XCOPA}: A Multilingual Dataset for Causal Commonsense Reasoning}",
    author = "Ponti, Edoardo Maria  and
      Glava{\v{s}}, Goran  and
      Majewska, Olga  and
      Liu, Qianchu  and
      Vuli{\'c}, Ivan  and
      Korhonen, Anna",
    editor = "Webber, Bonnie  and
      Cohn, Trevor  and
      He, Yulan  and
      Liu, Yang",
    booktitle = "Proceedings of the 2020 Conference on Empirical Methods in Natural Language Processing (EMNLP)",
    month = nov,
    year = "2020",
    address = "Online",
    publisher = "Association for Computational Linguistics",
    url = "https://aclanthology.org/2020.emnlp-main.185/",
    doi = "10.18653/v1/2020.emnlp-main.185",
    pages = "2362--2376"
}

@article{asgari2025morphbpe,
  title={{MorphBPE: A Morpho-Aware Tokenizer Bridging Linguistic Complexity for Efficient LLM Training Across Morphologies}},
  author={Asgari, Ehsaneddin and Kheir, Yassine El and Javaheri, Mohammad Ali Sadraei},
  journal={arXiv preprint arXiv:2502.00894},
  year={2025}
}

@inproceedings{dagan2024getting,
  title={{Getting the Most out of Your Tokenizer for Pre-Training and Domain Adaptation}},
  author={Dagan, Gautier and Synnaeve, Gabriel and Rozi{\`e}re, Baptiste},
  booktitle={Proceedings of the 41st International Conference on Machine Learning},
  year={2024},
  volume={235},
  series={Proceedings of Machine Learning Research},
  publisher={PMLR},
  pages={9784--9805},
  url={https://proceedings.mlr.press/v235/dagan24a.html}}

@inproceedings{ali-etal-2024-tokenizer,
    title = "{Tokenizer Choice For {LLM} Training: Negligible or Crucial?}",
    author = {Ali, Mehdi  and
      Fromm, Michael  and
      Thellmann, Klaudia  and
      Rutmann, Richard  and
      L{\"u}bbering, Max  and
      Leveling, Johannes  and
      Klug, Katrin  and
      Ebert, Jan  and
      Doll, Niclas  and
      Buschhoff, Jasper  and
      Jain, Charvi  and
      Weber, Alexander  and
      Jurkschat, Lena  and
      Abdelwahab, Hammam  and
      John, Chelsea  and
      Ortiz Suarez, Pedro  and
      Ostendorff, Malte  and
      Weinbach, Samuel  and
      Sifa, Rafet  and
      Kesselheim, Stefan  and
      Flores-Herr, Nicolas},
    editor = "Duh, Kevin  and
      Gomez, Helena  and
      Bethard, Steven",
    booktitle = "Findings of the Association for Computational Linguistics: NAACL 2024",
    month = jun,
    year = "2024",
    address = "Mexico City, Mexico",
    publisher = "Association for Computational Linguistics",
    url = "https://aclanthology.org/2024.findings-naacl.247/",
    doi = "10.18653/v1/2024.findings-naacl.247",
    pages = "3907--3924"
}

@inproceedings{bostrom-durrett-2020-byte,
    title = "Byte Pair Encoding is Suboptimal for Language Model Pretraining",
    author = "Bostrom, Kaj  and
      Durrett, Greg",
    editor = "Cohn, Trevor  and
      He, Yulan  and
      Liu, Yang",
    booktitle = "Findings of the Association for Computational Linguistics: EMNLP 2020",
    month = nov,
    year = "2020",
    address = "Online",
    publisher = "Association for Computational Linguistics",
    url = "https://aclanthology.org/2020.findings-emnlp.414",
    doi = "10.18653/v1/2020.findings-emnlp.414",
    pages = "4617--4624",
}

@inproceedings{arnett2025morphscore,
  title        = {Evaluating Morphological Alignment of Tokenizers in 70 Languages},
  author       = {Arnett, Catherine and Hudspeth, Marisa and O'Connor, Brendan},
  booktitle    = {Proceedings of the ICML 2025 Workshop on Tokenization (TokShop)},
  year         = {2025},
  month        = jul,
  url          = {https://openreview.net/forum?id=XYRri1s6pP},
}

@inproceedings{foroutan2025parity,
  title={Parity-Aware Byte-Pair Encoding: Improving Cross-lingual Fairness in Tokenization},
  author={Foroutan, Negar and Meister, Clara and Paul, Debjit and Niklaus, Joel and Ahmadi, Sina and Bosselut, Antoine and Sennrich, Rico},
  booktitle={Proceedings of the 64th Annual Meeting of the Association for Computational Linguistics (Volume 1: Long Papers)},
  year={2026},
  pages={7514--7538},
  publisher={Association for Computational Linguistics},
  url={https://aclanthology.org/2026.acl-long.342/}
}

@article{touvron2023llama1,
  title={{LLaMA: Open and Efficient Foundation Language Models}},
  author={Touvron, Hugo and Lavril, Thibaut and Izacard, Gautier and Martinet, Xavier and Lachaux, Marie-Anne and Lacroix, Timoth{\'e}e and Rozi{\`e}re, Baptiste and Goyal, Naman and Hambro, Eric and Azhar, Faisal and others},
  journal={arXiv preprint arXiv:2302.13971},
  year={2023},
  url={https://arxiv.org/abs/2302.13971}
}

@inproceedings{
sun2025curse,
title={The Curse of Depth in Large Language Models},
author={Wenfang Sun and Xinyuan Song and Pengxiang Li and Lu Yin and Yefeng Zheng and Shiwei Liu},
booktitle={The Thirty-ninth Annual Conference on Neural Information Processing Systems},
year={2025},
url={https://openreview.net/forum?id=Orpf8yDjdj}
}

@misc{radford2018improving,
  title={{Improving Language Understanding by Generative Pre-Training}},
  author={Radford, Alec and Narasimhan, Karthik and Salimans, Tim and Sutskever, Ilya},
  year={2018},
  howpublished={OpenAI Blog},
  url={https://cdn.openai.com/research-covers/language-unsupervised/language_understanding_paper.pdf}
}

@inproceedings{wolf2019huggingface,
  title={{Transformers: State-of-the-Art Natural Language Processing}},
  author={Wolf, Thomas and Debut, Lysandre and Sanh, Victor and Chaumond, Julien and Delangue, Clement and Moi, Anthony and Cistac, Pierric and Rault, Tim and Louf, R{\'e}mi and Funtowicz, Morgan and Davison, Joe and Shleifer, Sam and von Platen, Patrick and Ma, Clara and Jernite, Yacine and Plu, Julien and Xu, Canwen and Le Scao, Teven and Gugger, Sylvain and Drame, Mariama and Lhoest, Quentin and Rush, Alexander},
  booktitle={Proceedings of the 2020 Conference on Empirical Methods in Natural Language Processing: System Demonstrations},
  pages={38--45},
  year={2020},
  month={oct},
  address={Online},
  publisher={Association for Computational Linguistics},
  url={https://aclanthology.org/2020.emnlp-demos.6/},
  doi={10.18653/v1/2020.emnlp-demos.6}
}

@inproceedings{muennighoff2023scaling,
  title = {{Scaling Data-Constrained Language Models}},
  author = {Muennighoff, Niklas and Rush, Alexander and Barak, Boaz and Le Scao, Teven and Tazi, Nouamane and Piktus, Aleksandra and Pyysalo, Sampo and Wolf, Thomas and Raffel, Colin A.},
  booktitle = {Advances in Neural Information Processing Systems},
  volume = {36},
  pages = {50358--50376},
  year = {2023},
  publisher = {Curran Associates, Inc.},
  url = {https://neurips.cc/virtual/2023/poster/70706}
}

@inproceedings{arnett2024bit,
    title = "A Bit of a Problem: Measurement Disparities in Dataset Sizes across Languages",
    author = "Arnett, Catherine  and
      Chang, Tyler A.  and
      Bergen, Benjamin",
    editor = "Melero, Maite  and
      Sakti, Sakriani  and
      Soria, Claudia",
    booktitle = "Proceedings of the 3rd Annual Meeting of the Special Interest Group on Under-resourced Languages @ LREC-COLING 2024",
    month = may,
    year = "2024",
    address = "Torino, Italia",
    publisher = "ELRA and ICCL",
    url = "https://aclanthology.org/2024.sigul-1.1/",
    pages = "1--9"
}

@article{toksuite2025,
  title={{TokSuite: Measuring the Impact of Tokenizer Choice on Language Model Behavior}},
  author={Alt{\i}nta{\c{s}}, G{\"u}l Sena and Ehghaghi, Malikeh and Lester, Brian and Liu, Fengyuan and Zhao, Wanru and Ciccone, Marco and Raffel, Colin},
  year={2025},
  journal={arXiv preprint arXiv:2512.20757},
  url={https://arxiv.org/abs/2512.20757},
  note={Under review at ICLR 2026},
}

@inproceedings{
penedo2024fineweb,
title={{The FineWeb Datasets: Decanting the Web for the Finest Text Data at Scale}},
author={Guilherme Penedo and Hynek Kydl{\'\i}{\v{c}}ek and Loubna Ben allal and Anton Lozhkov and Margaret Mitchell and Colin Raffel and Leandro Von Werra and Thomas Wolf},
booktitle={The Thirty-eight Conference on Neural Information Processing Systems Datasets and Benchmarks Track},
year={2024},
  url={https://neurips.cc/virtual/2024/poster/97513},
}

@inproceedings{penedo2025fineweb2,
  title = {{{FineWeb2}: One Pipeline to Scale Them All---Adapting Pre-Training Data Processing to Every Language}},
  author = {Penedo, Guilherme and Kydl{\'\i}{\v{c}}ek, Hynek and Sabol{\v{c}}ec, Vinko and Messmer, Bettina and Foroutan, Negar and Kargaran, Amir Hossein and Raffel, Colin and Jaggi, Martin and Von Werra, Leandro and Wolf, Thomas},
  booktitle = {Proceedings of the Second Conference on Language Modeling},
  year = {2025},
  month = oct,
  address = {Montreal, Canada},
  url = {https://openreview.net/forum?id=jnRBe6zatP},
  note = {arXiv:2506.20920}
}

@inproceedings{beinborn-etal-2023-cognitive,
  title={Analyzing Cognitive Plausibility of Subword Tokenization},
  author={Beinborn, Lisa and Pinter, Yuval},
  booktitle={Proceedings of the 2023 Conference on Empirical Methods in Natural Language Processing},
  pages={4478--4486},
  year={2023},
  month={dec},
  address={Singapore},
  publisher={Association for Computational Linguistics},
  url={https://aclanthology.org/2023.emnlp-main.272/},
  doi={10.18653/v1/2023.emnlp-main.272}
}

@article{keuleers2012british,
  title={The {B}ritish {L}exicon {P}roject: Lexical decision data for 28,730 monosyllabic and disyllabic {E}nglish words},
  author={Keuleers, Emmanuel and Lacey, Paula and Rastle, Kathleen and Brysbaert, Marc},
  journal={Behavior Research Methods},
  volume={44},
  number={1},
  pages={287--304},
  year={2012},
  month={mar},
  publisher={Springer},
  doi={10.3758/s13428-011-0118-4},
  url={https://doi.org/10.3758/s13428-011-0118-4}
}

@inproceedings{limisiewicz-etal-2023-tokenization,
  title={{Tokenization Impacts Multilingual Language Modeling: Assessing Vocabulary Allocation and Overlap Across Languages}},
  author={Limisiewicz, Tomasz and Balhar, Ji{\v{r}}{\'i} and Mare{\v{c}}ek, David},
  booktitle={Findings of the Association for Computational Linguistics: ACL 2023},
  pages={5661--5681},
  year={2023},
  month={jul},
  address={Toronto, Canada},
  publisher={Association for Computational Linguistics},
  url={https://aclanthology.org/2023.findings-acl.350/},
  doi={10.18653/v1/2023.findings-acl.350}
}

@article{harris1955phoneme,
  title={From phoneme to morpheme},
  author={Harris, Zellig S},
  journal={Language},
  volume={31},
  number={2},
  pages={190--222},
  year={1955},
  publisher={JSTOR}
}

@inproceedings{jin-tanaka-ishii-2006-unsupervised,
  title={{Unsupervised Segmentation of {C}hinese Text by Use of Branching Entropy}},
  author={Jin, Zhihui and Tanaka-Ishii, Kumiko},
  booktitle={Proceedings of the COLING/ACL 2006 Main Conference Poster Sessions},
  pages={428--435},
  year={2006},
  address={Sydney, Australia},
  publisher={Association for Computational Linguistics},
  url={https://aclanthology.org/P06-2056}
}

@article{warstadt2019cola,
  title={{Neural Network Acceptability Judgments}},
  author={Warstadt, Alex and Singh, Amanpreet and Bowman, Samuel R.},
  journal={Transactions of the Association for Computational Linguistics},
  volume={7},
  pages={625--641},
  year={2019},
  publisher={MIT Press},
  url={https://aclanthology.org/Q19-1040/},
  doi={10.1162/tacl_a_00290}
}

@misc{song2022zhoblimp,
  title={{{ZhoBLiMP}: A Systematic Assessment of Language Models with Linguistic Minimal Pairs in {C}hinese}},
  author={Liu, Yikang and Shen, Yeting and Zhu, Hongao and Xu, Lilong and Qian, Zhiheng and Song, Siyuan and Zhang, Kejia and Tang, Jialong and Zhang, Pei and Yang, Baosong and Wang, Rui and Hu, Hai},
  year={2024},
  eprint={2411.06096},
  archivePrefix={arXiv},
  primaryClass={cs.CL},
  url={https://arxiv.org/abs/2411.06096}
}

@inproceedings{yuksekgonul2024turblimp,
  title={{TurBLiMP}: A {T}urkish Benchmark of Linguistic Minimal Pairs},
  author={Ba{\c{s}}ar, Ezgi and Padovani, Francesca and Jumelet, Jaap and Bisazza, Arianna},
  booktitle={Proceedings of the 2025 Conference on Empirical Methods in Natural Language Processing},
  year={2025},
  month={nov},
  address={Suzhou, China},
  publisher={Association for Computational Linguistics},
  url={https://aclanthology.org/2025.emnlp-main.834/},
  doi={10.18653/v1/2025.emnlp-main.834},
  pages={16495--16510}
}

@inproceedings{wang2019superglue,
 author = {Wang, Alex and Pruksachatkun, Yada and Nangia, Nikita and Singh, Amanpreet and Michael, Julian and Hill, Felix and Levy, Omer and Bowman, Samuel},
 booktitle = {Advances in Neural Information Processing Systems},
 editor = {H. Wallach and H. Larochelle and A. Beygelzimer and F. d\textquotesingle Alch\'{e}-Buc and E. Fox and R. Garnett},
 pages = {},
 publisher = {Curran Associates, Inc.},
 title = {{SuperGLUE: A Stickier Benchmark for General-Purpose Language Understanding Systems}},
 url = {https://proceedings.neurips.cc/paper_files/paper/2019/file/4496bf24afe7fab6f046bf4923da8de6-Paper.pdf},
 volume = {32},
 year = {2019}
}

@inproceedings{wijesiriwardene-etal-2023-analogical,
    title = "{{ANALOGICAL} - A Novel Benchmark for Long Text Analogy Evaluation in Large Language Models}",
    author = "Wijesiriwardene, Thilini  and
      Wickramarachchi, Ruwan  and
      Gajera, Bimal  and
      Gowaikar, Shreeyash  and
      Gupta, Chandan  and
      Chadha, Aman  and
      Reganti, Aishwarya Naresh  and
      Sheth, Amit  and
      Das, Amitava",
    editor = "Rogers, Anna  and
      Boyd-Graber, Jordan  and
      Okazaki, Naoaki",
    booktitle = "Findings of the Association for Computational Linguistics: ACL 2023",
    month = jul,
    year = "2023",
    address = "Toronto, Canada",
    publisher = "Association for Computational Linguistics",
    url = "https://aclanthology.org/2023.findings-acl.218/",
    doi = "10.18653/v1/2023.findings-acl.218",
    pages = "3534--3549"
}

@inproceedings{tikhonov2021xwinograd,
    title = "{{I}t{'}s {A}ll in the {H}eads: {U}sing {A}ttention {H}eads as a {B}aseline for {C}ross-{L}ingual {T}ransfer in {C}ommonsense {R}easoning}",
    author = "Tikhonov, Alexey  and
      Ryabinin, Max",
    editor = "Zong, Chengqing  and
      Xia, Fei  and
      Li, Wenjie  and
      Navigli, Roberto",
    booktitle = "Findings of the Association for Computational Linguistics: ACL-IJCNLP 2021",
    month = aug,
    year = "2021",
    address = "Online",
    publisher = "Association for Computational Linguistics",
    url = "https://aclanthology.org/2021.findings-acl.310/",
    doi = "10.18653/v1/2021.findings-acl.310",
    pages = "3534--3546"
}

@inproceedings{xu2020clue,
    title = "{{CLUE}: A {C}hinese Language Understanding Evaluation Benchmark}",
    author = "Xu, Liang  and
      Hu, Hai  and
      Zhang, Xuanwei  and
      Li, Lu  and
      Cao, Chenjie  and
      Li, Yudong  and
      Xu, Yechen  and
      Sun, Kai  and
      Yu, Dian  and
      Yu, Cong  and
      Tian, Yin  and
      Dong, Qianqian  and
      Liu, Weitang  and
      Shi, Bo  and
      Cui, Yiming  and
      Li, Junyi  and
      Zeng, Jun  and
      Wang, Rongzhao  and
      Xie, Weijian  and
      Li, Yanting  and
      Patterson, Yina  and
      Tian, Zuoyu  and
      Zhang, Yiwen  and
      Zhou, He  and
      Liu, Shaoweihua  and
      Zhao, Zhe  and
      Zhao, Qipeng  and
      Yue, Cong  and
      Zhang, Xinrui  and
      Yang, Zhengliang  and
      Richardson, Kyle  and
      Lan, Zhenzhong",
    editor = "Scott, Donia  and
      Bel, Nuria  and
      Zong, Chengqing",
    booktitle = "Proceedings of the 28th International Conference on Computational Linguistics",
    month = dec,
    year = "2020",
    address = "Barcelona, Spain (Online)",
    publisher = "International Committee on Computational Linguistics",
    url = "https://aclanthology.org/2020.coling-main.419/",
    doi = "10.18653/v1/2020.coling-main.419",
    pages = "4762--4772"
}

@article{amirkhani2020farstail,
  title={{FarsTail: A Persian Natural Language Inference Dataset}},
  author={Amirkhani, Hossein and AzariJafari, Mohammad and Pourjafari, Zohreh and Faridan-Jahromi, Soroush and Kouhkan, Zeinab and Amirak, Azadeh},
  journal={arXiv preprint arXiv:2009.08820},
  year={2020}
}

@inproceedings{budur2020turkishnli,
    title = "Data and {R}epresentation for {T}urkish {N}atural {L}anguage {I}nference",
    author = {Budur, Emrah  and
      {\"O}z{\c{c}}elik, R{\i}za  and
      Gungor, Tunga  and
      Potts, Christopher},
    editor = "Webber, Bonnie  and
      Cohn, Trevor  and
      He, Yulan  and
      Liu, Yang",
    booktitle = "Proceedings of the 2020 Conference on Empirical Methods in Natural Language Processing (EMNLP)",
    month = nov,
    year = "2020",
    address = "Online",
    publisher = "Association for Computational Linguistics",
    url = "https://aclanthology.org/2020.emnlp-main.662/",
    doi = "10.18653/v1/2020.emnlp-main.662",
    pages = "8253--8267"
}

@inproceedings{lin-etal-2022-shot,
    title = "Few-shot Learning with Multilingual Generative Language Models",
    author = "Lin, Xi Victoria  and
      Mihaylov, Todor  and
      Artetxe, Mikel  and
      Wang, Tianlu  and
      Chen, Shuohui  and
      Simig, Daniel  and
      Ott, Myle  and
      Goyal, Naman  and
      Bhosale, Shruti  and
      Du, Jingfei  and
      Pasunuru, Ramakanth  and
      Shleifer, Sam  and
      Koura, Punit Singh  and
      Chaudhary, Vishrav  and
      O{'}Horo, Brian  and
      Wang, Jeff  and
      Zettlemoyer, Luke  and
      Kozareva, Zornitsa  and
      Diab, Mona  and
      Stoyanov, Veselin  and
      Li, Xian",
    editor = "Goldberg, Yoav  and
      Kozareva, Zornitsa  and
      Zhang, Yue",
    booktitle = "Proceedings of the 2022 Conference on Empirical Methods in Natural Language Processing",
    month = dec,
    year = "2022",
    address = "Abu Dhabi, United Arab Emirates",
    publisher = "Association for Computational Linguistics",
    url = "https://aclanthology.org/2022.emnlp-main.616/",
    doi = "10.18653/v1/2022.emnlp-main.616",
    pages = "9019--9052"
}

@inproceedings{misra2023comps,
  title={{{COMPS}: Conceptual Minimal Pair Sentences for testing Robust Property Knowledge and its Inheritance in Pre-trained Language Models}},
  author={Misra, Kanishka and Rayz, Julia and Ettinger, Allyson},
  booktitle={Proceedings of the 17th Conference of the European Chapter of the Association for Computational Linguistics},
  pages={2928--2949},
  year={2023},
  month={may},
  address={Dubrovnik, Croatia},
  publisher={Association for Computational Linguistics},
  url={https://aclanthology.org/2023.eacl-main.213/},
}

@inproceedings{hendrycks2021mmlu,
  title={{Measuring Massive Multitask Language Understanding}},
  author={Hendrycks, Dan and Burns, Collin and Basart, Steven and Zou, Andy and Mazeika, Mantas and Song, Dawn and Steinhardt, Jacob},
  booktitle={The Ninth International Conference on Learning Representations (ICLR)},
  year={2021},
  url={https://openreview.net/forum?id=d7KBjmI3GmQ},
  note={arXiv:2009.03300}
}

@inproceedings{artetxe2020xquad,
    title = "{On the Cross-lingual Transferability of Monolingual Representations}",
    author = "Artetxe, Mikel  and
      Ruder, Sebastian  and
      Yogatama, Dani",
    editor = "Jurafsky, Dan  and
      Chai, Joyce  and
      Schluter, Natalie  and
      Tetreault, Joel",
    booktitle = "Proceedings of the 58th Annual Meeting of the Association for Computational Linguistics",
    month = jul,
    year = "2020",
    address = "Online",
    publisher = "Association for Computational Linguistics",
    url = "https://aclanthology.org/2020.acl-main.421/",
    doi = "10.18653/v1/2020.acl-main.421",
    pages = "4623--4637"
}

@inproceedings{li2023cmmlu,
  title={{{CMMLU}: Measuring Massive Multitask Language Understanding in {C}hinese}},
  author={Li, Haonan and Zhang, Yixuan and Koto, Fajri and Yang, Yifei and Zhao, Hai and Gong, Yeyun and Duan, Nan and Baldwin, Timothy},
  booktitle={Findings of the Association for Computational Linguistics: ACL 2024},
  year={2024},
  month={aug},
  address={Bangkok, Thailand},
  publisher={Association for Computational Linguistics},
  url={https://aclanthology.org/2024.findings-acl.671/},
  doi={10.18653/v1/2024.findings-acl.671},
  pages={11260--11285}
}

@inproceedings{singh2024globalmmlu,
  title={{Global {MMLU}: Understanding and Addressing Cultural and Linguistic Biases in Multilingual Evaluation}},
  author={Singh, Shivalika and Romanou, Angelika and Fourrier, Cl{\'e}mentine and Adelani, David Ifeoluwa and Ngui, Jian Gang and Vila-Suero, Daniel and Limkonchotiwat, Peerat and Marchisio, Kelly and Leong, Wei Qi and Susanto, Yosephine and Ng, Raymond and Longpre, Shayne and Ruder, Sebastian and Ko, Wei-Yin and Bosselut, Antoine and Oh, Alice and Martins, Andre and Choshen, Leshem and Ippolito, Daphne and Ferrante, Enzo and Fadaee, Marzieh and Ermis, Beyza and Hooker, Sara},
  booktitle={Proceedings of the 63rd Annual Meeting of the Association for Computational Linguistics (Volume 1: Long Papers)},
  year={2025},
  month={jul},
  address={Vienna, Austria},
  publisher={Association for Computational Linguistics},
  url={https://aclanthology.org/2025.acl-long.919/},
  doi={10.18653/v1/2025.acl-long.919},
  pages={18761--18799}
}

@inproceedings{romanou2024include,
  title={{INCLUDE}: Evaluating Multilingual Language Understanding with Regional Knowledge},
  author={Romanou, Angelika and Foroutan, Negar and Sotnikova, Anna and Chen, Zeming and Nelaturu, Sree Harsha and Singh, Shivalika and Maheshwary, Rishabh and Altomare, Micol and Haggag, Mohamed A. and Snegha, A. and Amayuelas, Alfonso and Amirudin, Azril Hafizi and Aryabumi, Viraat and Boiko, Danylo and Chang, Michael and Chim, Jenny and Cohen, Gal and Dalmia, Aditya Kumar and Diress, Abraham and Duwal, Sharad and Dzenhaliou, Daniil and Erazo Florez, Daniel Fernando and Farestam, Fabian and Imperial, Joseph Marvin and Islam, Shayekh Bin and Isotalo, Perttu and Jabbarishiviari, Maral and Karlsson, B{\"o}rje F. and Khalilov, Eldar and Klamm, Christopher and Koto, Fajri and Krzemi{\'n}ski, Dominik and de Melo, Gabriel Adriano and Montariol, Syrielle and Nan, Yiyang and Niklaus, Joel and Novikova, Jekaterina and Obando Ceron, Johan Samir and Paul, Debjit and Ploeger, Esther and Purbey, Jebish and Rajwal, Swati and Ravi, Selvan Sunitha and Rydell, Sara and Santhosh, Roshan and Sharma, Drishti and Skenduli, Marjana Prifti and Soltani Moakhar, Arshia and Soltani Moakhar, Bardia and Tamir, Ran and Tarun, Ayush Kumar and Wasi, Azmine Toushik and Weerasinghe, Thenuka Ovin and Yilmaz, Serhan and Zhang, Mike and Schlag, Imanol and Fadaee, Marzieh and Hooker, Sara and Bosselut, Antoine},
  booktitle={The Thirteenth International Conference on Learning Representations (ICLR)},
  year={2025},
  url={https://openreview.net/forum?id=k3gCieTXeY},
}

@misc{hupkes2025multiloko,
  title={{{MultiLoKo}: A Multilingual Local Knowledge Benchmark for {LLM}s Spanning 31 Languages}},
  author={Hupkes, Dieuwke and Bogoychev, Nikolay},
  year={2025},
  eprint={2504.10356},
  archivePrefix={arXiv},
  primaryClass={cs.CL},
  url={https://arxiv.org/abs/2504.10356}
}

@inproceedings{suzgun2023bbh,
  title={{Challenging BIG-Bench Tasks and Whether Chain-of-Thought Can Solve Them}},
    author = {Suzgun, Mirac  and
      Scales, Nathan  and
      Sch{\"a}rli, Nathanael  and
      Gehrmann, Sebastian  and
      Tay, Yi  and
      Chung, Hyung Won  and
      Chowdhery, Aakanksha  and
      Le, Quoc  and
      Chi, Ed  and
      Zhou, Denny  and
      Wei, Jason},
    editor = "Rogers, Anna  and
      Boyd-Graber, Jordan  and
      Okazaki, Naoaki",
    booktitle = "Findings of the Association for Computational Linguistics: ACL 2023",
    month = jul,
    year = "2023",
    address = "Toronto, Canada",
    publisher = "Association for Computational Linguistics",
    url = "https://aclanthology.org/2023.findings-acl.824/",
    doi = "10.18653/v1/2023.findings-acl.824",
    pages = "13003--13051"
}

@inproceedings{seveso2025italic,
  title={{{ITALIC}: An {I}talian Culture-Aware Natural Language Benchmark}},
  author={Seveso, Andrea and Potert{\`i}, Daniele and Federici, Edoardo and Mezzanzanica, Mario and Mercorio, Fabio},
  booktitle={Proceedings of the 2025 Conference of the Nations of the Americas Chapter of the Association for Computational Linguistics: Human Language Technologies (Volume 1: Long Papers)},
  year={2025},
  address={Albuquerque, New Mexico},
  publisher={Association for Computational Linguistics},
  url={https://aclanthology.org/2025.naacl-long.68/},
  doi={10.18653/v1/2025.naacl-long.68},
  pages={1469--1478}
}

@inproceedings{uzan2026charbench,
  title={CharBench: Evaluating the Role of Tokenization in Character-Level Tasks},
  author={Uzan, Omri and Pinter, Yuval},
  booktitle={Proceedings of the AAAI Conference on Artificial Intelligence},
  volume={40},
  pages={33296--33304},
  doi={10.1609/aaai.v40i39.40615},
  year={2026}
}

@inproceedings{efrat2023lmentry,
    title = "{{LM}entry: A Language Model Benchmark of Elementary Language Tasks}",
    author = "Efrat, Avia  and
      Honovich, Or  and
      Levy, Omer",
    editor = "Rogers, Anna  and
      Boyd-Graber, Jordan  and
      Okazaki, Naoaki",
    booktitle = "Findings of the Association for Computational Linguistics: ACL 2023",
    month = jul,
    year = "2023",
    address = "Toronto, Canada",
    publisher = "Association for Computational Linguistics",
    url = "https://aclanthology.org/2023.findings-acl.666/",
    doi = "10.18653/v1/2023.findings-acl.666",
    pages = "10476--10501"
}

@inproceedings{wang2025stringllm,
  title={{{StringLLM}: Understanding the String Processing Capability of Large Language Models}},
  author={Wang, Xilong and Fu, Hao and Wang, Jindong and Gong, Neil Zhenqiang},
  booktitle={The Thirteenth International Conference on Learning Representations (ICLR)},
  year={2025},
  url={https://openreview.net/forum?id=kTXChtaaNO},
  note={arXiv:2410.01208}
}

@article{silberling2024why,
  title        = {{Why {AI} Can’t Spell ‘Strawberry’}},
  author       = {Silberling, Amanda},
  journal      = {TechCrunch},
  year         = {2024},
  date         = {2024-08-27},
  url          = {https://techcrunch.com/2024/08/27/why-ai-cant-spell-strawberry/},
  note         = {Accessed: 2026-01-26},
}

@inproceedings{cosma-etal-2025-strawberry,
    title = "{The Strawberry Problem: Emergence of Character-level Understanding in Tokenized Language Models}",
    author = "Cosma, Adrian  and
      Ruseti, Stefan  and
      Radoi, Emilian  and
      Dascalu, Mihai",
    editor = "Christodoulopoulos, Christos  and
      Chakraborty, Tanmoy  and
      Rose, Carolyn  and
      Peng, Violet",
    booktitle = "Proceedings of the 2025 Conference on Empirical Methods in Natural Language Processing",
    month = nov,
    year = "2025",
    address = "Suzhou, China",
    publisher = "Association for Computational Linguistics",
    url = "https://aclanthology.org/2025.emnlp-main.1434/",
    doi = "10.18653/v1/2025.emnlp-main.1434",
    pages = "28252--28263",
    ISBN = "979-8-89176-332-6"
}

@article{alqahtani2026stop,
  title={{Stop Taking Tokenizers for Granted: They Are Core Design Decisions in Large Language Models}},
  author={Alqahtani, Sawsan and Nayeem, Mir Tafseer and Laskar, Md Tahmid Rahman and Mohiuddin, Tasnim and Bari, M Saiful},
  journal={arXiv preprint arXiv:2601.13260},
  year={2026}
}

@inproceedings{galle-2019-investigating,
    title = "{Investigating the Effectiveness of {BPE}: The Power of Shorter Sequences}",
    author = "Gall{\'e}, Matthias",
    editor = "Inui, Kentaro  and
      Jiang, Jing  and
      Ng, Vincent  and
      Wan, Xiaojun",
    booktitle = "Proceedings of the 2019 Conference on Empirical Methods in Natural Language Processing and the 9th International Joint Conference on Natural Language Processing (EMNLP-IJCNLP)",
    month = nov,
    year = "2019",
    address = "Hong Kong, China",
    publisher = "Association for Computational Linguistics",
    url = "https://aclanthology.org/D19-1141/",
    doi = "10.18653/v1/D19-1141",
    pages = "1375--1381"
}

@inproceedings{
hu2025entropydriven,
title={{Entropy-Driven Pre-Tokenization for Byte-Pair Encoding}},
author={Yifan Hu and Ningyue Liang and Dachuan Zhao and Jonathan Geuter and Varshini Reddy and Craig W Schmidt and Chris Tanner},
booktitle={Tokenization Workshop},
year={2025},
url={https://openreview.net/forum?id=1hcu9Vq4Or}
}

@inproceedings{pagnoni-etal-2025-byte,
    title = "{Byte Latent Transformer: Patches Scale Better Than Tokens}",
    author = "Pagnoni, Artidoro  and
      Pasunuru, Ramakanth  and
      Rodriguez, Pedro  and
      Nguyen, John  and
      Muller, Benjamin  and
      Li, Margaret  and
      Zhou, Chunting  and
      Yu, Lili  and
      Weston, Jason E  and
      Zettlemoyer, Luke  and
      Ghosh, Gargi  and
      Lewis, Mike  and
      Holtzman, Ari  and
      Iyer, Srini",
    editor = "Che, Wanxiang  and
      Nabende, Joyce  and
      Shutova, Ekaterina  and
      Pilehvar, Mohammad Taher",
    booktitle = "Proceedings of the 63rd Annual Meeting of the Association for Computational Linguistics (Volume 1: Long Papers)",
    month = jul,
    year = "2025",
    address = "Vienna, Austria",
    publisher = "Association for Computational Linguistics",
    url = "https://aclanthology.org/2025.acl-long.453/",
    doi = "10.18653/v1/2025.acl-long.453",
    pages = "9238--9258",
    ISBN = "979-8-89176-251-0"
}

@inproceedings{arnett2025crosslingual,
  author    = {Arnett, Catherine and Chang, Tyler A. and Biderman, Stella and Bergen, Benjamin K.},
  title     = {Explaining and Mitigating Crosslingual Tokenizer Inequities},
  booktitle = {Proceedings of the Thirty-Ninth Annual Conference on Neural Information Processing Systems (NeurIPS)},
  year      = {2025}
}

@article{allenai_arc,
      author    = {Peter Clark  and Isaac Cowhey and Oren Etzioni and Tushar Khot and
                    Ashish Sabharwal and Carissa Schoenick and Oyvind Tafjord},
      title     = {{Think you have Solved Question Answering? Try ARC, the AI2 Reasoning Challenge}},
      journal   = {arXiv:1803.05457v1},
      year      = {2018},
}

@inproceedings{aroca-ouellette-etal-2021-prost,
    title = "{PROST}: Physical Reasoning about Objects through Space and Time",
    author = "Aroca-Ouellette, St{\'e}phane  and
      Paik, Cory  and
      Roncone, Alessandro  and
      Kann, Katharina",
    booktitle = "Findings of the Association for Computational Linguistics: ACL-IJCNLP 2021",
    month = aug,
    year = "2021",
    address = "Online",
    publisher = "Association for Computational Linguistics",
    url = "https://aclanthology.org/2021.findings-acl.404",
    doi = "10.18653/v1/2021.findings-acl.404",
    pages = "4597--4608",
}

@inproceedings{gow-smith-etal-2022-improving,
    title = "{Improving Tokenisation by Alternative Treatment of Spaces}",
    author = "Gow-Smith, Edward  and
      Tayyar Madabushi, Harish  and
      Scarton, Carolina  and
      Villavicencio, Aline",
    editor = "Goldberg, Yoav  and
      Kozareva, Zornitsa  and
      Zhang, Yue",
    booktitle = "Proceedings of the 2022 Conference on Empirical Methods in Natural Language Processing",
    month = dec,
    year = "2022",
    address = "Abu Dhabi, United Arab Emirates",
    publisher = "Association for Computational Linguistics",
    url = "https://aclanthology.org/2022.emnlp-main.786/",
    doi = "10.18653/v1/2022.emnlp-main.786",
    pages = "11430--11443"
}

@inproceedings{hofmann-etal-2020-dagobert,
    title = "{{D}ago{BERT}: {G}enerating Derivational Morphology with a Pretrained Language Model}",
    author = {Hofmann, Valentin  and
      Pierrehumbert, Janet  and
      Sch{\"u}tze, Hinrich},
    editor = "Webber, Bonnie  and
      Cohn, Trevor  and
      He, Yulan  and
      Liu, Yang",
    booktitle = "Proceedings of the 2020 Conference on Empirical Methods in Natural Language Processing (EMNLP)",
    month = nov,
    year = "2020",
    address = "Online",
    publisher = "Association for Computational Linguistics",
    url = "https://aclanthology.org/2020.emnlp-main.316/",
    doi = "10.18653/v1/2020.emnlp-main.316",
    pages = "3848--3861"
}

@inproceedings{klein-tsarfaty-2020-getting,
    title = "{Getting the {\#}{\#}life out of living: How Adequate Are Word-Pieces for Modelling Complex Morphology?}",
    author = "Klein, Stav  and
      Tsarfaty, Reut",
    editor = "Nicolai, Garrett  and
      Gorman, Kyle  and
      Cotterell, Ryan",
    booktitle = "Proceedings of the 17th SIGMORPHON Workshop on Computational Research in Phonetics, Phonology, and Morphology",
    month = jul,
    year = "2020",
    address = "Online",
    publisher = "Association for Computational Linguistics",
    url = "https://aclanthology.org/2020.sigmorphon-1.24/",
    doi = "10.18653/v1/2020.sigmorphon-1.24",
    pages = "204--209"
}

@inproceedings{park-etal-2020-empirical,
    title = "{An Empirical Study of Tokenization Strategies for Various {K}orean {NLP} Tasks}",
    author = "Park, Kyubyong  and
      Lee, Joohong  and
      Jang, Seongbo  and
      Jung, Dawoon",
    editor = "Wong, Kam-Fai  and
      Knight, Kevin  and
      Wu, Hua",
    booktitle = "Proceedings of the 1st Conference of the Asia-Pacific Chapter of the Association for Computational Linguistics and the 10th International Joint Conference on Natural Language Processing",
    month = dec,
    year = "2020",
    address = "Suzhou, China",
    publisher = "Association for Computational Linguistics",
    url = "https://aclanthology.org/2020.aacl-main.17/",
    doi = "10.18653/v1/2020.aacl-main.17",
    pages = "133--142"
}

@inproceedings{vasiu2020romanian,
  author={Vasiu, Mihaela Alexandra and Potolea, Rodica},
  booktitle={2020 IEEE 16th International Conference on Intelligent Computer Communication and Processing (ICCP)}, 
  title={{Enhancing Tokenization by Embedding Romanian Language Specific Morphology}}, 
  year={2020},
  volume={},
  number={},
  pages={243-250},
  doi={10.1109/ICCP51029.2020.9266140}}

@inproceedings{nzeyimana-niyongabo-rubungo-2022-kinyabert,
    title = "{{K}inya{BERT}: a Morphology-aware {K}inyarwanda Language Model}",
    author = "Nzeyimana, Antoine  and
      Niyongabo Rubungo, Andre",
    editor = "Muresan, Smaranda  and
      Nakov, Preslav  and
      Villavicencio, Aline",
    booktitle = "Proceedings of the 60th Annual Meeting of the Association for Computational Linguistics (Volume 1: Long Papers)",
    month = may,
    year = "2022",
    address = "Dublin, Ireland",
    publisher = "Association for Computational Linguistics",
    url = "https://aclanthology.org/2022.acl-long.367/",
    doi = "10.18653/v1/2022.acl-long.367",
    pages = "5347--5363"
}

@article{toraman2023turkish,
author = {Toraman, Cagri and Yilmaz, Eyup Halit and {\c{S}}ahinu{\c{c}}, Furkan and Ozcelik, Oguzhan},
title = {{Impact of Tokenization on Language Models: An Analysis for Turkish}},
year = {2023},
issue_date = {April 2023},
publisher = {Association for Computing Machinery},
address = {New York, NY, USA},
volume = {22},
number = {4},
issn = {2375-4699},
url = {https://doi.org/10.1145/3578707},
doi = {10.1145/3578707},
journal = {ACM Trans. Asian Low-Resour. Lang. Inf. Process.},
month = mar,
articleno = {116},
numpages = {21}
}

@inproceedings{libovicky-helcl-2024-lexically,
    title = "{Lexically Grounded Subword Segmentation}",
    author = "Libovick{\'y}, Jind{\v{r}}ich  and
      Helcl, Jind{\v{r}}ich",
    editor = "Al-Onaizan, Yaser  and
      Bansal, Mohit  and
      Chen, Yun-Nung",
    booktitle = "Proceedings of the 2024 Conference on Empirical Methods in Natural Language Processing",
    month = nov,
    year = "2024",
    address = "Miami, Florida, USA",
    publisher = "Association for Computational Linguistics",
    url = "https://aclanthology.org/2024.emnlp-main.421/",
    doi = "10.18653/v1/2024.emnlp-main.421",
    pages = "7403--7420"
}

@inproceedings{bauwens-delobelle-2024-bpe,
    title = "{BPE}-knockout: {Pruning Pre-existing {BPE} Tokenisers with Backwards-compatible Morphological Semi-supervision}",
    author = "Bauwens, Thomas  and
      Delobelle, Pieter",
    editor = "Duh, Kevin  and
      Gomez, Helena  and
      Bethard, Steven",
    booktitle = "Proceedings of the 2024 Conference of the North American Chapter of the Association for Computational Linguistics: Human Language Technologies (Volume 1: Long Papers)",
    month = jun,
    year = "2024",
    address = "Mexico City, Mexico",
    publisher = "Association for Computational Linguistics",
    url = "https://aclanthology.org/2024.naacl-long.324/",
    doi = "10.18653/v1/2024.naacl-long.324",
    pages = "5810--5832"
}

@inproceedings{arnett-etal-2024-different,
    title = "Different Tokenization Schemes Lead to Comparable Performance in {S}panish Number Agreement",
    author = "Arnett, Catherine  and
      Rivi{\`e}re, Pamela D.  and
      Chang, Tyler A.  and
      Trott, Sean",
    editor = {Nicolai, Garrett  and
      Chodroff, Eleanor  and
      Mailhot, Frederic  and
      {\c{C}}{\"o}ltekin, {\c{C}}a{\u{g}}r{\i}},
    booktitle = "Proceedings of the 21st SIGMORPHON workshop on Computational Research in Phonetics, Phonology, and Morphology",
    month = jun,
    year = "2024",
    address = "Mexico City, Mexico",
    publisher = "Association for Computational Linguistics",
    url = "https://aclanthology.org/2024.sigmorphon-1.4/",
    doi = "10.18653/v1/2024.sigmorphon-1.4",
    pages = "32--38"
}

@inproceedings{machavcek2018morphological,
  title={{Morphological and Language-Agnostic Word Segmentation for NMT}},
author="Mach{\'a}{\v{c}}ek, Dominik
and Vidra, Jon{\'a}{\v{s}}
and Bojar, Ond{\v{r}}ej",
editor="Sojka, Petr
and Hor{\'a}k, Ale{\v{s}}
and Kope{\v{c}}ek, Ivan
and Pala, Karel",
booktitle="Text, Speech, and Dialogue",
year="2018",
publisher="Springer International Publishing",
address="Cham",
pages="277--284",
isbn="978-3-030-00794-2"
}

@inproceedings{saleva-lignos-2021-effectiveness,
    title = "{The Effectiveness of Morphology-aware Segmentation in Low-Resource Neural Machine Translation}",
    author = "Saleva, Jonne  and
      Lignos, Constantine",
    editor = "Sorodoc, Ionut-Teodor  and
      Sushil, Madhumita  and
      Takmaz, Ece  and
      Agirre, Eneko",
    booktitle = "Proceedings of the 16th Conference of the European Chapter of the Association for Computational Linguistics: Student Research Workshop",
    month = apr,
    year = "2021",
    address = "Online",
    publisher = "Association for Computational Linguistics",
    url = "https://aclanthology.org/2021.eacl-srw.22/",
    doi = "10.18653/v1/2021.eacl-srw.22",
    pages = "164--174"
}

@article{choo2023study,
  title={A study on the evaluation of tokenizer performance in natural language processing},
  author={Choo, Sanghyun and Kim, Wonjoon},
  journal={Applied Artificial Intelligence},
  volume={37},
  number={1},
  pages={2175112},
  year={2023},
  publisher={Taylor \& Francis}
}

@misc{MosaicML2023Introducing,
    author    = {{Mosaic Research}},
    title     = {{Introducing MPT-7B: A New Standard for Open-Source, Commercially Usable LLMs}},
    year      = {2023},
    url       = {https://www.databricks.com/blog/mpt-7b},
    note      = {Accessed: 2026-01-28},
}

@inproceedings{tao2024vocab_scaling,
 author = {Tao, Chaofan and Liu, Qian and Dou, Longxu and Muennighoff, Niklas and Wan, Zhongwei and Luo, Ping and Lin, Min and Wong, Ngai},
 booktitle = {Advances in Neural Information Processing Systems},
 doi = {10.52202/079017-3626},
 pages = {114147--114179},
 publisher = {Curran Associates, Inc.},
 title = {{Scaling Laws with Vocabulary: Larger Models Deserve Larger Vocabularies}},
 volume = {37},
 year = {2024}
}

@inproceedings{limisiewicz-etal-2024-myte,
    title = "{{MYTE}: Morphology-Driven Byte Encoding for Better and Fairer Multilingual Language Modeling}",
    author = "Limisiewicz, Tomasz  and
      Blevins, Terra  and
      Gonen, Hila  and
      Ahia, Orevaoghene  and
      Zettlemoyer, Luke",
    editor = "Ku, Lun-Wei  and
      Martins, Andre  and
      Srikumar, Vivek",
    booktitle = "Proceedings of the 62nd Annual Meeting of the Association for Computational Linguistics (Volume 1: Long Papers)",
    month = aug,
    year = "2024",
    address = "Bangkok, Thailand",
    publisher = "Association for Computational Linguistics",
    url = "https://aclanthology.org/2024.acl-long.804/",
    doi = "10.18653/v1/2024.acl-long.804",
    pages = "15059--15076"
}

@inproceedings{land-bartolo-2024-fishing,
    title = "{Fishing for Magikarp: Automatically Detecting Under-trained Tokens in Large Language Models}",
    author = "Land, Sander  and
      Bartolo, Max",
    editor = "Al-Onaizan, Yaser  and
      Bansal, Mohit  and
      Chen, Yun-Nung",
    booktitle = "Proceedings of the 2024 Conference on Empirical Methods in Natural Language Processing",
    month = nov,
    year = "2024",
    address = "Miami, Florida, USA",
    publisher = "Association for Computational Linguistics",
    url = "https://aclanthology.org/2024.emnlp-main.649/",
    doi = "10.18653/v1/2024.emnlp-main.649",
    pages = "11631--11646"
}

@book{Zipf49,
    author = {Zipf, George K.},
    title = {Human Behavior and the Principle of Least Effort},
    publisher = {Addison-Wesley},
    year = {1949}
}

@article{Piantadosi2014Zipf,
    author = {Piantadosi, Steven T.},
    title = {Zipf's word frequency law in natural language: A critical review and future directions},
    journal = {Psychonomic Bulletin \& Review},
    volume = {21},
    number = {5},
    pages = {1112--1130},
    year = {2014},
    month = oct,
    publisher = {Springer},
    doi = {10.3758/s13423-014-0585-6},
    url = {https://doi.org/10.3758/s13423-014-0585-6}
}

@article{clauset2009powerlaw,
    author = {Clauset, Aaron and Shalizi, Cosma Rohilla and Newman, M. E. J.},
    title = {Power-law distributions in empirical data},
    journal = {SIAM Review},
    volume = {51},
    number = {4},
    pages = {661--703},
    year = {2009},
    doi = {10.1137/070710111}
}

@article{hoffmann2022training,
  title={Training compute-optimal large language models},
  author={Hoffmann, Jordan and Borgeaud, Sebastian and Mensch, Arthur and Buchatskaya, Elena and Cai, Trevor and Rutherford, Eliza and de Las Casas, Diego and Hendricks, Lisa Anne and Welbl, Johannes and Clark, Aidan and others},
  journal={arXiv preprint arXiv:2203.15556},
  year={2022}
}

@misc{chang2025globalpiqa,
  title={{Global {PIQA}: Evaluating Physical Commonsense Reasoning Across 100+ Languages and Cultures}},
  author={Chang, Tyler A. and Arnett, Catherine and others},
  year={2025},
  eprint={2510.24081},
  archivePrefix={arXiv},
  primaryClass={cs.CL},
  url={https://arxiv.org/abs/2510.24081}
}

@misc{farsi2024syntranfa,
  title={{SynTran-fa: Generating Comprehensive Answers for {F}arsi {QA} Pairs via Syntactic Transformation}},
  author={Farsi, Farhan and Sabouri, Sadra and Kashfipour, Kian and Gooran, Soroush and Sameti, Hossein and Asgari, Ehsaneddin},
  year={2024},
  doi={10.20944/preprints202410.1684.v1},
  note={Preprint}
}

@misc{meister2026tokeval,
  title={{TokEval: A Tokenizer Evaluation Suite}},
  author={Meister, Clara},
  year={2026},
  eprint={2608.18062},
  archivePrefix={arXiv},
  primaryClass={cs.CL},
  url={https://arxiv.org/abs/2608.18062}
}

@inproceedings{moroni-etal-2024-towards,
  title={Towards a More Comprehensive Evaluation for {I}talian {LLM}s},
  author={Moroni, Luca and Conia, Simone and Martelli, Federico and Navigli, Roberto},
  booktitle={Proceedings of the Tenth Italian Conference on Computational Linguistics (CLiC-it 2024)},
  month={dec},
  year={2024},
  address={Pisa, Italy},
  publisher={CEUR Workshop Proceedings},
  url={https://aclanthology.org/2024.clicit-1.67/},
  pages={584--599}
}

@misc{PersianQA,
  author={Ayoubi, Sajjad and Davoodeh, Mohammad Yasin},
  title={{PersianQA: a dataset for Persian Question Answering}},
  year={2021},
  publisher={GitHub},
  journal={GitHub repository},
  howpublished={\url{https://github.com/SajjjadAyobi/PersianQA}}
}

\clearpage
\appendix

\section{Intrinsic Tokenizer Metrics} \label{app:tokenizer_metrics}

We evaluate tokenizers using a comprehensive suite of intrinsic metrics spanning information-theoretic, linguistic, statistical, and cognitive dimensions. Unless otherwise specified, we use implementations from the \texttt{tokenizer-intrinsic-evals} (TokEval) suite \citep{meister2026tokeval}, which we extended with additional metrics for this study. Metrics are computed on samples from the OSCAR 23.01 corpus \citep{abadji2022towards}, a large-scale multilingual web corpus derived from Common Crawl. Table~\ref{tab:metrics_summary} provides a summary of all metrics computed.

\begin{table*}[t]
\centering
\caption{Summary of intrinsic tokenizer metrics computed in this study. Metrics marked with $\dagger$ were implemented in our fork of \texttt{tokenizer-intrinsic-evals}. Arrows indicate the direction proposed in the literature to correlate with better downstream performance; blank cells indicate metrics without a clear proposed direction.}
\label{tab:metrics_summary}
\small
\begin{tabular}{@{}llcp{5.5cm}@{}}
\toprule
\textbf{Category} & \textbf{Metric} & & \textbf{Description} \\
\midrule
\multirow{6}{*}{Information Theoretic}
 & \texttt{renyi\_efficiency\_a\_*} & $\uparrow$ & Rényi entropy ($\alpha \in \{1, 2, 2.5\}$) \\
 & \texttt{unigram\_entropy} & $\uparrow$ & Shannon entropy of token distribution \\
 & \texttt{compression\_ratio} & $\uparrow$ & Lines per token \\
 & \texttt{bigrams\_entropy} & $\uparrow$ & Entropy over token bigrams \\
 & \texttt{fertility} & $\downarrow$ & Tokens per byte \\
 & \texttt{encoding\_length}$^\dagger$ & $\downarrow$ & Total tokens to encode corpus \\
 & \texttt{avg\_chars\_per\_token} & $\uparrow$ & Average character length per token \\
 & \texttt{avg\_tokens\_per\_line} & $\downarrow$ & Average tokens per line of text \\
\midrule
\multirow{3}{*}{Zipfianness}
 & \texttt{AUC\_zipf}$^\dagger$ & & Area under log-log rank-frequency curve \\
 & \texttt{SLOPE\_zipf}$^\dagger$ & $\approx\!-1$ & Slope of linear fit on log-log \\
 & \texttt{POWER\_LAW\_MAE\_zipf}$^\dagger$ & $\downarrow$ & Deviation from Zipf's law \\
\midrule
\multirow{4}{*}{Morphological}
 & \texttt{morphscore\_precision} & $\uparrow$ & Precision of morpheme boundary detection \\
 & \texttt{morphscore\_recall} & $\uparrow$ & Recall of morpheme boundary detection \\
 & \texttt{morphscore\_micro\_f1} & $\uparrow$ & Micro-averaged F1 score \\
 & \texttt{morphscore\_macro\_f1} & $\uparrow$ & Macro-averaged F1 score \\
\midrule
\multirow{11}{*}{Token Statistics}
 & \texttt{vocab\_size} & & Size of tokenizer vocabulary \\
 & \texttt{vocab\_utilization} & $\uparrow$ & Proportion of vocabulary used \\
 & \texttt{type\_token\_ratio} & & Ratio of unique to total tokens \\
 & \texttt{gini\_coefficient} & & Inequality of tokens-per-line across languages \\
 & \texttt{final\_tokens\_count}$^\dagger$ & & Tokens not used in merges \\
 & \texttt{used\_in\_merges}$^\dagger$ & & Tokens used as merge components \\
 & \texttt{leading\_space\_tokens\_*}$^\dagger$ & & Tokens with leading space marker \\
 & \texttt{average\_token\_rank} & & Mean frequency rank of tokens \\
\midrule
\multirow{3}{*}{Cognitive Plausibility}
 & \texttt{corr\_chunkability\_rt\_mean}$^\dagger$ & $\downarrow$ & Correlation with reaction time \\
 & \texttt{corr\_chunkability\_rt\_z}$^\dagger$ & $\downarrow$ & Correlation with z-scored RT \\
 & \texttt{corr\_chunkability\_accuracy}$^\dagger$ & $\uparrow$ & Correlation with accuracy \\
\midrule
\multirow{4}{*}{Boundary Statistics}
 & \texttt{u\_mean}$^\dagger$ & $\uparrow$ & Mean U-statistic (PMI + entropy) \\
 & \texttt{u\_median}$^\dagger$ & $\uparrow$ & Median U-statistic \\
 & \texttt{u\_mean\_weighted}$^\dagger$ & $\uparrow$ & Frequency-weighted mean U-statistic \\
 & \texttt{u\_mean\_log\_weighted}$^\dagger$ & $\uparrow$ & Log-frequency-weighted mean U-statistic \\
\bottomrule
\end{tabular}
\end{table*}

\subsection{Information Theoretic Metrics}

Information-theoretic metrics have traditionally served as the primary tool for evaluating compression-based hypotheses of tokenizer intrinsic performance.

\paragraph{Rényi Entropy.} 
The Rényi entropy of order $\alpha$ generalizes Shannon entropy and is defined as:
\[
H_\alpha(X) = \frac{1}{1-\alpha} \log_2 \left( \sum_i p_i^\alpha \right)
\]
where $X$ is the token distribution and $p_i$ is the probability of token $i$. For $\alpha = 1$, this reduces to Shannon entropy. Higher-order Rényi entropy (e.g., $\alpha = 2$) penalizes token distributions that are too unbalanced, particularly those with very high-frequency tokens \citep{zouhar-etal-2023-tokenization}. We compute Rényi entropy for $\alpha \in \{1.0, 2.0, 2.5\}$.

\paragraph{Compression Ratio and Fertility.} 
Compression ratio is the number of lines per token, averaged over the corpus samples:
\[
\text{Compression Ratio} = \frac{\text{\# lines}}{\text{\# tokens}}
\]
Higher values indicate better compression. Fertility is the average number of tokens per byte of text. Lower fertility indicates more aggressive compression.

\paragraph{N-gram Entropy.} 
Shannon entropy computed over token bigram (or higher-order n-gram) distributions:
\[
H(X_n) = -\sum_{g \in \mathcal{G}_n} p(g) \log_2 p(g)
\]
where $X_n$ is the n-gram distribution, $\mathcal{G}_n$ is the set of all observed n-grams, and $p(g)$ is the empirical probability of n-gram $g$. Higher values indicate that tokens appear in more diverse contexts, which may correlate with more informative tokenization.

\paragraph{Unigram Distribution Entropy.} 
Shannon entropy of the token frequency distribution, measuring the diversity of the unigram token distribution. Higher values indicate a more uniform usage of the vocabulary.

\paragraph{Encoding Length.} 
The total number of tokens required to encode a corpus. This raw count provides an absolute measure of tokenization efficiency for a fixed text.

\paragraph{Token Length (Average Characters per Token).} 
The average number of characters per token when encoding a given corpus. Longer tokens indicate more aggressive merging during vocabulary construction.

\paragraph{Average Tokens per Line.} 
The mean number of tokens produced per line of text. This provides a coarse measure of tokenization granularity relative to natural text structure.

\subsection{Zipfianness Metrics}

Word frequency distributions in natural language follow Zipf's law \citep{Zipf49}, which states that the frequency of a word is inversely proportional to its rank: $f(r) \propto r^{-\alpha}$ where $r$ is the rank and $\alpha \approx 1$. 
This pattern holds across languages \citep{Piantadosi2014Zipf}. On a log-log plot, a perfect Zipfian distribution appears as a straight line with slope $-1$. 

Recent work has shown that tokenizers whose frequency distributions adhere more closely to Zipf's law tend to yield better downstream model performance, particularly in multilingual settings \citep{lotz2025beyond}. 
The intuition is that tokenizers yielding token distributions closely aligned with a Zipfian power law may be better suited for modeling natural language. 
Following \citet{lotz2025beyond}, we compute three metrics on the log-log rank-frequency curve. Following evidence that power laws only apply above some minimum rank \citep{clauset2009powerlaw}, we restrict our analysis to tokens with $\log(\text{rank}) \leq 6$.

\begin{figure*}[t]
\centering
\begin{subfigure}{0.24\textwidth}
    \centering
    \includegraphics[width=\textwidth]{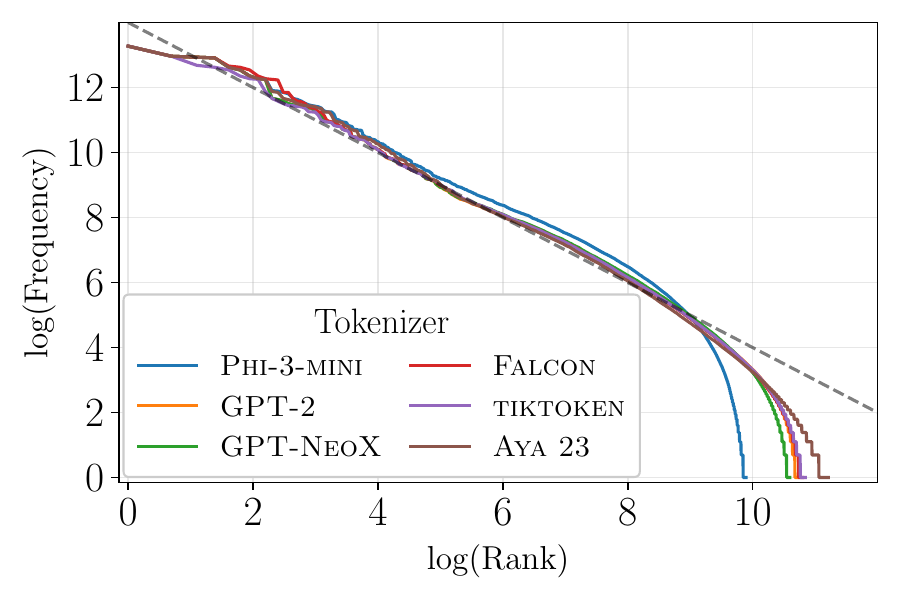}
    \caption{English}
\end{subfigure}
\hfill
\begin{subfigure}{0.24\textwidth}
    \centering
    \includegraphics[width=\textwidth]{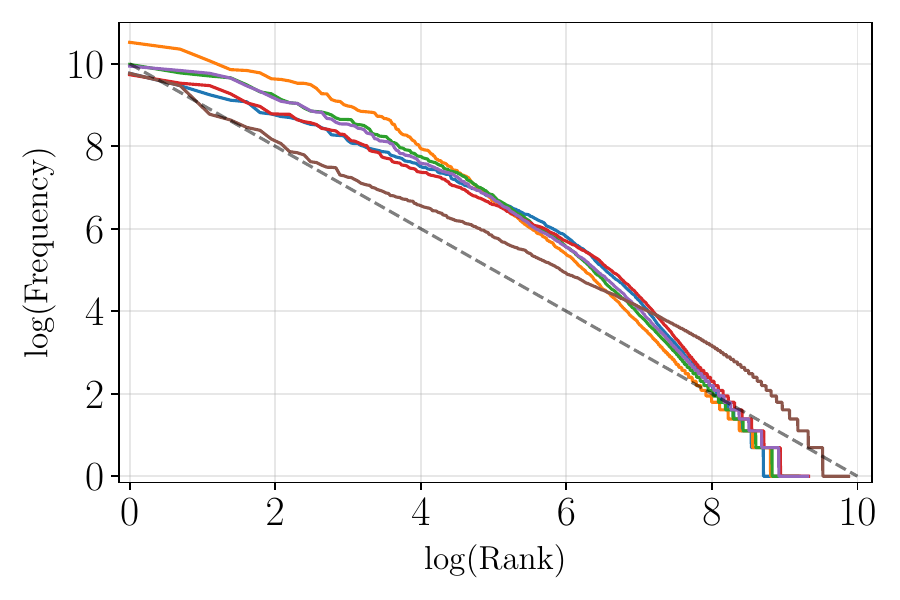}
    \caption{Czech}
\end{subfigure}
\hfill
\begin{subfigure}{0.24\textwidth}
    \centering
    \includegraphics[width=\textwidth]{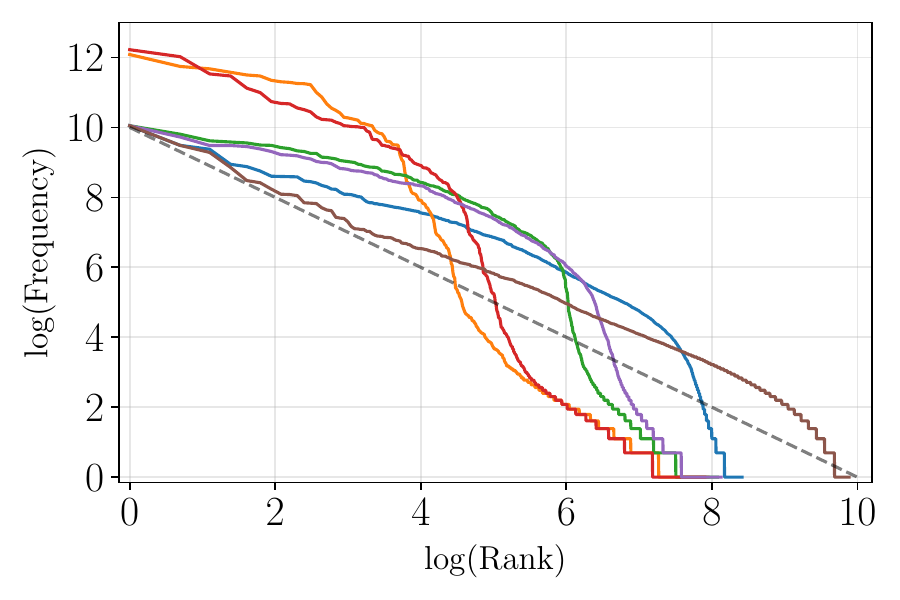}
    \caption{Russian}
\end{subfigure}
\hfill
\begin{subfigure}{0.24\textwidth}
    \centering
    \includegraphics[width=\textwidth]{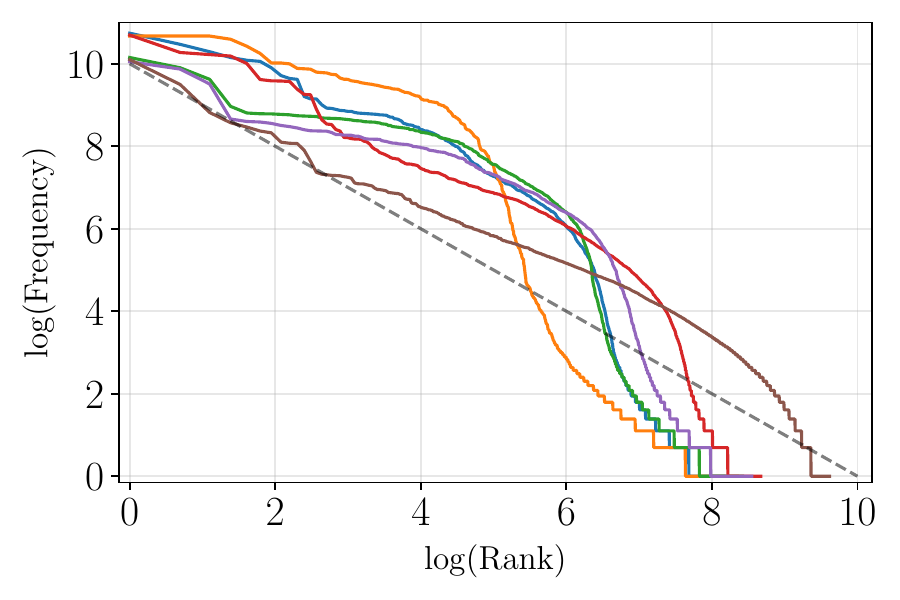}
    \caption{Chinese}
\end{subfigure}
\caption{Token frequency vs.\ frequency rank in log-log scale. Dashed lines (slope $-1$) reference Zipfian distributions. Figure from \citet{lotz2025beyond}.}
\label{fig:zipf_distributions}
\end{figure*}

Figure~\ref{fig:zipf_distributions} illustrates token frequencies against frequency rank on log-log scales for English and three non-English languages. Rank-frequency patterns for English differ minimally across tokenizers, while non-Latin scripts like Russian and Chinese exhibit the largest distributional differences across tokenizers.

\paragraph{Rank-Frequency AUC.} 
The area under the curve of sorted token frequency against frequency rank in log-log scale, computed using Simpson's rule. This captures the overall shape of the rank-frequency distribution, summarizing how token frequencies decay with rank.

\paragraph{Slope (SLOPE).} 
The slope $\beta_1$ from fitting a linear function $f(x) = \beta_0 + \beta_1 x$ on the log-log points (where $x = \log(\text{rank})$ and $y = \log(\text{count})$), approximating Zipf's law. For a perfect Zipfian distribution, the slope would be approximately $-1$. Deviations from $-1$ indicate that the tokenizer's frequency distribution differs from natural language patterns.

\paragraph{Power Law Deviation (POWER\_LAW\_MAE).} 
The mean absolute error from the fitted linear function:
\[
\text{MAE} = \frac{1}{n} \sum_{i=1}^{n} |\beta_0 + \beta_1 x_i - y_i|
\]
This metric quantifies how closely the token distribution aligns with a Zipfian power law. Lower values indicate closer adherence; high MAE suggests the token frequency distribution deviates substantially from a power law, which may indicate suboptimal vocabulary construction.

\subsection{Linguistic and Morphological Metrics}

\paragraph{MorphScore.} 
Assesses the extent to which a tokenizer segments words along morpheme boundaries \citep{arnett-bergen-2025-language, arnett2025morphscore}. If tokens follow morphological boundaries, they are more likely to be linguistically meaningful rather than arbitrary statistical artifacts. We report both precision (proportion of predicted boundaries that are correct) and recall (proportion of true boundaries that are recovered), as well as micro and macro F1 scores across languages.

\subsection{Token Statistics Metrics}

We include metrics based on shallow token statistics that may capture phenomena complementary to information-theoretic measures.

\paragraph{Vocabulary Utilization.} 
The proportion of the vocabulary actually used when encoding a corpus:
\[
\text{Utilization} = \frac{\text{\# unique tokens used}}{\text{vocabulary size}}
\]
Higher utilization suggests more efficient use of the available vocabulary.

\paragraph{Type-Token Ratio (TTR).} 
The ratio of unique tokens (types) to total tokens in the corpus, measuring lexical diversity:
\[
\text{TTR} = \frac{\text{\# unique tokens}}{\text{\# total tokens}}
\]
Higher values indicate more diverse token usage.

\paragraph{Gini Coefficient (Tokenizer Fairness).}
A measure of how equitably a tokenizer distributes encoding costs across languages \citep{ahia-etal-2023-languages,petrov2023language}. For each language $\ell$, the token cost $c_\ell$ is the number of tokens required per line of a content-matched corpus. The metric is the Gini coefficient of these per-language costs:
\[
G = \frac{\sum_{i=1}^{n} \sum_{j=1}^{n} |c_i - c_j|}{2n^2 \mu}
\]
where $n$ is the number of languages and $\mu$ is the mean cost. Values range from 0 (all languages equally costly) upward, with higher values indicating that some languages require disproportionately more tokens than others for equivalent content. The metric is only defined across multiple languages; for the monolingual TIMTC and VISTA runs it is degenerate.

\paragraph{Average Token Rank.} 
The expected rank of a token sampled from a corpus, where tokens are ranked by their frequency:
\[
\text{AR} = \sum_{v \in \mathcal{V}} \text{rank}(v) \cdot p(v)
\]
where $\mathcal{V}$ is the vocabulary, $\text{rank}(v)$ is the position of token $v$ when vocabulary items are sorted by frequency (most frequent = rank 1), and $p(v) = \frac{f(v)}{\sum_{u \in \mathcal{V}} f(u)}$ is the empirical probability of token $v$ in the corpus, with $f(v)$ being the frequency count of token $v$. Lower values indicate that the tokenizer relies more heavily on frequent tokens \citep{limisiewicz-etal-2023-tokenization}.

\subsection{Cognitive Plausibility}

\paragraph{Chunkability Correlation.} 
Following \citet{beinborn-etal-2023-cognitive}, we evaluate tokenizers through the lens of cognitive plausibility by correlating tokenizer behavior with human lexical decision performance. 
We use the British Lexicon Project \citep{keuleers2012british}, which contains lexical decision data (reaction time and accuracy) for 28,730 English words. 
Chunkability measures how aggressively a tokenizer segments a word:
\[
\text{Chunkability}(w) = 1 - \frac{\text{\# tokens for word } w}{\text{\# characters in } w}
\]
Chunkability ranges from 0 (each character is a separate token) to values approaching 1 (the word is kept as a single token). 
Higher chunkability indicates that the tokenizer treats the word as a more cohesive unit. 
We compute Pearson correlations between chunkability and three human performance measures: 
(1) mean reaction time (RT), the average time in milliseconds participants take to decide whether a letter string is a valid word—faster responses indicate easier processing; 
(2) z-scored reaction time, which normalizes RT within each participant to control for individual differences in baseline speed, providing a more reliable measure of relative processing difficulty; and 
(3) accuracy percentage, the proportion of correct responses for each word. 

Cognitively plausible tokenization should show positive correlation with accuracy (words kept intact are recognized more accurately) and negative correlation with reaction time (words kept intact are recognized faster). Prior work has shown that the Unigram algorithm yields less cognitively plausible tokenization behavior than BPE and WordPiece, and also exhibits worse coverage of derivational morphemes \citep{beinborn-etal-2023-cognitive}.

\subsection{Boundary Statistics (U-Statistics)}

The U-statistic evaluates whether a token's boundaries align with natural linguistic boundaries by combining two signals from unsupervised segmentation: \textit{internal cohesion} and \textit{contextual separability} \citep{hu2025entropydriven}. The combined score for a token $w$ is defined in Equation~\ref{eq:ustat} (Section~\ref{sec:ustat_deep_dive}). Here we detail each component.

\paragraph{Internal cohesion: minimum adjacent-character PMI.}
For a token $w$ consisting of characters $c_1, c_2, \ldots, c_n$, we compute the pointwise mutual information for each adjacent character pair:
\[
\text{PMI}(c_i, c_{i+1}) = \log \frac{f(c_i c_{i+1}) \cdot T}{f(c_i) \cdot f(c_{i+1})}
\]
where $f(\cdot)$ denotes corpus frequency and $T$ is the total number of character bigrams in the corpus. High PMI means the two characters co-occur far more often than expected by chance, indicating they are part of the same linguistic unit (e.g., ``t'' and ``h'' in English, or ``qu'' in French).

We take the \textit{minimum} PMI across all adjacent pairs within the token:
\[
\text{PMI}_{\min}(w) = \min_{(c_i, c_{i+1}) \subset w} \text{PMI}(c_i, c_{i+1})
\]
This acts as a bottleneck: a single weak link---a character pair with low PMI---lowers the overall cohesion score, preventing loosely connected parts from being treated as a single unit \citep{hu2025entropydriven}. A token like ``the'' has uniformly high internal PMI, while a poorly segmented token like ``e\_of'' would have a low-PMI pair at the boundary it straddles.

\paragraph{Contextual separability: branching entropy.}
The second component measures whether the token's edges coincide with natural segmentation points. We compute the Shannon entropy over the distribution of characters observed at each boundary:
\begin{align*}
H_{\text{left}}(w) &= -\sum_{l} P(l \mid w) \log P(l \mid w) \\
H_{\text{right}}(w) &= -\sum_{r} P(r \mid w) \log P(r \mid w)
\end{align*}
where $l$ and $r$ index the set of distinct characters that can occur immediately to the left and right of $w$ in the corpus, respectively.

The intuition comes from \citet{harris1955phoneme}: at natural linguistic boundaries (e.g., between words or morphemes), many different characters can appear, leading to high entropy. Mid-morpheme, the set of possible continuations is constrained (e.g., after ``qu'' in English, ``e'', ``i'', ``a'' are likely but few others), yielding low entropy. A token whose boundaries both have high branching entropy is likely to start and end at natural break points. We take the minimum of $H_{\text{left}}$ and $H_{\text{right}}$, so that a single side with low diversity keeps the token from being split prematurely.

\paragraph{Combining the signals and aggregation.}
The U-statistic combines internal cohesion and contextual separability as in Equation~\ref{eq:ustat}. The two terms are complementary: PMI captures whether the token's \textit{interior} is coherent, while branching entropy captures whether the token's \textit{edges} are well-placed. A high U-statistic requires both---the token must be internally cohesive \textit{and} its boundaries must fall at positions with diverse continuations. The weighting $\lambda = 4.0$ balances the two terms, as PMI values span a wider range than branching entropy; we adopt the optimal value found by \citet{hu2025entropydriven} for Chinese word segmentation on the PKU benchmark. We compute $U_{\text{stat}}$ for every token in the tokenizer's vocabulary using character statistics estimated from a corpus, then report the mean (\texttt{u\_mean}) and median (\texttt{u\_median}) across the vocabulary. We also report two frequency-weighted variants designed to dampen the relationship between U-statistic values and vocabulary size, by giving more weight to tokens the model encounters more often. Let $f(w)$ denote the corpus frequency of token $w$ and let $\mathcal{V}$ be the set of vocabulary items with a valid U-statistic. The frequency-weighted mean is:
\[
\texttt{u\_mean\_weighted} = \frac{\sum_{w \in \mathcal{V}} f(w) \cdot U_{\text{stat}}(w)}{\sum_{w \in \mathcal{V}} f(w)}
\]
The log-frequency-weighted mean dampens the influence of extremely frequent tokens:
\[
\texttt{u\_mean\_log\_weighted} = \tfrac{\sum_{w \in \mathcal{V}} \log(1+f(w))\cdot U_{\text{stat}}(w)}{\sum_{w \in \mathcal{V}} \log(1+f(w))}
\]

\section{Model Details} \label{app:model_details}

We use three model suites in this work: TIMTC, TokSuite, and VISTA. 
The TIMTC models from \citet{schmidt-etal-2024-tokenization} are released under the AI Pubs Open RAIL-M license.\footnote{\url{https://www.licenses.ai/ai-pubs-open-railm-vz1}} 
The TokSuite models from \citet{toksuite2025} are released under the MIT license. 
Our VISTA models are released under the Apache~2.0 license.

\subsection{TIMTC} \label{app:timtc_details}

The TIMTC (Tokenization Is More Than Compression) models from \citet{schmidt-etal-2024-tokenization} consist of 350M parameter decoder-only language models trained on the Pile corpus. 
The suite includes 54 models at this scale, plus additional 1.3B and 2.4B parameter models for select configurations. 
Models were trained with three vocabulary sizes (32,768, 40,960, and 49,152 tokens) across 18 different tokenizer configurations. 
Table~\ref{tab:timtc_conditions} summarizes these conditions, which vary along four dimensions:
The models use an MPT (MosaicML Pretrained Transformer; \citealp{MosaicML2023Introducing}) decoder-only architecture and were trained on the Pile corpus, with tokenizer vocabularies constructed over MiniPile, a 6GB subset of English-only data.

\paragraph{Vocabulary Construction (Vocab Constr)} The algorithm used to build the vocabulary. 
BPE iteratively merges the most frequent byte pairs; 
Unigram starts from a large vocabulary and prunes tokens that minimally decrease corpus likelihood; 
WordPiece merges pairs based on pointwise mutual information; 
SaGe incorporates contextual information via a skip-gram objective; 
and PathPiece minimizes the total token count via shortest-path segmentation (with PathPieceL breaking ties by longest token and PathPieceR breaking ties randomly).

\paragraph{Initial Vocabulary (Init Voc)} PathPiece and SaGe require a large starting vocabulary from which tokens are iteratively pruned. 
Options include a vocabulary pre-trained with BPE or Unigram, or the most frequent byte $n$-grams from the corpus. 
This column is empty for methods like BPE and WordPiece, which build vocabularies bottom-up by merging smaller units.

\paragraph{Pre-tokenization (Pre-tok)} 
Rules that restrict token boundaries before vocabulary construction. 
``FirstSpace'' requires spaces to appear only at the start of tokens; 
``SpaceDigit'' treats each space and digit as its own token; 
``FirstSpDigit'' combines both constraints; 
``None'' allows unrestricted token boundaries.

\paragraph{Segmentation (Segment)} The algorithm for splitting text into tokens at inference time. 
``Merge'' applies BPE merge rules; 
``Likelihood'' maximizes unigram probability; 
``Greedy'' takes the longest matching token left-to-right; 
``PathPieceL/R'' finds the minimum-token segmentation.

\begin{table}[ht!]
    \small
    \centering
    \caption{Summary of 350M parameter model vocabulary construction methods, showing initialization, pre-tokenization, and segmentation strategies. All models have been trained with different vocab sizes, specifically 32,768, 40,960 and 49,152 tokens. }
    \label{tab:timtc_conditions}
    \begin{tabular}{llll}
    \toprule
\textbf{Vocab Constr} & \textbf{Init Voc} & \textbf{Pre-tok} & \textbf{Segment} \\
\midrule

\multirow{4}{*}{PathPieceL} & BPE     & FirstSpace   & \multirow{4}{*}{PathPieceL} \\
                            & Unigram & FirstSpace   & \\
                            & $n$-gram & FirstSpDigit & \\
                            & $n$-gram & FirstSpace   & \\
\midrule
\multirow{3}{*}{Unigram}    &         & \multirow{3}{*}{FirstSpace} & Likelihood \\
                            &         &                             & Greedy \\
                            &         &                             & PathPieceL \\
\midrule
\multirow{3}{*}{BPE}        &         & \multirow{3}{*}{FirstSpace} & Merge \\
                            &         &                             & Greedy \\
                            &         &                             & PathPieceL \\
\midrule
WordPiece                   &         & FirstSpace   & Greedy \\
\midrule
\multirow{4}{*}{SaGe}       & BPE     & FirstSpace   & \multirow{4}{*}{Greedy} \\
                            & $n$-gram & FirstSpace  & \\
                            & Unigram & FirstSpace  & \\
                            & $n$-gram & FirstSpDigit & \\
\midrule
\multirow{3}{*}{PathPieceR} & \multirow{3}{*}{$n$-gram} & SpaceDigit   & \multirow{3}{*}{PathPieceR} \\
                            &                           & FirstSpDigit & \\
                            &                           & None         & \\
\bottomrule
    \end{tabular}
\end{table}

\subsection{TokSuite}

The TokSuite models from \citet{toksuite2025} are designed to isolate the impact of tokenizer choice on language model behavior. 
The suite consists of 14 models sharing identical architecture, training data, compute budget, and initialization, differing only in their tokenizer.
Tokenizers are drawn from popular open-source models including GPT-2, BERT, Llama, Qwen, Gemma, Phi, Aya, BLOOM, XGLM, Tekken, GPT-4o, Comma, and TokenMonster, as well as the byte-level ByT5 tokenizer. 
This controlled setup enables direct comparison of tokenizer effects on downstream performance across English and multilingual tasks.

\subsection{VISTA Models} \label{app:vista_models}
\subsubsection{Model Architecture\label{sec:detailed-training-protocol}}
We train two families of LLAMA~1~\cite{touvron2023llama1} language models:
\begin{itemize}
    \item \textbf{VISTA-130M} ($\sim$130M parameters, comparable to GPT-1~\cite{radford2018improving}): 12 layers, hidden size 768, 12 attention heads.
    \item \textbf{VISTA-350M} ($\sim$350M parameters): 24 layers, hidden size 1024, 16 attention heads.
\end{itemize}
Both architectures use SiLU activations, RoPE positional embeddings, a maximum sequence length of 1024, and RMS normalization ($\epsilon{=}10^{-6}$). The embedding and language modeling head dimensions are adjusted to match each vocabulary size; all other hyperparameters are held fixed across tokenizers. Our implementation is built on the codebase of~\citet{sun2025curse} using the Hugging Face Transformers library~\cite{wolf2019huggingface}. Table~\ref{tab:hyperparameters} details the full model and training configurations for the 130M and 350M scales, respectively. Note that the parameter sizes for both families vary significantly with the vocabulary sizes.

\subsubsection{Hyperparameters}
For the 130M models, we closely follow the pre-training protocol from \citet{chang2024goldfish}. All 130M models are trained for 10 epochs on their respective BP-scaled monolingual training data, following evidence that multiple-epoch pre-training is beneficial for smaller datasets~\cite{muennighoff2023scaling}. We use a global batch size of 512 and apply attention dropout (0.1) to mitigate overfitting, with gradient clipping disabled (following ~\cite{chang2024goldfish}).

For the 350M models, we train on substantially larger data: the English models are trained for 1 epoch on the Chinchilla-optimal 7 billion tokens from FineWeb~\cite{penedo2024fineweb}, while non-English models are trained on equivalent bytes of FineWeb-2~\cite{penedo2025fineweb2} data scaled by the language-specific byte premium. The global batch size remains at 512, with gradient clipping enabled, and attention dropout removed. For fair comparison, all configurations within each model scale are held constant across languages and vocabulary sizes.

\subsubsection{Data}

\paragraph{VISTA 130M.}
We utilize monolingual datasets from~\citet{chang2024goldfish}, covering five languages: English (\texttt{eng\_latn}), Thai (\texttt{tha\_thai}), Urdu (\texttt{urd\_arab}), Amharic (\texttt{amh\_ethi}), and Vietnamese (\texttt{vie\_latn}). Training splits are defined by Byte Premium (BP) scaling, such that each language contributes an information-theoretically equivalent amount of data (approximately 1\,GB $\times$ the Byte Premium). The BP multipliers are derived from \citet{arnett2024bit} and detailed in Table~\ref{tab:vista_bp_scaling}. Evaluation splits consist of the disjoint last (BP-scaled) 8{,}000 lines of each monolingual dataset.

\paragraph{VISTA 350M.}
At the 350M scale, we follow a Chinchilla-optimal data regime. English models are trained on 7 billion tokens from FineWeb's 10 Billion token sample~\cite{penedo2024fineweb}, yielding approximately 31\,GiB of tokenized training data (for vocab size 32{,}768). For non-English languages, we source data from FineWeb-2~\cite{penedo2025fineweb2} and define each language's training data target as:
\[
\texttt{target\_bytes} = \text{BP} \times \texttt{eng\_ref\_bytes}
\]
where \texttt{eng\_ref\_bytes} is the byte size of the 7B-token, tokenized English training file, for that corresponding tokenizer type and vocabulary size. Table~\ref{tab:fw2-350m-data} details the data budget for the BPE tokenizer at vocabulary size 32{,}768 (this is representative; the same BP-scaling procedure is applied per tokenizer/vocabulary combination with respect to the corresponding English reference).

Not all languages have sufficient data in FineWeb-2 to meet the BP-scaled target. Amharic has only 2.69\,GiB of raw text---less than 5\% of the required ${\sim}$54\,GiB (tokenized)---and was {excluded from 350M training} to avoid severe overfitting. Urdu has approximately 25\% of the required data; we compensate with multi-epoch training (3--4$\times$ depending on tokenizer/vocabulary), which converged well empirically. Thai and Vietnamese have ample data in FineWeb-2 and are trained for a single epoch.

\subsubsection{Compute Costs}
All VISTA models are trained on 4 or 8 NVIDIA H100/A100 GPUs in \texttt{bfloat16} mixed precision.

\paragraph{VISTA 130M.} Each 130M model is trained for 10 epochs on the BP-scaled datasets (${\sim}$1--2.7\,GB per language). Wall-clock time per run ranges between 2 and 3 hours (${\sim}$8--12 GPU-hours). The full 130M experimental suite comprises 80 runs (5 languages $\times$ 2 tokenizers $\times$ 8 vocabulary sizes), for a total of 640--960 GPU-hours.

\paragraph{VISTA 350M.} Each 350M model is trained for 1 epoch on the Chinchilla-optimal data (31--85\,GiB tokenized per language), except Urdu which trains for 3--4 epochs to compensate for data scarcity. The full 350M suite comprises 20 runs (4 languages $\times$ 2 tokenizers $\times$ 2 vocabulary sizes, plus 2 more vocabulary sizes for English), as detailed in Table~\ref{tab:vista_variations}.

\subsubsection{Language Specific Tokenization}
For both BPE and Unigram variants, tokenizers trained on the corresponding ~\citet{arnett2025crosslingual} monolingual corpus with varying vocabulary sizes from~\citet{arnett2025crosslingual} are used. At the 350M scale, we reuse the same monolingual tokenizers for the FineWeb-2 languages (Thai, Urdu, Vietnamese), as for the 130M experiments (e.g., FineWeb-2 Thai data is tokenized with the same \texttt{tha\_thai} tokenizer). All text data is pre-tokenized into integer IDs before training. Each line in the training file corresponds to a sequence of space-separated token integers. Consecutive non-empty lines from the source text are concatenated and packed to fill the maximum sequence length of 1024 tokens. We prepend a \texttt{CLS} token to each packed sequence and append a \texttt{SEP} token after every individual text segment within that sequence. This packing is performed offline with no padding tokens written to disk, ensuring the model sees a continuous stream of data at high throughput. For the 350M English models, the tokenized data is truncated to get the first 7 billion tokens. For non-English 350M models, the split is defined by the target byte count (\texttt{BP} $\times$ \texttt{eng\_ref\_bytes}), with the evaluation set taken from the disjoint tail of the tokenized corpus.

\begin{table*}
    \centering
    \caption{FineWeb-2 data budget for VISTA 350M models (for the BPE, vocab\,=\,32{,}768 case). The \emph{BP-scaled target} is the Chinchilla-optimal tokenized data size scaled by each language's byte premium. \emph{Tokenized Data Size} is the actual tokenized data after processing downloaded FineWeb-2 text using the pre-trained language specific tokenizers.}
    \label{tab:fw2-350m-data}
    \renewcommand{\arraystretch}{1.15}
    \begin{tabular}{lccccc}
    \toprule
     & \textbf{English} & \textbf{Thai} & \textbf{Urdu} & \textbf{Amharic$^{\dagger}$} & \textbf{Vietnamese} \\
    \midrule
    Byte Premium & 1.00 & 2.74 & 1.71 & 1.72 & 1.35 \\
    \midrule
    English reference (7B tokens) & \multicolumn{5}{c}{31.11\,GiB} \\
    BP-scaled target & 31.11\,GiB & 85.25\,GiB & 53.20\,GiB & 53.52\,GiB & 42.00\,GiB \\
    \midrule
    FineWeb2 Raw Available & -- & 321.88\,GiB & 22.48\,GiB & 2.69\,GiB & 402.69\,GiB \\
    Tokenized Data Size & 31.11\,GiB & $>$85\,GiB & 12.54\,GiB & -- & 38.27\,GiB \\
    Coverage & 100\% & $>$100\% & $\sim$24\% & $<$5\% & $\sim$91\% \\
    Compute-matched Training epochs & 1 & 1 & 3--4${}^{\ddagger}$ & $\sim$20$\times$$^{\dagger}$ & 1 \\
    \bottomrule
    \end{tabular}\\[4pt]
    \begin{minipage}{\linewidth}
    \footnotesize ${}^{\dagger}$Amharic was excluded from 350M training: $<$5\% data coverage would require ${\sim}$20$\times$ epoch repetition, causing severe overfitting. ${}^{\ddagger}$Urdu uses 3--4$\times$ epoch compensation across configs, which converged well empirically.
    \end{minipage}
\end{table*}

\begin{table*}
\centering
\caption{VISTA Pre-training Variations.}
\label{tab:vista_variations}
\renewcommand{\arraystretch}{1.25}
\begin{tabular}{@{}l l l@{}}
\toprule
 & \textbf{VISTA-130M} & \textbf{VISTA-350M} \\
\midrule
Languages
  & English, Thai, Urdu, Amharic, Vietnamese
  & English, Thai, Urdu, Vietnamese \\
Tokenizers
  & BPE, Unigram
  & BPE, Unigram \\
Vocabulary sizes
  & \begin{tabular}[t]{@{}l@{}}8192, 16384, 32768, 49152,\\65536, 81920, 98304, 114688\end{tabular}
  & \begin{tabular}[t]{@{}l@{}}8192, 98304\\(English also 32768, 65536)\end{tabular} \\
\midrule
Total runs
  & $5 \times 2 \times 8 = \mathbf{80}$
  & $4 \times 2 \times 2 + 2 \times 2 = \mathbf{20}$ \\
\bottomrule
\end{tabular}
\end{table*}

\begin{table*}
\centering
\caption{Monolingual corpora and byte-premium (BP) statistics used in our experiments. The left column shows the BP-scaled data size (in GB), and the right column shows the raw UTF-8 byte size. BP multipliers are derived from ~\citet{arnett2024bit}; BP scaling ensures languages with less information-dense scripts are allocated proportionately more raw text data.}
\label{tab:data-bp}
\label{tab:vista_bp_scaling}
\begin{tabular}{lrr}
\toprule
\textbf{Language (tag)} & \textbf{BP-scaled GB} & \textbf{Raw GB} \\
\midrule
English (eng\_latn)     & 1 & 1.00 \\
Thai (tha\_thai)        & 1 & 2.74 \\
Urdu (urd\_arab)        & 1  & 1.71 \\
Amharic (amh\_ethi)     & 1  & 1.72 \\
Vietnamese (vie\_latn)  & 1 & 1.35 \\
\bottomrule
\end{tabular}
\end{table*}

\begin{table*}[t]
\centering
\caption{Model architecture and training hyperparameters for VISTA-130M and VISTA-350M. Where the two scales differ, both values are shown; shared values appear once spanning both columns.}
\label{tab:hyperparameters}
\begin{tabular}{lcc}
\toprule
\textbf{Configuration} & \textbf{VISTA-130M} & \textbf{VISTA-350M} \\
\midrule
\multicolumn{3}{c}{\textit{\textbf{Model Architecture}}} \\
\midrule
Model Type & \multicolumn{2}{c}{LLaMA 1} \\
Hidden Size ($d_{\text{model}}$) & 768 & 1024 \\
Intermediate Size ($d_{\text{ff}}$) & 2048 & 2736 \\
Number of Hidden Layers & 12 & 24 \\
Number of Attention Heads & 12 & 16 \\
Max Sequence Length & \multicolumn{2}{c}{1024} \\
Vocabulary Size & \multicolumn{2}{c}{\textit{Varying}} \\
Activation Function & \multicolumn{2}{c}{SiLU} \\
RMS Norm Epsilon & \multicolumn{2}{c}{$1 \times 10^{-6}$} \\
Initializer Range & \multicolumn{2}{c}{0.02} \\
Positional Embeddings & \multicolumn{2}{c}{RoPE} \\
\midrule
\multicolumn{3}{c}{\textit{\textbf{Training Hyperparameters}}} \\
\midrule
Learning Rate (Max) & \multicolumn{2}{c}{$1 \times 10^{-3}$} \\
Learning Rate (Min) & \multicolumn{2}{c}{$1 \times 10^{-4}$ (min\_lr\_ratio = 0.1)} \\
LR Scheduler & \multicolumn{2}{c}{Cosine with Warmup} \\
Warmup Steps Ratio & \multicolumn{2}{c}{0.10} \\
Attention Dropout & 0.1 & 0.0 (Disabled) \\
Adam $\beta_1, \beta_2$ & \multicolumn{2}{c}{0.9, 0.999} \\
Adam Epsilon & \multicolumn{2}{c}{$1 \times 10^{-6}$} \\
Weight Decay & \multicolumn{2}{c}{0.0} \\
Precision & \multicolumn{2}{c}{bfloat16} \\
Gradient Clipping & 0.0 (Disabled) & 1.0 \\
Epochs & 10 & 1$^{*}$ \\
\midrule
\multicolumn{3}{c}{\textit{\textbf{Batch Configuration}}} \\
\midrule
World Size & \multicolumn{2}{c}{4 GPUs (H100/A100)} \\
Per-GPU Micro-batch Size & 32 & 16 \\
Gradient Accumulation Steps & 4 & 8 \\
Global Batch Size & \multicolumn{2}{c}{512} \\
\midrule
\multicolumn{3}{c}{\textit{\textbf{Data}}} \\
\midrule
Training data source & Goldfish~\cite{chang2024goldfish} & FineWeb/FineWeb-2 \\
Training data size & ${\sim}$1\,GB $\times$ BP & 7B tokens (Eng); BP $\times$ Eng ref bytes \\
\bottomrule
\multicolumn{3}{l}{\footnotesize $^{*}$Urdu: 3--4 epochs to compensate for limited FineWeb-2 data (${\sim}$24\% of BP-scaled target).} \\
\end{tabular}
\end{table*}

\clearpage

\section{Eval Dataset Details}
\label{app:eval-data-details}

We list the benchmarks we used to evaluate each model suite: 
TokSuite (Table \ref{tab:toksuite_eval_list}), 
TIMTC (Table \ref{tab:timtc_evals}), and
VISTA (Table \ref{tab:vista_list}). 
All benchmarks are released under permissive open-source licenses, suitable for this work. FLORES, Global PIQA, ARC, and Belebele are released under CC-BY-SA-4.0 licenses. BLiMP, MultiBLiMP, Winograd\_th, XWinograd, and XCOPA are released under CC-BY-4.0 licenses. HellaSwag and MultiLoko are released under MIT licenses. Global MMLU, AfriMMLU, and INCLUDE are released under Apache-2.0 licenses.

\begin{table*}[t]
\centering
\small
\caption{List of language model evaluations used for TokSuite models. Tasks marked with $\dagger$ were implemented in our fork of the LM evaluation harness.}
\label{tab:toksuite_eval_list}
\begin{tabular}{@{}lll@{}}
\toprule
\textbf{Task Type}                                                                        & \textbf{Benchmark}                    & \textbf{Language(s)}             \\ \midrule
\multirow{5}{*}{\textbf{Linguistic}}                                                      & CoLA \citep{warstadt2019cola}                           & English                          \\
                                                                                          & BLiMP \citep{warstadt2020blimp}                           & English                          \\
                                                                                          & ZhoBLiMP \citep{song2022zhoblimp}                              & Chinese                          \\
                                                                                          & TurBLiMP \citep{yuksekgonul2024turblimp}                              & Turkish                          \\
                                                                                          & MultiBLiMP \citep{jumelet2025multiblimp}                            & English, Italian, Farsi, Turkish \\ \midrule
\multirow{7}{*}{\textbf{NLU}}                                                             & SuperGLUE \citep{wang2019superglue}     & English                          \\
                                                                                          & Winogrande \citep{sakaguchi2020winogrande}     & English, Turkish                 \\
                                                                                          & ANALOGICAL $\dagger$ \citep{wijesiriwardene-etal-2023-analogical}     & English                          \\
                                                                                          & XWinograd \citep{tikhonov2021xwinograd}                             & Chinese                          \\
                                                                                          & CLUE $\dagger$ \citep{xu2020clue}                                  & Chinese                          \\
                                                                                          & FarsTail $\dagger$ \citep{amirkhani2020farstail}                              & Farsi                            \\
                                                                                          & Turkish NLI $\dagger$ \citep{budur2020turkishnli}                  & Turkish                          \\ \midrule
\multirow{6}{*}{\textbf{\begin{tabular}[c]{@{}l@{}}Commonsense\\ Reasoning\end{tabular}}} & HellaSwag \citep{zellers2019hellaswag}                      & English                          \\
                                                                                          & PIQA \citep{bisk2020piqa}                      & English                          \\
                                                                                          & PROST \citep{aroca-ouellette-etal-2021-prost}                    & English                          \\
                                                                                          & XCOPA \citep{ponti-etal-2020-xcopa}              & Chinese, Turkish                 \\
                                                                                          & Global PIQA \citep{chang2025globalpiqa}                           & Italian, Turkish, Chinese        \\
                                                                                          & XStoryCloze \citep{lin-etal-2022-shot}                           & Chinese                          \\ \midrule
\multirow{12}{*}{\textbf{\begin{tabular}[c]{@{}l@{}}World Knowledge/\\ QA\end{tabular}}}  & COMPS $\dagger$ \citep{misra2023comps}                     & English                          \\
                                                                                          & MMLU \citep{hendrycks2021mmlu}                     & English                          \\
                                                                                          & EWoK $\dagger$ \citep{ivanova2024elements}                     & English                          \\
                                                                                          & XCOMPS $\dagger$ \citep{he2025xcompsmultilingualbenchmarkconceptual}                                & Farsi, Turkish, Chinese          \\
                                                                                          & XQuAD \citep{artetxe2020xquad}                                 & English, Turkish, Chinese        \\
                                                                                          & CMMLU \citep{li2023cmmlu}                                 & Chinese                          \\
                                                                                          & Global MMLU \citep{singh2024globalmmlu}                           & English, Italian, Farsi, Turkish, Chinese \\
                                                                                          & Belebele \citep{bandarkar-etal-2024-belebele}                              & English, Italian, Farsi, Turkish, Chinese \\
                                                                                          & INCLUDE \citep{romanou2024include}                   & Chinese, Farsi, Turkish          \\
                                                                                          & MultiLoKo $\dagger$ \citep{hupkes2025multiloko}                   & Farsi, Chinese, Turkish          \\
                                                                                          & PersianQA $\dagger$ \citep{PersianQA}                    & Farsi                            \\
                                                                                          & SynTran-fa $\dagger$ \citep{farsi2024syntranfa}                    & Farsi                            \\ \midrule
\multirow{3}{*}{\textbf{Multi-Task}}                                                      & BBH \citep{suzgun2023bbh}                                   & English                          \\
                                                                                          & ITALIC $\dagger$ \citep{seveso2025italic}                      & Italian                          \\
                                                                                          & ITA-Bench $\dagger$ \citep{moroni-etal-2024-towards}                      & Italian                          \\ \midrule
\multirow{6}{*}{\textbf{Token-Level}}                                                     & CUTE $\dagger$ \citep{edman-etal-2024-cute} & English                          \\
                                                                                          & CharBench $\dagger$ \citep{uzan2026charbench} & English                          \\
                                                                                          & LMentry $\dagger$ \citep{efrat2023lmentry} & English                          \\
                                                                                          & stringbench $\dagger$ \citep{wang2025stringllm} & English                          \\
                                                                                          & EXECUTE $\dagger$ \citep{edman2025execute}                               & English, Chinese                 \\
                                                                                          & TokSuite \citep{toksuite2025}                              & English, Chinese, Farsi, Italian \\ \midrule
\textbf{Perplexity}                                                                       & FLORES $\dagger$ \citep{nllb2022flores200}                                & English, Italian, Farsi, Turkish, Chinese \\ \bottomrule 
\end{tabular}
\end{table*}

\begin{table*}[t]
\centering
\caption{List of English language model evaluations used for ``Tokenization is More Than Compression'' models. Tasks marked with $\dagger$ were implemented in our fork of the LM evaluation harness.}
\label{tab:timtc_evals}
\begin{tabular}{@{}ll@{}}
\toprule
\textbf{Task Type}                                                                        & \textbf{Benchmark}                    \\ \midrule
\multirow{3}{*}{\textbf{Linguistic}}                                                      & SuperGLUE \citep{wang2019superglue}                           \\
                                                                                          & Winogrande \citep{sakaguchi2020winogrande}                           \\
                                                                                          & BLiMP \citep{warstadt2020blimp}                          \\
                                                                                          & MultiBLiMP \citep{jumelet2025multiblimp}                            \\ \midrule
\multirow{5}{*}{\textbf{\begin{tabular}[c]{@{}l@{}}World Knowledge/\\ Semantics\end{tabular}}}  & COMPS $\dagger$ \citep{misra2023comps}                     \\
                                                                                          & ANALOGICAL $\dagger$ \citep{wijesiriwardene-etal-2023-analogical}                     \\
                                                                                          & EWoK $\dagger$ \citep{ivanova2024elements}                     \\
                                                                                          & MMLU \citep{hendrycks2021mmlu}                     \\
                                                                                          & Global MMLU \citep{singh2024globalmmlu}                     \\ \midrule
\multirow{3}{*}{\textbf{\begin{tabular}[c]{@{}l@{}}Commonsense\\ Reasoning\end{tabular}}} & PIQA \citep{bisk2020piqa}                      \\
                                                                                          & HellaSwag \citep{zellers2019hellaswag}                      \\
                                                                                          & PROST \citep{aroca-ouellette-etal-2021-prost}                      \\ \midrule
\multirow{2}{*}{\textbf{QA / Reading}}                                                    & XQuAD \citep{artetxe2020xquad}                                 \\
                                                                                          & Belebele \citep{bandarkar-etal-2024-belebele}                              \\ \midrule
\textbf{Multi-Task}                                                                       & BBH \citep{suzgun2023bbh}                                   \\ \midrule
\multirow{6}{*}{\textbf{Token-Level}}                                                     & CUTE $\dagger$ \citep{edman-etal-2024-cute} \\
                                                                                          & EXECUTE $\dagger$ \citep{edman2025execute}                               \\
                                                                                          & CharBench $\dagger$ \citep{uzan2026charbench} \\
                                                                                          & stringbench $\dagger$ \citep{wang2025stringllm} \\
                                                                                          & LMentry $\dagger$ \citep{efrat2023lmentry} \\
                                                                                          & TokSuite \citep{toksuite2025}                              \\ \midrule
\textbf{Perplexity}                                                                       & FLORES $\dagger$ \citep{nllb2022flores200}                                   \\ \bottomrule 
\end{tabular}
\end{table*}

\clearpage

\begin{table*}[t]
\centering
\small
\caption{List of language model evaluations used for VISTA. Tasks marked with $\dagger$ were implemented in our fork of LM evaluation harness.}
\label{tab:vista_list}
\begin{tabular}{@{}lll@{}}
\toprule
\textbf{Task Type} & \textbf{Benchmark} & \textbf{Language(s)} \\ \midrule
\textbf{Perplexity} & FLORES $\dagger$ \cite{nllb2022flores200} & Amharic, English, Thai, Urdu, Vietnamese \\ \midrule

\multirow{2}{*}{\textbf{Linguistic}} & BLiMP \cite{warstadt2020blimp} & English \\
& MultiBLiMP \cite{jumelet2025multiblimp} & Amharic, English, Urdu \\ \midrule

\multirow{4}{*}{\textbf{NLU}} & SuperGLUE \cite{wang2019superglue} & English \\
& Winograd\_th $\dagger$ & Thai \\
& Winogrande \cite{sakaguchi2020winogrande} & English \\
& XNLI \cite{conneau-etal-2018-xnli} & Vietnamese \\ \midrule

\multirow{6}{*}{\textbf{\begin{tabular}[c]{@{}l@{}}Commonsense\\ Reasoning\end{tabular}}} & HellaSwag \cite{zellers2019hellaswag} & English, Vietnamese \\
& PIQA \cite{bisk2020piqa} & English \\
& PROST \cite{aroca-ouellette-etal-2021-prost} & English \\
& Global PIQA \cite{chang2025globalpiqa} & Amharic, English, Thai, Urdu, Vietnamese \\
& XCOPA \cite{ponti-etal-2020-xcopa} & Thai, Vietnamese \\
& ARC \cite{allenai_arc} & Vietnamese \\ \midrule

\multirow{14}{*}{\textbf{\begin{tabular}[c]{@{}l@{}}World Knowledge/\\ QA\end{tabular}}} & COMPS $\dagger$ \cite{misra2023comps} & English \\
& ANALOGICAL $\dagger$ \cite{wijesiriwardene-etal-2023-analogical} & English \\
& EWoK $\dagger$ \cite{ivanova2024elements} & English \\
& MMLU \cite{hendrycks2021mmlu} & English \\
& XQuAD \cite{artetxe2020xquad} & English, Thai, Vietnamese \\
& Global MMLU \cite{singh2024globalmmlu} & Amharic, English, Vietnamese \\
& MMLU Prox $\dagger$ & Thai, Urdu \\
& AfriMMLU $\dagger$ & Amharic \\
& Belebele \cite{bandarkar-etal-2024-belebele} & Amharic, English, Thai, Urdu, Vietnamese \\
& MultiLoKo $\dagger$ \cite{hupkes2025multiloko} & Thai, Urdu, Vietnamese \\
& INCLUDE \cite{romanou2024include} & Urdu, Vietnamese \\
& Amharic QA $\dagger$ & Amharic \\
& ViQuAD $\dagger$ & Vietnamese \\
& VMLU $\dagger$ & Vietnamese \\ \midrule

\textbf{Multi-Task} & BBH \cite{suzgun2023bbh} & English \\ \midrule

\multirow{6}{*}{\textbf{Token-Level}} & CUTE $\dagger$ \cite{edman-etal-2024-cute} & English \\
& EXECUTE $\dagger$ \cite{edman2025execute} & English \\
& CharBench $\dagger$ \cite{uzan2026charbench} & English \\
& stringbench $\dagger$ \cite{wang2025stringllm} & English \\
& LMentry $\dagger$ \cite{efrat2023lmentry} & English \\
& TokSuite \cite{toksuite2025} & English \\ \bottomrule

\end{tabular}
\end{table*}

\clearpage

\section{Full Eval Results} \label{app:full_eval_results}

\subsection{TokSuite}

The evaluation results of the TokSuite models for each of the languages the models are trained on: English (Table \ref{tab:toksuite_eng_raw}), Chinese (Table \ref{tab:toksuite_zho_raw}), Turkish (Table \ref{tab:toksuite_tur_raw}), Farsi (Table \ref{tab:toksuite_fas_raw}), and Italian (Table \ref{tab:toksuite_ita_raw}).

\begin{figure}[ht!]
    \centering
    \includegraphics[width=0.85\linewidth]{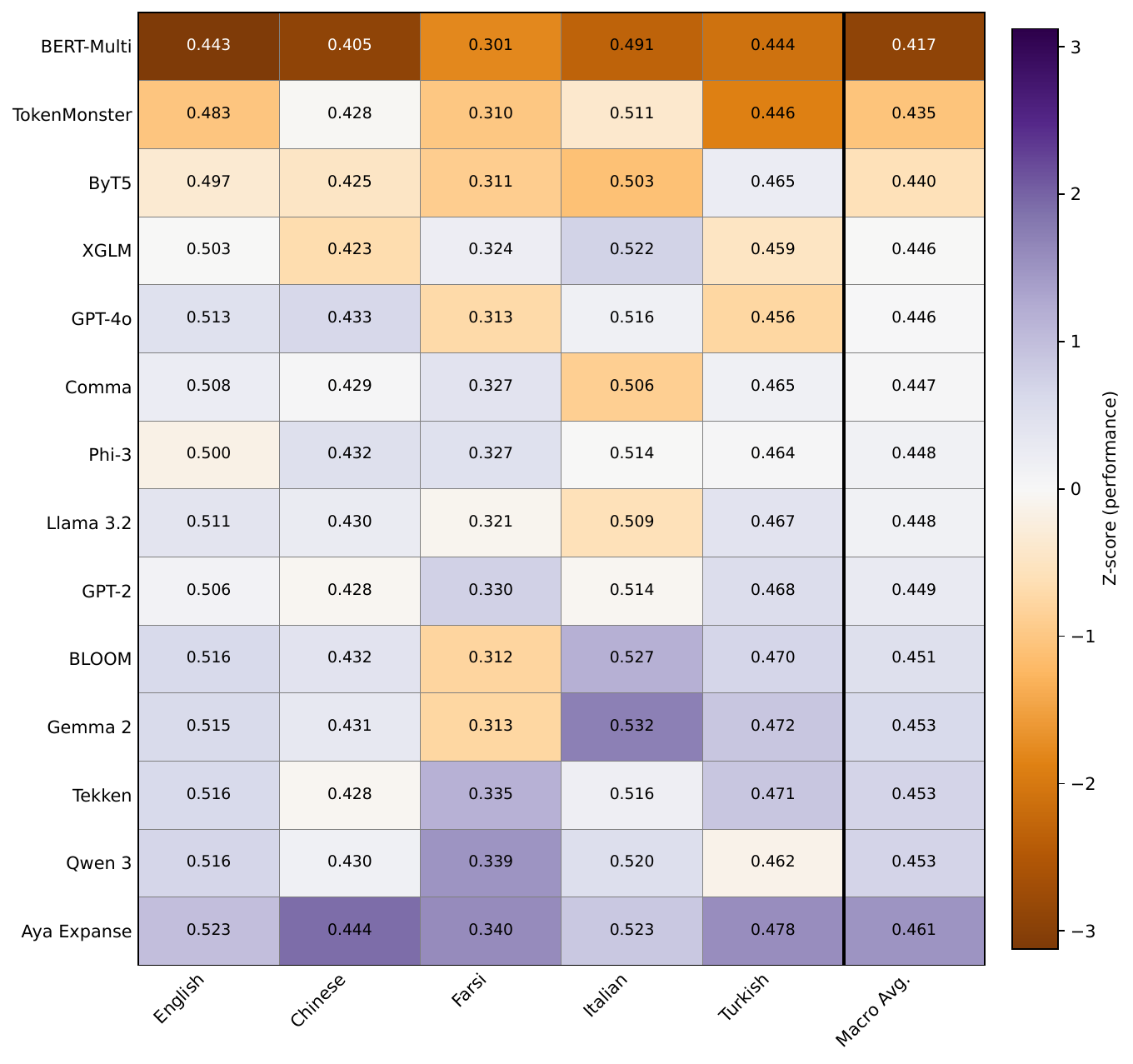}
    \caption{Downstream performance for TokSuite models aggregated by language and macro average across language. Each cell value shows aggregate performance. The color of each cell corresponds to the performance z-scored within the column, where the darker the purple the better the performance.}
    \label{fig:toksuite_by_lang_performance}
\end{figure}

Fig. \ref{fig:toksuite_by_lang_performance} shows the model performance for each of the TokSuite models by language and macro averaged across language.

\begin{table*}[t]
\centering
\caption{Evaluation scores for TokSuite models on English benchmarks. Avg.\ is their mean, the aggregate performance of the text.}
\label{tab:toksuite_eng_raw}
\tiny
\resizebox{\textwidth}{!}{%
\begin{tabular}{@{}lrrrrrrrrrrrrrrr@{}}
\toprule
Benchmark & Aya & Qwen & BLOOM & Comma & XGLM & BERT & ByT5 & Gemma & GPT-2 & Llama & Phi & Tekken & GPT-4o & TokenMonster & Chance \\
\midrule
ANALOGICAL & 0.813 & 0.813 & 0.808 & 0.809 & 0.787 & 0.361 & 0.782 & 0.790 & 0.767 & 0.798 & 0.751 & \textbf{0.823} & 0.792 & 0.697 & 0.250 \\
BBH & 0.232 & 0.224 & 0.236 & 0.229 & 0.200 & 0.202 & 0.209 & 0.212 & \textbf{0.243} & 0.237 & 0.238 & 0.237 & 0.229 & 0.174 & 0.000 \\
Belebele & 0.373 & 0.371 & 0.367 & 0.364 & \textbf{0.393} & 0.256 & 0.362 & 0.378 & 0.359 & 0.388 & 0.357 & 0.392 & 0.373 & 0.329 & 0.250 \\
BLiMP & 0.836 & 0.836 & 0.832 & 0.844 & 0.835 & 0.833 & 0.837 & 0.839 & 0.828 & 0.843 & 0.836 & 0.836 & \textbf{0.845} & 0.787 & 0.500 \\
COMPS & 0.683 & 0.670 & 0.669 & 0.651 & 0.680 & 0.680 & \textbf{0.688} & 0.660 & 0.669 & 0.680 & 0.671 & 0.660 & 0.673 & 0.666 & 0.500 \\
EWoK & 0.626 & 0.623 & 0.621 & 0.607 & 0.614 & 0.565 & 0.617 & 0.628 & 0.612 & \textbf{0.632} & 0.613 & 0.615 & 0.622 & 0.599 & 0.500 \\
Global MMLU & 0.263 & 0.233 & 0.265 & 0.265 & 0.235 & 0.235 & 0.228 & 0.225 & \textbf{0.278} & 0.237 & 0.240 & 0.242 & 0.263 & 0.250 & 0.250 \\
GLUE & 0.487 & 0.484 & 0.454 & 0.473 & 0.385 & 0.289 & 0.410 & \textbf{0.505} & 0.462 & 0.360 & 0.388 & 0.486 & 0.472 & 0.465 & 0.407 \\
HellaSwag & 0.571 & 0.557 & 0.565 & 0.530 & \textbf{0.582} & 0.507 & 0.512 & 0.568 & 0.531 & 0.557 & 0.534 & 0.557 & 0.562 & 0.503 & 0.250 \\
MMLU & 0.234 & \textbf{0.261} & 0.234 & 0.242 & 0.247 & 0.229 & 0.236 & 0.255 & 0.259 & 0.251 & 0.253 & 0.241 & 0.246 & 0.235 & 0.250 \\
MultiBLiMP & 0.990 & 0.992 & 0.991 & 0.984 & 0.988 & 0.992 & 0.990 & 0.988 & 0.988 & 0.990 & \textbf{0.995} & 0.991 & 0.990 & 0.986 & 0.500 \\
PIQA & 0.736 & 0.726 & 0.724 & 0.718 & 0.733 & 0.649 & 0.706 & \textbf{0.738} & 0.709 & 0.726 & 0.709 & 0.727 & 0.729 & 0.698 & 0.500 \\
PROST & 0.316 & \textbf{0.327} & 0.302 & 0.281 & 0.292 & 0.307 & 0.296 & 0.275 & 0.296 & 0.300 & 0.279 & 0.289 & 0.296 & 0.257 & 0.250 \\
Winogrande & 0.577 & 0.571 & 0.588 & 0.556 & 0.576 & 0.534 & 0.552 & 0.563 & 0.540 & \textbf{0.592} & 0.578 & 0.579 & 0.581 & 0.545 & 0.500 \\
XQuAD & \textbf{0.106} & 0.059 & 0.082 & 0.065 & 0.000 & 0.000 & 0.024 & 0.103 & 0.049 & 0.076 & 0.062 & 0.059 & 0.019 & 0.056 & 0.000 \\
\midrule
Avg. & \textbf{0.523} & 0.516 & 0.516 & 0.508 & 0.503 & 0.443 & 0.497 & 0.515 & 0.506 & 0.511 & 0.500 & 0.516 & 0.513 & 0.483 & 0.327 \\
\bottomrule
\end{tabular}
}
\end{table*}

\begin{table*}[t]
\centering
\caption{Evaluation scores for TokSuite models on Chinese benchmarks. Avg.\ is their mean, the aggregate performance of the text.}
\label{tab:toksuite_zho_raw}
\tiny
\resizebox{\textwidth}{!}{%
\begin{tabular}{@{}lrrrrrrrrrrrrrrr@{}}
\toprule
Benchmark & Aya & Qwen & BLOOM & Comma & XGLM & BERT & ByT5 & Gemma & GPT-2 & Llama & Phi & Tekken & GPT-4o & TokenMonster & Chance \\
\midrule
Belebele & 0.340 & 0.321 & 0.311 & \textbf{0.359} & 0.346 & 0.256 & 0.328 & 0.342 & 0.322 & 0.340 & 0.340 & 0.336 & 0.341 & 0.337 & 0.250 \\
CLUE & 0.419 & 0.435 & 0.440 & \textbf{0.441} & 0.438 & 0.345 & 0.436 & 0.434 & 0.434 & 0.435 & 0.435 & 0.432 & 0.435 & 0.435 & 0.389 \\
CMMLU & 0.250 & 0.251 & 0.248 & 0.252 & 0.247 & \textbf{0.253} & 0.252 & 0.251 & 0.248 & 0.249 & 0.251 & 0.249 & 0.248 & 0.249 & 0.250 \\
Global MMLU & \textbf{0.277} & 0.261 & 0.265 & 0.261 & 0.252 & 0.236 & 0.267 & 0.269 & 0.272 & 0.272 & 0.264 & 0.251 & 0.269 & 0.269 & 0.250 \\
Global PIQA & 0.510 & 0.520 & 0.530 & 0.510 & 0.500 & 0.490 & 0.510 & 0.470 & 0.480 & 0.500 & 0.540 & 0.490 & \textbf{0.570} & 0.560 & 0.500 \\
INCLUDE & \textbf{0.294} & 0.250 & 0.244 & 0.244 & 0.261 & 0.246 & 0.248 & 0.257 & 0.264 & 0.242 & 0.228 & 0.255 & 0.239 & 0.240 & 0.250 \\
MultiLoKo & \textbf{0.116} & 0.068 & 0.052 & 0.012 & 0.004 & 0.000 & 0.008 & 0.092 & 0.020 & 0.024 & 0.044 & 0.068 & 0.008 & 0.000 & 0.000 \\
XCOMPS & 0.676 & 0.675 & \textbf{0.701} & 0.671 & 0.666 & 0.699 & 0.686 & 0.634 & 0.687 & 0.666 & 0.690 & 0.674 & 0.668 & 0.687 & 0.500 \\
XCOPA & 0.640 & 0.634 & 0.632 & 0.628 & 0.622 & 0.604 & 0.596 & 0.642 & 0.618 & \textbf{0.652} & 0.634 & 0.644 & 0.650 & 0.620 & 0.500 \\
XQuAD & \textbf{0.059} & 0.007 & 0.034 & 0.004 & 0.000 & 0.000 & 0.052 & 0.013 & 0.035 & 0.018 & 0.010 & 0.015 & 0.009 & 0.003 & 0.000 \\
XStoryCloze & 0.592 & 0.594 & 0.595 & 0.576 & \textbf{0.598} & 0.567 & 0.563 & 0.593 & 0.567 & 0.589 & 0.570 & 0.592 & 0.586 & 0.571 & 0.500 \\
XWinograd & 0.724 & 0.698 & 0.687 & 0.736 & 0.694 & 0.698 & 0.714 & 0.728 & 0.736 & 0.728 & 0.732 & 0.683 & \textbf{0.740} & 0.704 & 0.500 \\
ZhoBLiMP & 0.872 & 0.872 & 0.876 & 0.878 & 0.871 & 0.872 & 0.860 & 0.876 & 0.878 & 0.879 & 0.882 & 0.875 & 0.872 & \textbf{0.891} & 0.500 \\
\midrule
Avg. & \textbf{0.444} & 0.430 & 0.432 & 0.429 & 0.423 & 0.405 & 0.425 & 0.431 & 0.428 & 0.430 & 0.432 & 0.428 & 0.433 & 0.428 & 0.338 \\
\bottomrule
\end{tabular}
}
\end{table*}

\begin{table*}[t]
\centering
\caption{Evaluation scores for TokSuite models on Turkish benchmarks. Avg.\ is their mean, the aggregate performance of the text.}
\label{tab:toksuite_tur_raw}
\tiny
\resizebox{\textwidth}{!}{%
\begin{tabular}{@{}lrrrrrrrrrrrrrrr@{}}
\toprule
Benchmark & Aya & Qwen & BLOOM & Comma & XGLM & BERT & ByT5 & Gemma & GPT-2 & Llama & Phi & Tekken & GPT-4o & TokenMonster & Chance \\
\midrule
Belebele & 0.353 & 0.346 & 0.337 & 0.349 & 0.348 & 0.229 & 0.336 & 0.337 & 0.329 & 0.353 & 0.338 & 0.346 & \textbf{0.363} & 0.312 & 0.250 \\
Global MMLU & 0.274 & 0.274 & 0.274 & 0.270 & 0.276 & 0.241 & 0.262 & \textbf{0.277} & 0.266 & 0.275 & 0.270 & 0.276 & \textbf{0.277} & 0.260 & 0.250 \\
Global PIQA & 0.690 & 0.690 & 0.680 & 0.710 & 0.650 & 0.670 & 0.690 & \textbf{0.740} & 0.700 & 0.670 & 0.690 & 0.710 & 0.670 & 0.660 & 0.500 \\
INCLUDE & 0.266 & 0.228 & 0.248 & 0.250 & 0.232 & \textbf{0.290} & 0.286 & 0.237 & 0.255 & 0.248 & 0.248 & 0.234 & 0.215 & 0.223 & 0.250 \\
MultiBLiMP & 0.958 & 0.950 & 0.958 & 0.963 & \textbf{0.964} & 0.963 & 0.962 & 0.955 & 0.953 & 0.942 & 0.945 & 0.947 & 0.945 & 0.961 & 0.500 \\
MultiLoKo & 0.120 & 0.060 & 0.068 & 0.076 & 0.020 & 0.000 & 0.068 & 0.068 & 0.096 & 0.088 & 0.104 & \textbf{0.136} & 0.024 & 0.008 & 0.000 \\
NLI & 0.342 & 0.318 & \textbf{0.357} & 0.332 & 0.342 & 0.317 & 0.341 & 0.342 & 0.346 & 0.314 & 0.324 & 0.346 & 0.350 & 0.344 & 0.333 \\
TurBLiMP & 0.831 & 0.840 & 0.827 & 0.839 & 0.836 & \textbf{0.844} & 0.831 & 0.835 & 0.841 & 0.835 & 0.834 & 0.834 & 0.828 & 0.803 & 0.500 \\
Winogrande & 0.532 & 0.509 & \textbf{0.539} & 0.490 & 0.517 & 0.496 & 0.519 & 0.504 & 0.509 & 0.503 & 0.511 & 0.500 & 0.497 & 0.498 & 0.500 \\
XCOMPS & 0.661 & 0.661 & 0.653 & 0.652 & \textbf{0.680} & 0.674 & 0.655 & 0.646 & 0.663 & 0.665 & 0.638 & 0.659 & 0.660 & 0.611 & 0.500 \\
XCOPA & 0.632 & 0.642 & 0.604 & 0.620 & 0.638 & 0.600 & 0.616 & 0.630 & 0.628 & \textbf{0.644} & 0.610 & 0.632 & 0.624 & 0.626 & 0.500 \\
XQuAD & 0.073 & 0.025 & \textbf{0.091} & 0.026 & 0.001 & 0.000 & 0.017 & 0.087 & 0.032 & 0.068 & 0.051 & 0.037 & 0.020 & 0.044 & 0.000 \\
\midrule
Avg. & \textbf{0.478} & 0.462 & 0.470 & 0.465 & 0.459 & 0.444 & 0.465 & 0.472 & 0.468 & 0.467 & 0.464 & 0.471 & 0.456 & 0.446 & 0.340 \\
\bottomrule
\end{tabular}
}
\end{table*}

\clearpage

\begin{table*}[t]
\centering
\caption{Evaluation scores for TokSuite models on Farsi benchmarks. Avg.\ is their mean, the aggregate performance of the text.}
\label{tab:toksuite_fas_raw}
\tiny
\resizebox{\textwidth}{!}{%
\begin{tabular}{@{}lrrrrrrrrrrrrrrr@{}}
\toprule
Benchmark & Aya & Qwen & BLOOM & Comma & XGLM & BERT & ByT5 & Gemma & GPT-2 & Llama & Phi & Tekken & GPT-4o & TokenMonster & Chance \\
\midrule
Belebele & 0.340 & 0.352 & 0.324 & \textbf{0.377} & 0.358 & 0.271 & 0.348 & 0.348 & 0.350 & 0.348 & 0.359 & 0.357 & 0.356 & 0.331 & 0.250 \\
FarsTail & 0.338 & 0.336 & 0.332 & 0.346 & 0.350 & 0.332 & 0.336 & 0.326 & 0.332 & 0.339 & 0.325 & \textbf{0.355} & 0.331 & 0.335 & 0.333 \\
Global MMLU & 0.268 & 0.270 & 0.270 & 0.264 & 0.271 & 0.243 & 0.257 & 0.268 & 0.262 & 0.265 & 0.267 & 0.267 & \textbf{0.272} & 0.257 & 0.250 \\
INCLUDE & 0.250 & \textbf{0.279} & 0.214 & 0.241 & 0.243 & 0.221 & 0.234 & 0.254 & 0.265 & 0.232 & 0.245 & 0.243 & 0.250 & 0.243 & 0.250 \\
MultiBLiMP & 0.929 & 0.942 & 0.931 & 0.947 & \textbf{0.974} & 0.969 & 0.927 & 0.872 & 0.945 & 0.897 & 0.942 & 0.909 & 0.901 & 0.946 & 0.500 \\
MultiLoKo & \textbf{0.144} & 0.124 & 0.060 & 0.072 & 0.024 & 0.000 & 0.008 & 0.040 & 0.080 & 0.096 & 0.112 & 0.128 & 0.024 & 0.004 & 0.000 \\
Persian QA & \textbf{0.112} & 0.066 & 0.048 & 0.037 & 0.022 & 0.000 & 0.028 & 0.049 & 0.069 & 0.046 & 0.028 & 0.089 & 0.025 & 0.003 & 0.000 \\
SynTran & \textbf{0.018} & 0.013 & 0.007 & 0.007 & 0.000 & 0.000 & 0.007 & 0.007 & 0.006 & 0.007 & 0.010 & 0.012 & 0.007 & 0.002 & 0.000 \\
XCOMPS & 0.660 & 0.666 & 0.626 & 0.648 & \textbf{0.675} & \textbf{0.675} & 0.659 & 0.654 & 0.661 & 0.656 & 0.656 & 0.655 & 0.655 & 0.669 & 0.500 \\
\midrule
Avg. & \textbf{0.340} & 0.339 & 0.312 & 0.327 & 0.324 & 0.301 & 0.311 & 0.313 & 0.330 & 0.321 & 0.327 & 0.335 & 0.313 & 0.310 & 0.231 \\
\bottomrule
\end{tabular}
}
\end{table*}

\begin{table*}[t]
\centering
\caption{Evaluation scores for TokSuite models on Italian benchmarks. Avg.\ is their mean, the aggregate performance of the text.}
\label{tab:toksuite_ita_raw}
\tiny
\resizebox{\textwidth}{!}{%
\begin{tabular}{@{}lrrrrrrrrrrrrrrr@{}}
\toprule
Benchmark & Aya & Qwen & BLOOM & Comma & XGLM & BERT & ByT5 & Gemma & GPT-2 & Llama & Phi & Tekken & GPT-4o & TokenMonster & Chance \\
\midrule
Belebele & 0.353 & 0.352 & \textbf{0.360} & 0.332 & 0.357 & 0.260 & 0.326 & 0.346 & 0.328 & 0.334 & 0.334 & 0.343 & 0.346 & 0.327 & 0.250 \\
Global MMLU & 0.253 & 0.260 & 0.274 & 0.253 & 0.261 & 0.243 & 0.250 & 0.273 & 0.267 & \textbf{0.280} & 0.253 & 0.271 & 0.264 & 0.246 & 0.250 \\
Global PIQA & 0.740 & 0.720 & 0.740 & 0.680 & 0.720 & 0.730 & 0.690 & \textbf{0.770} & 0.710 & 0.650 & 0.720 & 0.700 & 0.690 & 0.740 & 0.500 \\
ITA-Bench & 0.494 & 0.488 & 0.488 & 0.477 & \textbf{0.501} & 0.434 & 0.467 & 0.496 & 0.479 & 0.490 & 0.479 & 0.488 & 0.492 & 0.468 & 0.350 \\
ITALIC & 0.319 & 0.321 & 0.321 & 0.315 & 0.319 & 0.300 & 0.309 & \textbf{0.327} & 0.321 & 0.322 & 0.323 & 0.317 & 0.323 & 0.310 & 0.250 \\
MultiBLiMP & 0.980 & 0.978 & 0.976 & 0.976 & 0.974 & 0.981 & 0.979 & 0.981 & 0.979 & 0.976 & 0.977 & 0.979 & \textbf{0.983} & 0.973 & 0.500 \\
\midrule
Avg. & 0.523 & 0.520 & 0.527 & 0.506 & 0.522 & 0.491 & 0.503 & \textbf{0.532} & 0.514 & 0.509 & 0.514 & 0.516 & 0.516 & 0.511 & 0.350 \\
\bottomrule
\end{tabular}
}
\end{table*}

\subsubsection{Correlation Between Vocab Size and Performance} 

\begin{figure*}[ht!]
    \centering
    \includegraphics[width=0.95\linewidth]{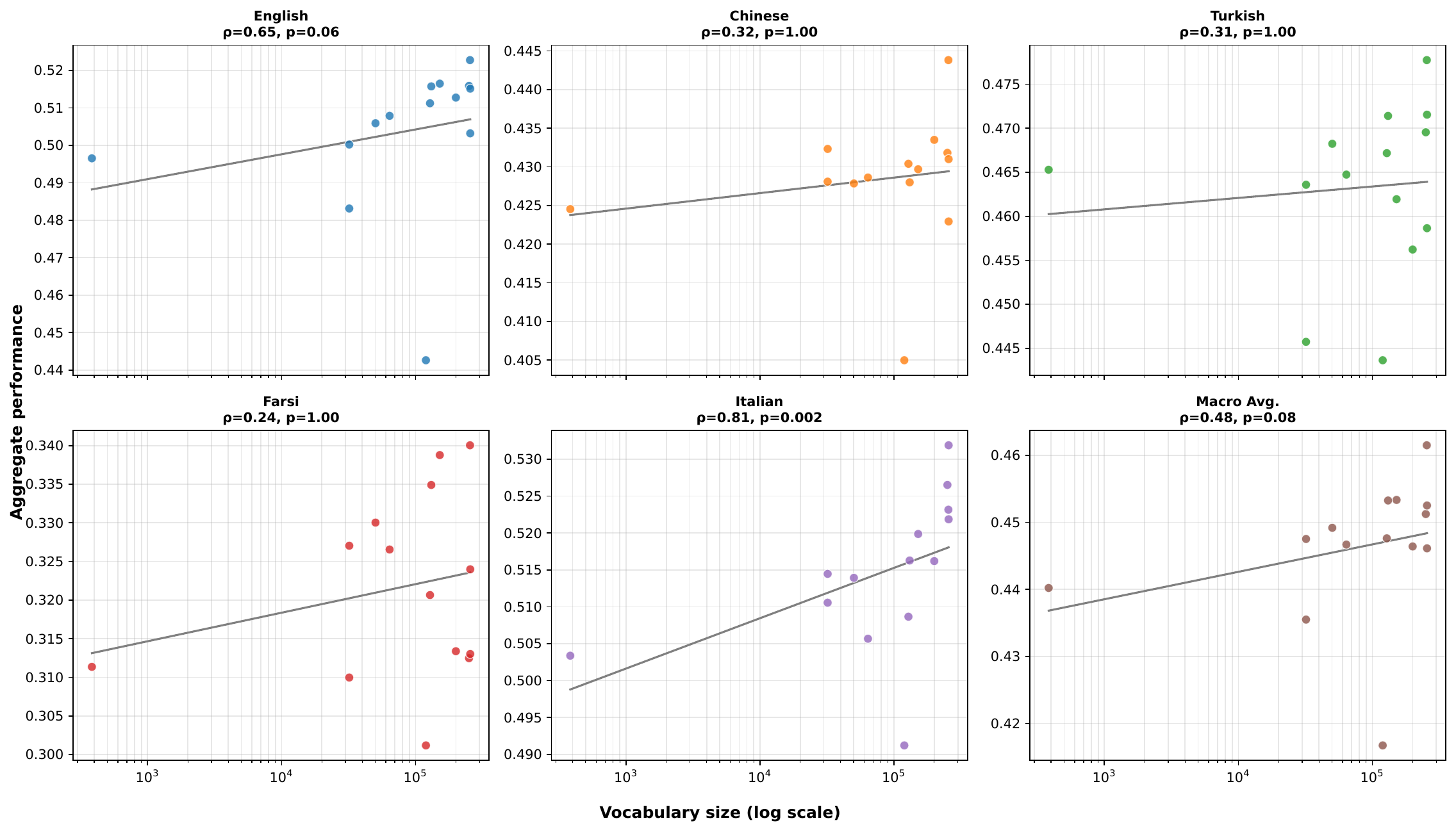}
    \caption{Correlation between vocab size and model performance for the TokSuite models. Facets represent evaluation language, and the last facet the macro average across languages. p-values of the language facets are Bonferroni-corrected for the five languages.}
    \label{fig:toksuite_by_vocab_size_performance}
\end{figure*}

Figure \ref{fig:toksuite_by_vocab_size_performance} plots the relationship between vocabulary size (log-scaled; x-axis) and aggregate performance (mean for each subset, and macro average across language; y-axis) for the TokSuite models. For each facet, Spearman's $\rho$ and p-value are reported.

\clearpage

\subsection{TIMTC}

The evaluation results for the TIMTC models are shown in Tables~\ref{tab:timtc_eng_32k}--\ref{tab:timtc_eng_49k}, one per vocabulary size. Column labels abbreviate the tokenizer configurations described in Table~\ref{tab:timtc_conditions}.

\begin{table*}[t]
\centering
\caption{Evaluation scores for TIMTC models (vocab size 32,768) on English benchmarks. Avg.\ is their mean, the aggregate performance of the text.}
\label{tab:timtc_eng_32k}
\tiny
\resizebox{\textwidth}{!}{%
\begin{tabular}{@{}lrrrrrrrrrrrrrrrr@{}}
\toprule
Benchmark & BPE & PPL-ng-fs & PPL-ng-fsd & PPL-uni-fs & PPR-fsd & PPR-none & PPR-sd & SAGE-bpe & SAGE-ng-fs & SAGE-ng-fsd & SAGE-uni & UNI-gr & UNI-lk & UNI-ppl & WP & Chance \\
\midrule
ANALOGICAL & \textbf{0.648} & 0.442 & 0.428 & 0.428 & 0.491 & 0.487 & 0.628 & 0.629 & 0.563 & 0.581 & 0.582 & 0.582 & 0.574 & 0.385 & 0.608 & 0.250 \\
BBH & 0.203 & 0.206 & 0.200 & 0.209 & \textbf{0.255} & 0.215 & 0.227 & 0.226 & 0.219 & 0.244 & 0.194 & 0.223 & 0.199 & 0.176 & 0.101 & 0.000 \\
Belebele & 0.322 & 0.323 & 0.306 & 0.308 & 0.302 & 0.281 & 0.310 & 0.307 & 0.310 & \textbf{0.324} & 0.316 & 0.308 & 0.322 & 0.301 & 0.318 & 0.250 \\
BLiMP & 0.835 & 0.725 & 0.744 & 0.771 & 0.739 & 0.780 & 0.805 & 0.828 & 0.823 & 0.812 & 0.825 & 0.834 & 0.834 & 0.752 & \textbf{0.839} & 0.500 \\
COMPS & \textbf{0.616} & 0.541 & 0.559 & 0.600 & 0.538 & 0.587 & 0.597 & 0.603 & 0.604 & 0.605 & 0.612 & 0.596 & 0.614 & 0.581 & 0.615 & 0.500 \\
EWoK & 0.578 & 0.524 & 0.524 & 0.553 & 0.529 & 0.543 & 0.563 & \textbf{0.582} & \textbf{0.582} & 0.575 & 0.580 & 0.575 & 0.578 & 0.548 & 0.579 & 0.500 \\
Global MMLU & 0.223 & 0.242 & 0.220 & 0.242 & 0.240 & 0.255 & 0.230 & 0.242 & \textbf{0.258} & 0.235 & 0.245 & 0.235 & 0.240 & 0.233 & 0.235 & 0.250 \\
GLUE & 0.478 & 0.453 & 0.477 & 0.464 & 0.459 & 0.481 & 0.477 & 0.465 & \textbf{0.482} & 0.457 & 0.452 & 0.467 & 0.464 & 0.466 & 0.468 & 0.407 \\
HellaSwag & \textbf{0.386} & 0.314 & 0.319 & 0.329 & 0.333 & 0.349 & 0.371 & 0.374 & 0.372 & 0.374 & 0.375 & 0.379 & 0.380 & 0.332 & 0.377 & 0.250 \\
MMLU & 0.231 & \textbf{0.245} & 0.240 & 0.233 & 0.238 & 0.241 & 0.235 & 0.240 & 0.240 & 0.230 & 0.231 & 0.233 & 0.232 & 0.232 & 0.231 & 0.250 \\
MultiBLiMP & 0.988 & 0.919 & 0.925 & 0.906 & 0.916 & 0.948 & 0.979 & \textbf{0.992} & 0.987 & 0.981 & 0.977 & 0.990 & 0.984 & 0.894 & 0.983 & 0.500 \\
PIQA & 0.658 & 0.596 & 0.587 & 0.613 & 0.587 & 0.615 & 0.650 & \textbf{0.666} & 0.649 & 0.654 & 0.662 & \textbf{0.666} & \textbf{0.666} & 0.606 & 0.658 & 0.500 \\
PROST & 0.298 & 0.299 & 0.304 & 0.286 & 0.268 & 0.284 & 0.296 & 0.286 & 0.321 & 0.317 & 0.279 & 0.285 & \textbf{0.323} & 0.266 & 0.296 & 0.250 \\
Winogrande & 0.511 & 0.515 & 0.509 & 0.504 & 0.533 & 0.487 & 0.514 & 0.522 & 0.519 & \textbf{0.535} & 0.515 & 0.513 & 0.507 & 0.514 & 0.509 & 0.500 \\
XQuAD & 0.012 & 0.018 & \textbf{0.046} & 0.038 & 0.024 & 0.021 & 0.004 & 0.024 & 0.008 & 0.030 & 0.030 & 0.003 & 0.029 & 0.019 & 0.000 & 0.000 \\
\midrule
Avg. & \textbf{0.466} & 0.424 & 0.426 & 0.432 & 0.430 & 0.438 & 0.459 & \textbf{0.466} & 0.462 & 0.464 & 0.458 & 0.459 & 0.463 & 0.420 & 0.454 & 0.327 \\
\bottomrule
\end{tabular}
}
\end{table*}

\begin{table*}[t]
\centering
\caption{Evaluation scores for TIMTC models (vocab size 40,960) on English benchmarks. Avg.\ is their mean, the aggregate performance of the text.}
\label{tab:timtc_eng_41k}
\tiny
\resizebox{\textwidth}{!}{%
\begin{tabular}{@{}lrrrrrrrrrrrrrrrr@{}}
\toprule
Benchmark & BPE & PPL-ng-fs & PPL-ng-fsd & PPL-uni-fs & PPR-fsd & PPR-none & PPR-sd & SAGE-bpe & SAGE-ng-fs & SAGE-ng-fsd & SAGE-uni & UNI-gr & UNI-lk & UNI-ppl & WP & Chance \\
\midrule
ANALOGICAL & 0.654 & 0.456 & 0.468 & 0.423 & 0.538 & 0.455 & 0.505 & 0.650 & 0.569 & 0.502 & 0.589 & 0.566 & 0.596 & 0.410 & \textbf{0.663} & 0.250 \\
BBH & 0.185 & 0.211 & 0.226 & 0.221 & 0.190 & 0.183 & 0.216 & 0.217 & \textbf{0.232} & 0.217 & 0.196 & 0.215 & 0.193 & 0.188 & 0.122 & 0.000 \\
Belebele & 0.309 & 0.298 & 0.311 & 0.296 & 0.304 & 0.259 & 0.312 & 0.301 & 0.331 & 0.293 & \textbf{0.334} & 0.306 & 0.322 & 0.308 & 0.318 & 0.250 \\
BLiMP & \textbf{0.835} & 0.763 & 0.747 & 0.762 & 0.734 & 0.776 & 0.722 & 0.832 & 0.816 & 0.807 & 0.826 & 0.834 & 0.832 & 0.749 & 0.833 & 0.500 \\
COMPS & 0.617 & 0.531 & 0.542 & 0.592 & 0.546 & 0.585 & 0.541 & 0.622 & 0.604 & 0.567 & 0.607 & 0.601 & \textbf{0.627} & 0.579 & 0.615 & 0.500 \\
EWoK & 0.579 & 0.526 & 0.531 & 0.546 & 0.534 & 0.548 & 0.543 & \textbf{0.585} & 0.584 & 0.573 & 0.572 & 0.582 & 0.581 & 0.542 & \textbf{0.585} & 0.500 \\
Global MMLU & 0.242 & 0.235 & 0.250 & 0.228 & 0.225 & 0.200 & 0.240 & 0.228 & \textbf{0.278} & 0.242 & 0.233 & 0.240 & 0.258 & 0.250 & 0.250 & 0.250 \\
GLUE & 0.471 & 0.457 & 0.458 & 0.455 & 0.470 & 0.466 & \textbf{0.482} & 0.462 & 0.475 & 0.465 & 0.460 & 0.454 & 0.469 & 0.457 & 0.460 & 0.407 \\
HellaSwag & 0.385 & 0.314 & 0.320 & 0.322 & 0.338 & 0.349 & 0.336 & 0.378 & 0.372 & 0.359 & 0.376 & 0.380 & 0.378 & 0.328 & \textbf{0.386} & 0.250 \\
MMLU & 0.237 & 0.229 & 0.235 & \textbf{0.247} & 0.228 & 0.242 & 0.230 & 0.229 & \textbf{0.247} & 0.231 & 0.231 & 0.232 & 0.230 & 0.236 & 0.240 & 0.250 \\
MultiBLiMP & 0.979 & 0.930 & 0.930 & 0.869 & 0.918 & 0.958 & 0.900 & \textbf{0.991} & 0.987 & 0.978 & 0.986 & 0.988 & 0.984 & 0.871 & 0.983 & 0.500 \\
PIQA & 0.667 & 0.588 & 0.587 & 0.614 & 0.593 & 0.616 & 0.598 & 0.666 & 0.655 & 0.640 & 0.663 & 0.661 & 0.659 & 0.603 & \textbf{0.677} & 0.500 \\
PROST & \textbf{0.324} & 0.290 & 0.288 & 0.256 & 0.294 & 0.258 & 0.307 & 0.280 & 0.319 & 0.314 & 0.306 & 0.294 & 0.295 & 0.283 & 0.315 & 0.250 \\
Winogrande & \textbf{0.534} & 0.505 & 0.507 & 0.513 & 0.531 & 0.509 & 0.504 & 0.511 & 0.507 & 0.494 & 0.505 & 0.522 & 0.498 & 0.492 & 0.515 & 0.500 \\
XQuAD & 0.030 & 0.007 & \textbf{0.052} & 0.019 & 0.035 & 0.020 & 0.000 & 0.010 & 0.007 & 0.013 & 0.026 & 0.014 & 0.011 & 0.027 & 0.000 & 0.000 \\
\midrule
Avg. & \textbf{0.470} & 0.423 & 0.430 & 0.424 & 0.432 & 0.428 & 0.429 & 0.464 & 0.466 & 0.446 & 0.461 & 0.459 & 0.462 & 0.421 & 0.464 & 0.327 \\
\bottomrule
\end{tabular}
}
\end{table*}

\begin{table*}[t]
\centering
\caption{Evaluation scores for TIMTC models (vocab size 49,152) on English benchmarks. Avg.\ is their mean, the aggregate performance of the text.}
\label{tab:timtc_eng_49k}
\tiny
\resizebox{\textwidth}{!}{%
\begin{tabular}{@{}lrrrrrrrrrrrrrrrr@{}}
\toprule
Benchmark & BPE & PPL-ng-fs & PPL-ng-fsd & PPL-uni-fs & PPR-fsd & PPR-none & PPR-sd & SAGE-bpe & SAGE-ng-fs & SAGE-ng-fsd & SAGE-uni & UNI-gr & UNI-lk & UNI-ppl & WP & Chance \\
\midrule
ANALOGICAL & 0.621 & 0.446 & 0.491 & 0.412 & 0.548 & 0.426 & 0.637 & 0.629 & 0.592 & 0.446 & 0.641 & 0.594 & 0.586 & 0.415 & \textbf{0.652} & 0.250 \\
BBH & 0.243 & 0.221 & 0.208 & 0.193 & 0.225 & 0.210 & 0.223 & 0.242 & 0.240 & \textbf{0.262} & 0.217 & 0.195 & 0.205 & 0.187 & 0.099 & 0.000 \\
Belebele & 0.312 & 0.304 & 0.296 & 0.313 & 0.298 & 0.267 & 0.320 & 0.317 & 0.319 & 0.318 & 0.314 & \textbf{0.327} & 0.323 & 0.302 & 0.303 & 0.250 \\
BLiMP & 0.821 & 0.742 & 0.751 & 0.760 & 0.741 & 0.775 & 0.818 & 0.828 & 0.822 & 0.793 & 0.832 & 0.830 & 0.833 & 0.752 & \textbf{0.837} & 0.500 \\
COMPS & 0.616 & 0.543 & 0.551 & 0.579 & 0.538 & 0.584 & 0.590 & \textbf{0.625} & 0.617 & 0.575 & 0.612 & 0.611 & 0.616 & 0.603 & 0.615 & 0.500 \\
EWoK & 0.582 & 0.525 & 0.524 & 0.548 & 0.533 & 0.548 & 0.569 & 0.576 & 0.583 & 0.557 & 0.588 & \textbf{0.593} & 0.573 & 0.554 & 0.574 & 0.500 \\
Global MMLU & 0.263 & 0.233 & 0.235 & 0.225 & 0.250 & 0.242 & 0.240 & 0.223 & 0.242 & 0.240 & 0.245 & \textbf{0.273} & 0.253 & 0.240 & 0.212 & 0.250 \\
GLUE & \textbf{0.475} & 0.453 & 0.456 & 0.461 & 0.461 & 0.457 & 0.456 & 0.451 & 0.460 & 0.432 & 0.459 & 0.445 & 0.461 & 0.467 & 0.465 & 0.407 \\
HellaSwag & \textbf{0.383} & 0.318 & 0.317 & 0.325 & 0.332 & 0.348 & 0.371 & 0.377 & 0.374 & 0.352 & 0.380 & 0.376 & 0.380 & 0.324 & \textbf{0.383} & 0.250 \\
MMLU & 0.234 & 0.231 & 0.230 & 0.233 & 0.233 & 0.237 & 0.233 & \textbf{0.248} & 0.237 & 0.234 & 0.236 & 0.233 & 0.238 & 0.230 & 0.232 & 0.250 \\
MultiBLiMP & 0.977 & 0.926 & 0.916 & 0.861 & 0.899 & 0.949 & 0.981 & \textbf{0.988} & 0.982 & 0.974 & \textbf{0.988} & 0.982 & 0.984 & 0.869 & 0.984 & 0.500 \\
PIQA & 0.658 & 0.584 & 0.589 & 0.610 & 0.600 & 0.615 & 0.645 & 0.661 & 0.661 & 0.645 & 0.655 & 0.665 & \textbf{0.666} & 0.614 & 0.651 & 0.500 \\
PROST & 0.305 & 0.296 & 0.307 & 0.255 & 0.265 & 0.295 & \textbf{0.322} & 0.276 & 0.302 & 0.286 & 0.302 & 0.268 & 0.310 & 0.294 & 0.317 & 0.250 \\
Winogrande & 0.530 & 0.501 & 0.507 & 0.504 & 0.507 & \textbf{0.535} & 0.524 & 0.506 & 0.501 & 0.514 & 0.512 & 0.513 & 0.505 & 0.507 & 0.507 & 0.500 \\
XQuAD & 0.019 & 0.012 & \textbf{0.064} & 0.026 & 0.035 & 0.010 & 0.000 & 0.018 & 0.014 & 0.027 & 0.013 & 0.025 & 0.022 & 0.020 & 0.000 & 0.000 \\
\midrule
Avg. & \textbf{0.469} & 0.422 & 0.429 & 0.420 & 0.431 & 0.433 & 0.462 & 0.464 & 0.463 & 0.444 & 0.466 & 0.462 & 0.464 & 0.425 & 0.455 & 0.327 \\
\bottomrule
\end{tabular}
}
\end{table*}

\clearpage

\subsection{VISTA}

The results for each of the VISTA models are presented by language: Amharic (Table \ref{tab:vista_amh_raw}), English 130M (Table \ref{tab:vista_eng_raw}), English 350M (Table \ref{tab:vista_eng_350m_raw}), Vietnamese 130M and 350M (Tables \ref{tab:vista_vie_raw} and \ref{tab:vista_vie_350m_raw}), Thai 130M and 350M (Tables \ref{tab:vista_tha_raw} and \ref{tab:vista_tha_350m_raw}), and Urdu 130M and 350M (Tables \ref{tab:vista_urd_raw} and \ref{tab:vista_urd_350m_raw}).



\begin{table*}[t]
\centering
\caption{Evaluation scores for Amharic VISTA 130M models on Amharic benchmarks. Avg.\ is their mean, the aggregate performance of the text.}
\label{tab:vista_amh_raw}
\tiny
\begin{tabular}{@{}llrrrrrrr@{}}
\toprule
Tokenizer Type & Vocab Size & AfriMMLU & AmharicQA & Belebele & Global MMLU & Global PIQA & MultiBLiMP & Avg. \\
\midrule
\multirow{8}{*}{BPE}
 & 8192 & 0.222 & 0.000 & 0.223 & 0.237 & \textbf{0.510} & 0.955 & 0.358 \\
 & 16384 & 0.236 & 0.000 & 0.217 & 0.250 & 0.470 & 0.946 & 0.353 \\
 & 32768 & 0.226 & 0.000 & 0.227 & 0.249 & 0.500 & 0.884 & 0.348 \\
 & 49152 & 0.236 & 0.000 & 0.229 & 0.251 & 0.440 & \textbf{1.000} & 0.359 \\
 & 65536 & 0.232 & 0.000 & 0.224 & \textbf{0.259} & 0.490 & 0.973 & \textbf{0.363} \\
 & 81920 & \textbf{0.250} & 0.000 & 0.232 & 0.250 & 0.470 & 0.964 & 0.361 \\
 & 98304 & 0.220 & 0.000 & 0.220 & 0.256 & 0.460 & 0.955 & 0.352 \\
 & 114688 & 0.244 & 0.000 & 0.226 & 0.240 & 0.420 & 0.964 & 0.349 \\
\midrule
\multirow{8}{*}{Unigram}
 & 8192 & 0.240 & 0.000 & 0.227 & 0.230 & 0.450 & 0.982 & 0.355 \\
 & 16384 & 0.230 & 0.000 & 0.230 & 0.232 & 0.450 & 0.929 & 0.345 \\
 & 32768 & 0.216 & 0.000 & 0.234 & 0.243 & 0.450 & 0.955 & 0.350 \\
 & 49152 & 0.234 & 0.000 & 0.227 & 0.238 & 0.430 & 0.982 & 0.352 \\
 & 65536 & 0.230 & 0.000 & 0.234 & 0.230 & 0.430 & 0.973 & 0.350 \\
 & 81920 & 0.222 & 0.000 & 0.232 & 0.230 & 0.380 & 0.991 & 0.343 \\
 & 98304 & 0.228 & 0.000 & \textbf{0.238} & 0.229 & 0.410 & 0.982 & 0.348 \\
 & 114688 & 0.230 & 0.000 & 0.234 & 0.230 & 0.400 & 0.938 & 0.339 \\
\midrule
Chance & & 0.250 & 0.000 & 0.250 & 0.250 & 0.500 & 0.500 & 0.292 \\
\bottomrule
\end{tabular}
\end{table*}



\begin{table*}[t]
\centering
\caption{Evaluation scores for English VISTA 130M models on English benchmarks. Avg.\ is their mean, the aggregate performance of the text.}
\label{tab:vista_eng_raw}
\tiny
\resizebox{\textwidth}{!}{%
\begin{tabular}{@{}llrrrrrrrrrrrrrrrr@{}}
\toprule
Tokenizer Type & Vocab Size & ANALOGICAL & BBH & Belebele & BLiMP & COMPS & EWoK & Global MMLU & GLUE & HellaSwag & MMLU & MultiBLiMP & PIQA & PROST & Winogrande & XQuAD & Avg. \\
\midrule
\multirow{8}{*}{BPE}
 & 8192 & \textbf{0.246} & 0.004 & 0.229 & 0.586 & 0.512 & 0.510 & \textbf{0.240} & 0.300 & 0.271 & 0.230 & 0.819 & 0.521 & 0.291 & 0.489 & 0.000 & 0.350 \\
 & 16384 & 0.206 & 0.002 & 0.216 & 0.600 & 0.517 & 0.524 & 0.235 & 0.383 & 0.256 & 0.230 & 0.786 & 0.516 & 0.316 & 0.478 & 0.000 & 0.351 \\
 & 32768 & 0.223 & 0.012 & 0.219 & 0.613 & 0.515 & 0.530 & 0.235 & 0.367 & 0.255 & 0.229 & 0.827 & 0.517 & 0.314 & 0.508 & 0.000 & 0.358 \\
 & 49152 & 0.226 & 0.005 & 0.213 & 0.654 & 0.538 & 0.528 & 0.235 & 0.438 & 0.257 & 0.229 & 0.812 & \textbf{0.527} & 0.303 & 0.504 & 0.000 & 0.365 \\
 & 65536 & 0.228 & 0.004 & 0.218 & 0.641 & 0.533 & 0.527 & 0.235 & \textbf{0.457} & 0.253 & 0.229 & 0.839 & 0.509 & 0.315 & 0.504 & 0.000 & \textbf{0.366} \\
 & 81920 & 0.222 & 0.004 & 0.219 & 0.645 & \textbf{0.567} & 0.525 & 0.235 & 0.374 & 0.253 & 0.230 & 0.844 & 0.517 & 0.313 & 0.504 & 0.000 & 0.363 \\
 & 98304 & 0.227 & 0.004 & 0.228 & \textbf{0.661} & 0.535 & 0.525 & 0.235 & 0.378 & 0.258 & 0.230 & \textbf{0.845} & 0.520 & 0.309 & 0.512 & 0.000 & 0.364 \\
 & 114688 & 0.216 & 0.005 & 0.212 & 0.658 & 0.541 & 0.520 & 0.235 & 0.354 & 0.255 & 0.229 & 0.836 & 0.522 & 0.311 & 0.492 & 0.000 & 0.359 \\
\midrule
\multirow{8}{*}{Unigram}
 & 8192 & 0.226 & 0.007 & 0.223 & 0.587 & 0.510 & 0.510 & 0.225 & 0.316 & \textbf{0.272} & \textbf{0.231} & 0.762 & 0.513 & 0.274 & 0.512 & 0.000 & 0.345 \\
 & 16384 & 0.230 & 0.009 & \textbf{0.231} & 0.601 & 0.501 & 0.523 & 0.237 & 0.385 & 0.263 & 0.230 & 0.805 & 0.514 & \textbf{0.319} & 0.487 & 0.000 & 0.356 \\
 & 32768 & 0.210 & 0.012 & 0.216 & 0.617 & 0.511 & \textbf{0.532} & 0.235 & 0.401 & 0.260 & 0.230 & 0.805 & 0.514 & 0.307 & 0.488 & 0.000 & 0.356 \\
 & 49152 & 0.221 & 0.004 & 0.226 & 0.640 & 0.519 & 0.521 & 0.235 & 0.296 & 0.258 & 0.229 & 0.796 & 0.511 & 0.304 & 0.494 & 0.000 & 0.350 \\
 & 65536 & 0.230 & \textbf{0.018} & 0.218 & 0.626 & 0.516 & 0.524 & 0.237 & 0.295 & 0.258 & 0.229 & 0.795 & 0.503 & 0.300 & 0.502 & 0.000 & 0.350 \\
 & 81920 & 0.221 & 0.011 & 0.220 & 0.602 & 0.520 & 0.523 & 0.235 & 0.317 & 0.259 & 0.229 & 0.801 & 0.507 & 0.316 & 0.496 & 0.000 & 0.351 \\
 & 98304 & 0.209 & 0.003 & 0.222 & 0.638 & 0.523 & 0.526 & 0.235 & 0.346 & 0.259 & 0.229 & 0.781 & 0.513 & 0.307 & 0.495 & 0.000 & 0.352 \\
 & 114688 & 0.228 & 0.013 & 0.211 & 0.593 & 0.524 & 0.531 & 0.235 & 0.295 & 0.262 & 0.230 & 0.804 & 0.521 & 0.302 & \textbf{0.539} & 0.000 & 0.353 \\
\midrule
Chance & & 0.250 & 0.000 & 0.250 & 0.500 & 0.500 & 0.500 & 0.250 & 0.407 & 0.250 & 0.250 & 0.500 & 0.500 & 0.250 & 0.500 & 0.000 & 0.327 \\
\bottomrule
\end{tabular}
}
\end{table*}


\begin{table*}[t]
\centering
\caption{Evaluation scores for English VISTA 350M models on English benchmarks. Avg.\ is their mean, the aggregate performance of the text.}
\label{tab:vista_eng_350m_raw}
\tiny
\resizebox{\textwidth}{!}{%
\begin{tabular}{@{}llrrrrrrrrrrrrrrrr@{}}
\toprule
Tokenizer Type & Vocab Size & ANALOGICAL & BBH & Belebele & BLiMP & COMPS & EWoK & Global MMLU & GLUE & HellaSwag & MMLU & MultiBLiMP & PIQA & PROST & Winogrande & XQuAD & Avg. \\
\midrule
\multirow{4}{*}{BPE}
 & 8192 & 0.398 & 0.058 & 0.284 & 0.593 & 0.539 & 0.526 & 0.220 & \textbf{0.468} & 0.318 & \textbf{0.234} & \textbf{0.878} & 0.613 & 0.307 & 0.505 & 0.000 & 0.396 \\
 & 32768 & 0.367 & 0.145 & 0.298 & 0.624 & 0.526 & 0.530 & 0.245 & 0.453 & 0.327 & 0.231 & 0.816 & 0.596 & 0.304 & 0.508 & 0.007 & 0.398 \\
 & 65536 & 0.377 & 0.164 & \textbf{0.321} & 0.578 & 0.551 & 0.534 & \textbf{0.250} & 0.464 & 0.327 & 0.229 & 0.799 & 0.608 & 0.320 & \textbf{0.514} & \textbf{0.008} & 0.403 \\
 & 98304 & \textbf{0.449} & \textbf{0.181} & 0.297 & 0.629 & \textbf{0.574} & 0.540 & 0.235 & 0.466 & \textbf{0.341} & 0.229 & 0.851 & \textbf{0.631} & 0.327 & 0.502 & 0.001 & \textbf{0.417} \\
\midrule
\multirow{4}{*}{Unigram}
 & 8192 & 0.363 & 0.089 & 0.300 & 0.582 & 0.513 & 0.514 & 0.233 & 0.417 & 0.312 & 0.230 & 0.866 & 0.587 & 0.256 & 0.482 & 0.000 & 0.383 \\
 & 32768 & 0.371 & 0.135 & 0.277 & 0.595 & 0.542 & 0.524 & 0.242 & 0.441 & 0.323 & 0.232 & 0.816 & 0.609 & 0.313 & 0.512 & 0.000 & 0.396 \\
 & 65536 & 0.446 & 0.133 & 0.291 & 0.584 & 0.545 & 0.541 & 0.242 & 0.463 & 0.325 & 0.230 & 0.816 & 0.620 & 0.324 & 0.512 & \textbf{0.008} & 0.405 \\
 & 98304 & 0.441 & 0.126 & 0.294 & \textbf{0.633} & 0.569 & \textbf{0.545} & 0.235 & 0.397 & 0.333 & 0.228 & 0.845 & 0.609 & \textbf{0.339} & 0.504 & \textbf{0.008} & 0.407 \\
\midrule
Chance & & 0.250 & 0.000 & 0.250 & 0.500 & 0.500 & 0.500 & 0.250 & 0.407 & 0.250 & 0.250 & 0.500 & 0.500 & 0.250 & 0.500 & 0.000 & 0.327 \\
\bottomrule
\end{tabular}
}
\end{table*}


\begin{table*}[t]
\centering
\caption{Evaluation scores for Vietnamese VISTA 130M models on Vietnamese benchmarks. Avg.\ is their mean, the aggregate performance of the text.}
\label{tab:vista_vie_raw}
\tiny
\resizebox{\textwidth}{!}{%
\begin{tabular}{@{}llrrrrrrrrrrrrr@{}}
\toprule
Tokenizer Type & Vocab Size & ARC & Belebele & Global MMLU & Global PIQA & HellaSwag & INCLUDE & MultiLoKo & ViQuAD & VMLU & XCOPA & XNLI & XQuAD & Avg. \\
\midrule
\multirow{8}{*}{BPE}
 & 8192 & 0.221 & \textbf{0.270} & 0.245 & 0.500 & \textbf{0.266} & 0.235 & 0.000 & 0.006 & 0.249 & 0.544 & 0.339 & 0.000 & 0.239 \\
 & 16384 & 0.221 & 0.238 & 0.249 & 0.580 & 0.260 & 0.255 & 0.000 & 0.015 & 0.272 & 0.542 & 0.341 & 0.000 & 0.248 \\
 & 32768 & 0.227 & 0.236 & 0.245 & 0.570 & 0.261 & \textbf{0.273} & 0.000 & 0.032 & 0.276 & 0.530 & 0.353 & 0.000 & 0.250 \\
 & 49152 & 0.233 & 0.230 & 0.250 & 0.590 & 0.259 & 0.262 & 0.000 & 0.020 & 0.265 & 0.530 & 0.343 & 0.000 & 0.249 \\
 & 65536 & 0.238 & 0.238 & 0.242 & 0.540 & 0.259 & 0.249 & 0.000 & 0.001 & \textbf{0.289} & 0.532 & 0.335 & 0.000 & 0.244 \\
 & 81920 & 0.226 & 0.233 & 0.242 & 0.590 & 0.258 & 0.260 & 0.000 & 0.057 & 0.265 & 0.508 & 0.330 & 0.000 & 0.247 \\
 & 98304 & 0.234 & 0.241 & 0.238 & 0.570 & 0.259 & 0.267 & 0.000 & 0.039 & 0.251 & 0.532 & 0.340 & 0.000 & 0.248 \\
 & 114688 & 0.240 & 0.234 & 0.240 & \textbf{0.600} & 0.259 & 0.265 & 0.000 & 0.060 & 0.282 & 0.530 & 0.341 & 0.000 & 0.254 \\
\midrule
\multirow{8}{*}{Unigram}
 & 8192 & \textbf{0.243} & 0.266 & 0.241 & 0.510 & 0.265 & 0.256 & 0.000 & 0.024 & 0.259 & 0.524 & 0.333 & 0.000 & 0.243 \\
 & 16384 & 0.232 & 0.226 & 0.249 & 0.500 & 0.262 & 0.238 & 0.000 & \textbf{0.196} & 0.274 & 0.538 & \textbf{0.360} & 0.000 & \textbf{0.256} \\
 & 32768 & 0.235 & 0.238 & 0.249 & 0.560 & 0.259 & 0.253 & 0.000 & 0.037 & 0.253 & 0.538 & \textbf{0.360} & 0.000 & 0.248 \\
 & 49152 & 0.226 & 0.231 & 0.252 & 0.550 & 0.259 & 0.231 & 0.000 & 0.041 & 0.263 & \textbf{0.552} & 0.357 & 0.000 & 0.247 \\
 & 65536 & 0.239 & 0.242 & 0.243 & 0.550 & 0.258 & 0.233 & 0.000 & 0.036 & 0.247 & 0.530 & 0.330 & 0.000 & 0.242 \\
 & 81920 & 0.237 & 0.236 & \textbf{0.254} & 0.580 & 0.262 & 0.227 & 0.000 & 0.046 & 0.272 & 0.550 & 0.349 & 0.000 & 0.251 \\
 & 98304 & 0.238 & 0.232 & 0.252 & 0.550 & 0.262 & 0.220 & 0.000 & 0.030 & 0.265 & 0.544 & 0.339 & 0.000 & 0.244 \\
 & 114688 & 0.226 & 0.232 & \textbf{0.254} & 0.530 & 0.259 & 0.231 & 0.000 & 0.084 & 0.272 & 0.538 & 0.347 & 0.000 & 0.248 \\
\midrule
Chance & & 0.250 & 0.250 & 0.250 & 0.500 & 0.250 & 0.250 & 0.000 & 0.000 & 0.250 & 0.500 & 0.333 & 0.000 & 0.236 \\
\bottomrule
\end{tabular}
}
\end{table*}

\begin{table*}[t]
\centering
\caption{Evaluation scores for Vietnamese VISTA 350M models on Vietnamese benchmarks. Avg.\ is their mean, the aggregate performance of the text.}
\label{tab:vista_vie_350m_raw}
\tiny
\resizebox{\textwidth}{!}{%
\begin{tabular}{@{}llrrrrrrrrrrrrr@{}}
\toprule
Tokenizer Type & Vocab Size & ARC & Belebele & Global MMLU & Global PIQA & HellaSwag & INCLUDE & MultiLoKo & ViQuAD & VMLU & XCOPA & XNLI & XQuAD & Avg. \\
\midrule
\multirow{2}{*}{BPE}
 & 8192 & 0.239 & 0.241 & \textbf{0.247} & 0.460 & 0.256 & 0.251 & 0.000 & 0.000 & 0.228 & 0.530 & 0.333 & 0.000 & 0.232 \\
 & 98304 & 0.237 & 0.292 & 0.230 & 0.670 & \textbf{0.298} & 0.275 & 0.000 & 0.006 & \textbf{0.251} & 0.536 & 0.382 & \textbf{0.001} & 0.265 \\
\midrule
\multirow{2}{*}{Unigram}
 & 8192 & 0.232 & \textbf{0.294} & 0.229 & 0.660 & \textbf{0.298} & \textbf{0.276} & 0.000 & 0.011 & 0.235 & \textbf{0.556} & 0.384 & 0.000 & 0.265 \\
 & 98304 & \textbf{0.243} & 0.291 & 0.230 & \textbf{0.680} & \textbf{0.298} & 0.275 & 0.000 & \textbf{0.019} & 0.245 & 0.550 & \textbf{0.406} & \textbf{0.001} & \textbf{0.270} \\
\midrule
Chance & & 0.250 & 0.250 & 0.250 & 0.500 & 0.250 & 0.250 & 0.000 & 0.000 & 0.250 & 0.500 & 0.333 & 0.000 & 0.236 \\
\bottomrule
\end{tabular}
}
\end{table*}


\begin{table*}[t]
\centering
\caption{Evaluation scores for Thai VISTA 130M models on Thai benchmarks. Avg.\ is their mean, the aggregate performance of the text.}
\label{tab:vista_tha_raw}
\tiny
\begin{tabular}{@{}llrrrrrrrr@{}}
\toprule
Tokenizer Type & Vocab Size & Belebele & Global PIQA & MMLU ProX & MultiLoKo & Winograd & XCOPA & XQuAD & Avg. \\
\midrule
\multirow{8}{*}{BPE}
 & 8192 & 0.257 & \textbf{0.540} & 0.081 & 0.000 & 0.481 & 0.540 & 0.000 & 0.271 \\
 & 16384 & 0.240 & \textbf{0.540} & 0.084 & 0.000 & \textbf{0.540} & 0.560 & 0.000 & \textbf{0.281} \\
 & 32768 & 0.234 & \textbf{0.540} & 0.080 & 0.000 & 0.526 & 0.544 & 0.000 & 0.275 \\
 & 49152 & 0.221 & \textbf{0.540} & 0.084 & \textbf{0.004} & 0.526 & 0.538 & 0.000 & 0.273 \\
 & 65536 & 0.231 & 0.510 & 0.081 & 0.000 & 0.526 & 0.562 & 0.000 & 0.273 \\
 & 81920 & 0.239 & 0.510 & 0.083 & 0.000 & 0.512 & 0.550 & 0.000 & 0.271 \\
 & 98304 & 0.239 & 0.510 & 0.082 & 0.000 & 0.505 & 0.550 & 0.000 & 0.269 \\
 & 114688 & 0.241 & 0.500 & 0.084 & 0.000 & 0.516 & 0.540 & 0.000 & 0.269 \\
\midrule
\multirow{8}{*}{Unigram}
 & 8192 & \textbf{0.269} & 0.480 & 0.081 & 0.000 & 0.470 & 0.566 & 0.000 & 0.267 \\
 & 16384 & 0.227 & 0.510 & 0.085 & 0.000 & 0.488 & 0.526 & 0.000 & 0.262 \\
 & 32768 & 0.232 & 0.510 & 0.083 & 0.000 & 0.523 & 0.542 & 0.000 & 0.270 \\
 & 49152 & 0.242 & \textbf{0.540} & \textbf{0.086} & 0.000 & 0.516 & 0.556 & 0.000 & 0.277 \\
 & 65536 & 0.230 & 0.500 & 0.084 & 0.000 & 0.463 & 0.564 & 0.000 & 0.263 \\
 & 81920 & 0.223 & 0.530 & 0.082 & 0.000 & 0.530 & 0.564 & 0.000 & 0.276 \\
 & 98304 & 0.219 & 0.490 & 0.084 & 0.000 & 0.519 & \textbf{0.576} & 0.000 & 0.270 \\
 & 114688 & 0.229 & 0.520 & 0.083 & 0.000 & 0.502 & 0.570 & 0.000 & 0.272 \\
\midrule
Chance & & 0.250 & 0.500 & 0.100 & 0.000 & 0.500 & 0.500 & 0.000 & 0.264 \\
\bottomrule
\end{tabular}
\end{table*}

\begin{table*}[t]
\centering
\caption{Evaluation scores for Thai VISTA 350M models on Thai benchmarks. Avg.\ is their mean, the aggregate performance of the text.}
\label{tab:vista_tha_350m_raw}
\tiny
\begin{tabular}{@{}llrrrrrrrr@{}}
\toprule
Tokenizer Type & Vocab Size & Belebele & Global PIQA & MMLU ProX & MultiLoKo & Winograd & XCOPA & XQuAD & Avg. \\
\midrule
\multirow{2}{*}{BPE}
 & 8192 & 0.314 & \textbf{0.670} & \textbf{0.083} & 0.000 & 0.505 & 0.540 & 0.000 & 0.302 \\
 & 98304 & 0.278 & 0.590 & 0.082 & 0.000 & \textbf{0.565} & 0.550 & 0.000 & 0.295 \\
\midrule
\multirow{2}{*}{Unigram}
 & 8192 & \textbf{0.327} & 0.610 & 0.079 & 0.000 & 0.540 & \textbf{0.576} & 0.000 & \textbf{0.305} \\
 & 98304 & 0.296 & 0.510 & \textbf{0.083} & 0.000 & 0.516 & 0.544 & 0.000 & 0.278 \\
\midrule
Chance & & 0.250 & 0.500 & 0.100 & 0.000 & 0.500 & 0.500 & 0.000 & 0.264 \\
\bottomrule
\end{tabular}
\end{table*}


\begin{table*}[t]
\centering
\caption{Evaluation scores for Urdu VISTA 130M models on Urdu benchmarks. Avg.\ is their mean, the aggregate performance of the text.}
\label{tab:vista_urd_raw}
\tiny
\begin{tabular}{@{}llrrrrrrr@{}}
\toprule
Tokenizer Type & Vocab Size & Belebele & Global PIQA & INCLUDE & MMLU ProX & MultiBLiMP & MultiLoKo & Avg. \\
\midrule
\multirow{8}{*}{BPE}
 & 8192 & 0.226 & 0.580 & 0.236 & 0.085 & 0.931 & 0.000 & 0.343 \\
 & 16384 & 0.233 & 0.610 & 0.239 & 0.083 & \textbf{0.942} & 0.000 & 0.351 \\
 & 32768 & 0.240 & 0.610 & 0.196 & 0.086 & 0.911 & 0.000 & 0.340 \\
 & 49152 & 0.249 & 0.610 & 0.253 & 0.087 & 0.913 & 0.000 & 0.352 \\
 & 65536 & 0.246 & \textbf{0.620} & 0.179 & 0.086 & 0.920 & 0.000 & 0.342 \\
 & 81920 & 0.242 & 0.580 & 0.179 & 0.087 & 0.927 & 0.000 & 0.336 \\
 & 98304 & 0.244 & \textbf{0.620} & 0.205 & 0.087 & 0.918 & 0.000 & 0.346 \\
 & 114688 & 0.252 & 0.580 & 0.253 & 0.086 & 0.922 & 0.000 & 0.349 \\
\midrule
\multirow{8}{*}{Unigram}
 & 8192 & 0.238 & 0.570 & 0.338 & \textbf{0.089} & 0.915 & 0.000 & 0.358 \\
 & 16384 & \textbf{0.257} & 0.570 & \textbf{0.344} & 0.088 & 0.925 & 0.000 & \textbf{0.364} \\
 & 32768 & 0.239 & 0.580 & 0.239 & 0.087 & 0.918 & 0.000 & 0.344 \\
 & 49152 & 0.237 & 0.540 & 0.230 & 0.086 & 0.891 & 0.000 & 0.331 \\
 & 65536 & 0.240 & 0.540 & 0.318 & 0.088 & 0.895 & 0.000 & 0.347 \\
 & 81920 & 0.240 & 0.510 & 0.273 & 0.086 & 0.907 & 0.000 & 0.336 \\
 & 98304 & 0.234 & 0.520 & 0.324 & 0.087 & 0.913 & 0.000 & 0.346 \\
 & 114688 & 0.246 & 0.560 & 0.338 & 0.088 & 0.918 & 0.000 & 0.358 \\
\midrule
Chance & & 0.250 & 0.500 & 0.250 & 0.100 & 0.500 & 0.000 & 0.267 \\
\bottomrule
\end{tabular}
\end{table*}

\begin{table*}[t]
\centering
\caption{Evaluation scores for Urdu VISTA 350M models on Urdu benchmarks. Avg.\ is their mean, the aggregate performance of the text.}
\label{tab:vista_urd_350m_raw}
\tiny
\begin{tabular}{@{}llrrrrrrr@{}}
\toprule
Tokenizer Type & Vocab Size & Belebele & Global PIQA & INCLUDE & MMLU ProX & MultiBLiMP & MultiLoKo & Avg. \\
\midrule
\multirow{2}{*}{BPE}
 & 8192 & 0.251 & 0.680 & 0.290 & \textbf{0.084} & \textbf{0.951} & 0.000 & 0.376 \\
 & 98304 & 0.276 & 0.740 & 0.324 & \textbf{0.084} & 0.947 & 0.000 & \textbf{0.395} \\
\midrule
\multirow{2}{*}{Unigram}
 & 8192 & \textbf{0.278} & 0.680 & 0.330 & 0.083 & 0.909 & 0.000 & 0.380 \\
 & 98304 & 0.260 & \textbf{0.750} & \textbf{0.332} & 0.083 & 0.935 & 0.000 & 0.393 \\
\midrule
Chance & & 0.250 & 0.500 & 0.250 & 0.100 & 0.500 & 0.000 & 0.267 \\
\bottomrule
\end{tabular}
\end{table*}

\clearpage
\subsubsection{Correlation Between Vocab Size and Performance} \label{app:vista_vocab_size_perf_corr}

\begin{figure*}[hb!]
    \centering
    \includegraphics[width=0.9\linewidth]{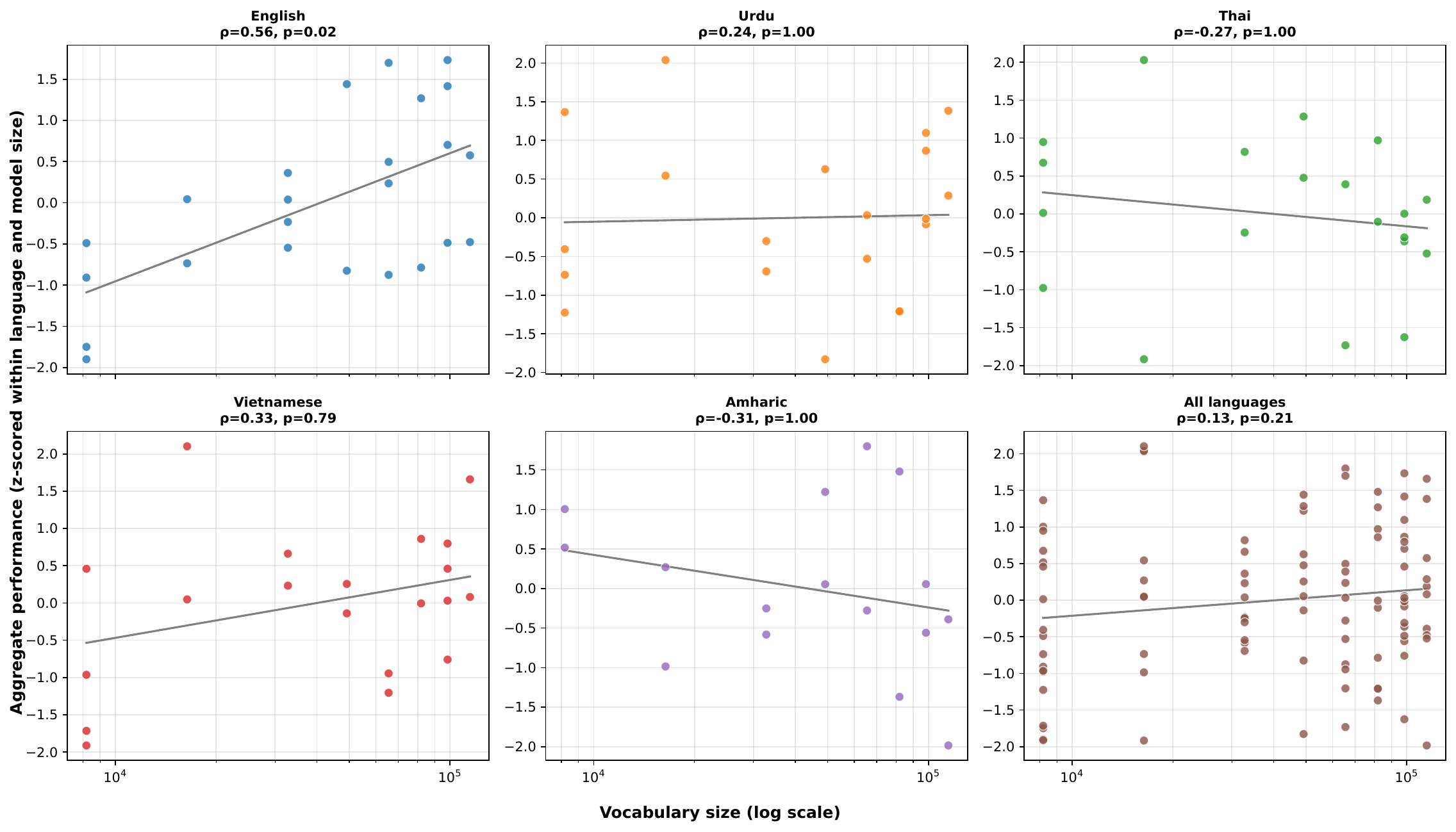}
    \caption{Correlation between vocabulary size (x-axis) and aggregated (z-scored) model performance by language and pooled over languages (bottom right facet). Each facet also reports correlation coefficient and p-value; p-values of the language facets are Bonferroni-corrected for the five languages.}
    \label{fig:vista_by_language_vocab_correlation}
\end{figure*}

Figure \ref{fig:vista_by_language_vocab_correlation} plots the relationship between vocabulary size (log-scaled; x-axis) and aggregate performance (z-scored within each language and model size, and pooled across languages; y-axis) for the VISTA models. For each facet, Spearman's $\rho$ and p-value are reported.

\clearpage

\section{Full Metric-Performance Correlation Results} \label{app:full_corr_results}

\begin{figure*}[t]
    \centering

    \begin{subfigure}{0.45\textwidth}
        \centering
        \includegraphics[height=5.5cm, keepaspectratio]{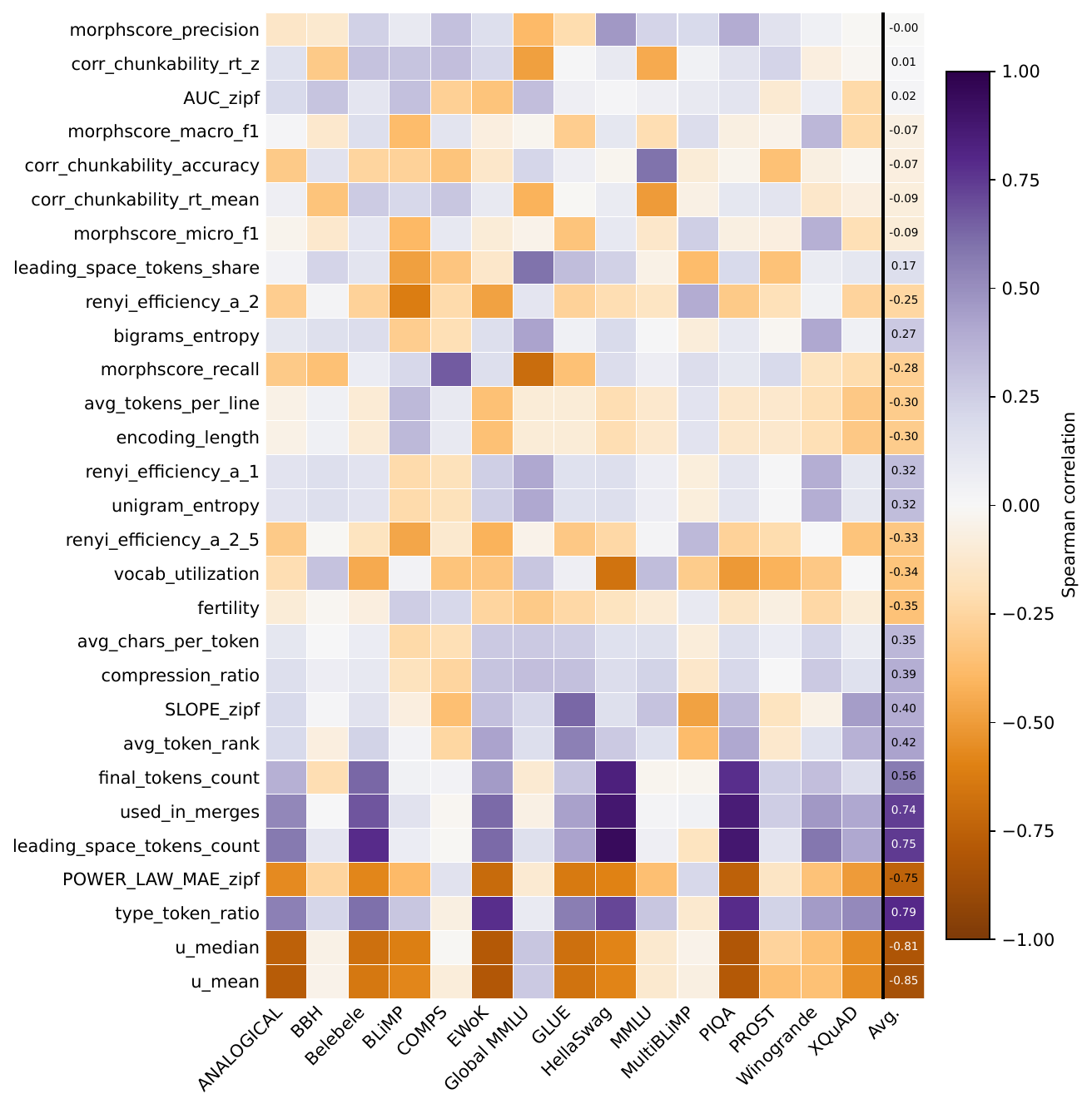}
        \caption{English}
    \end{subfigure}
    \hspace{2em}
    \begin{subfigure}{0.45\textwidth}
        \centering
        \includegraphics[height=5.5cm, keepaspectratio]{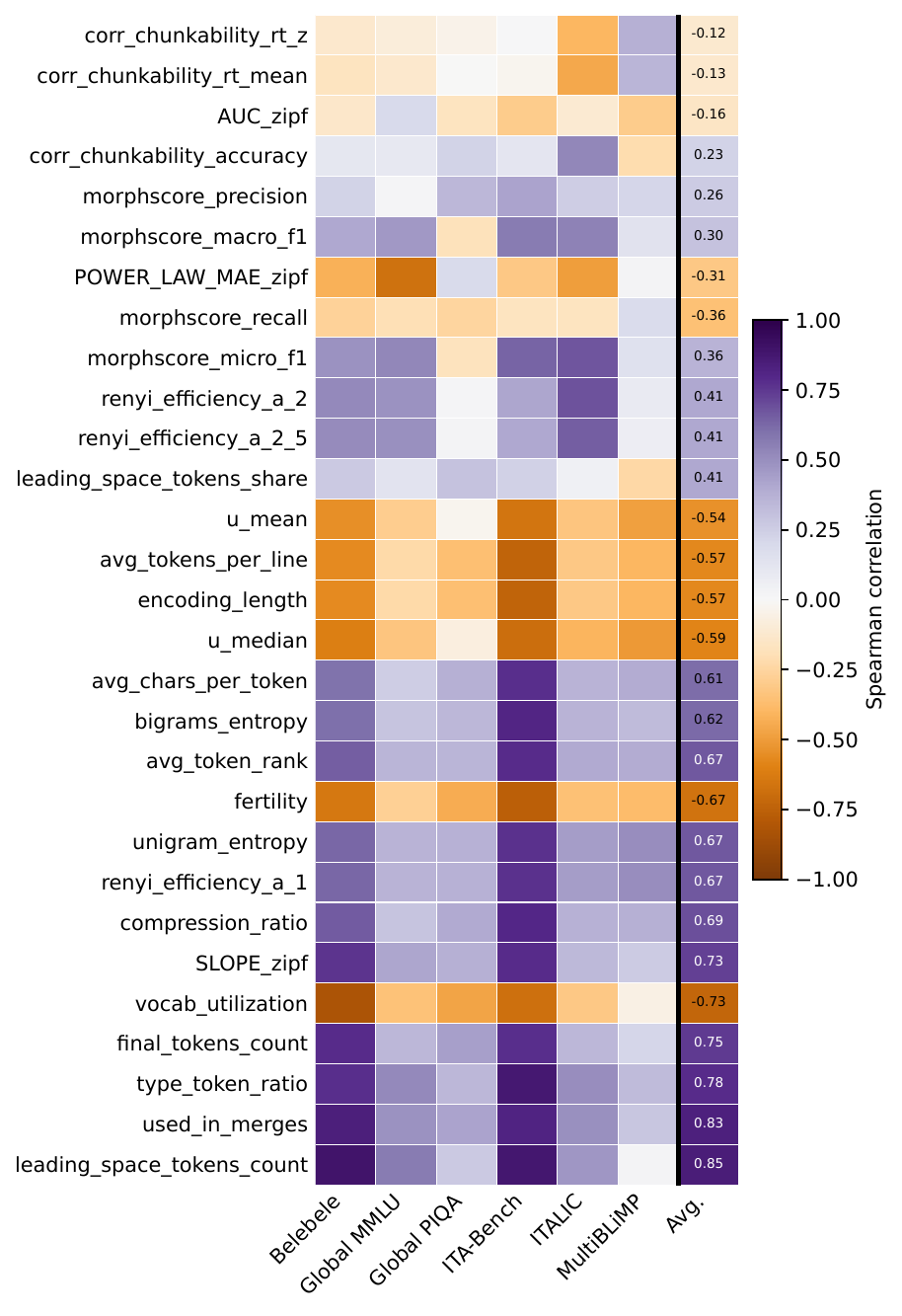}
        \caption{Italian}
    \end{subfigure}

    \vspace{0.8em}

    \begin{subfigure}{0.45\textwidth}
        \centering
        \includegraphics[height=5.5cm, keepaspectratio]{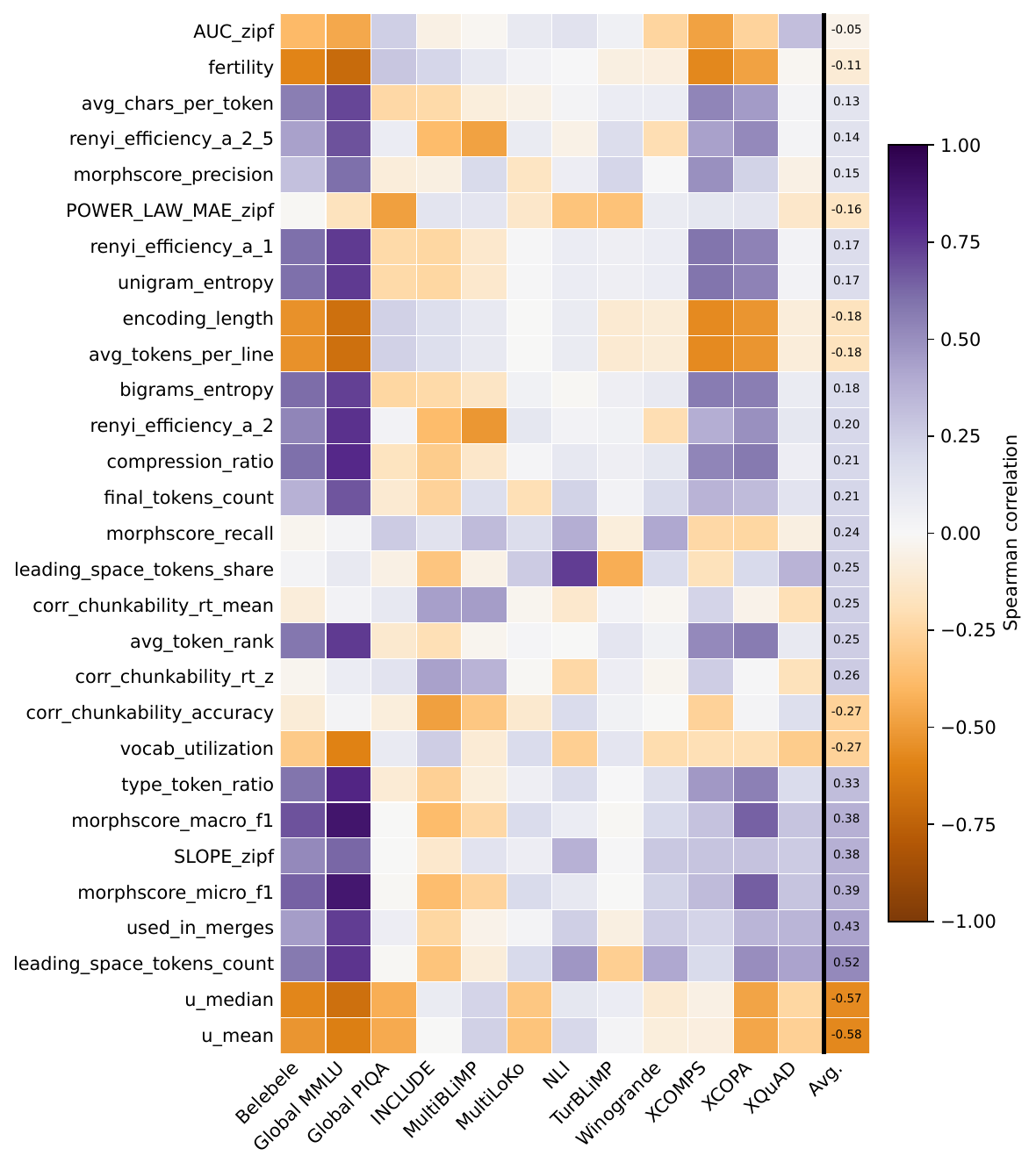}
        \caption{Turkish}
    \end{subfigure}
    \hspace{2em}
    \begin{subfigure}{0.45\textwidth}
        \centering
        \includegraphics[height=5.5cm, keepaspectratio]{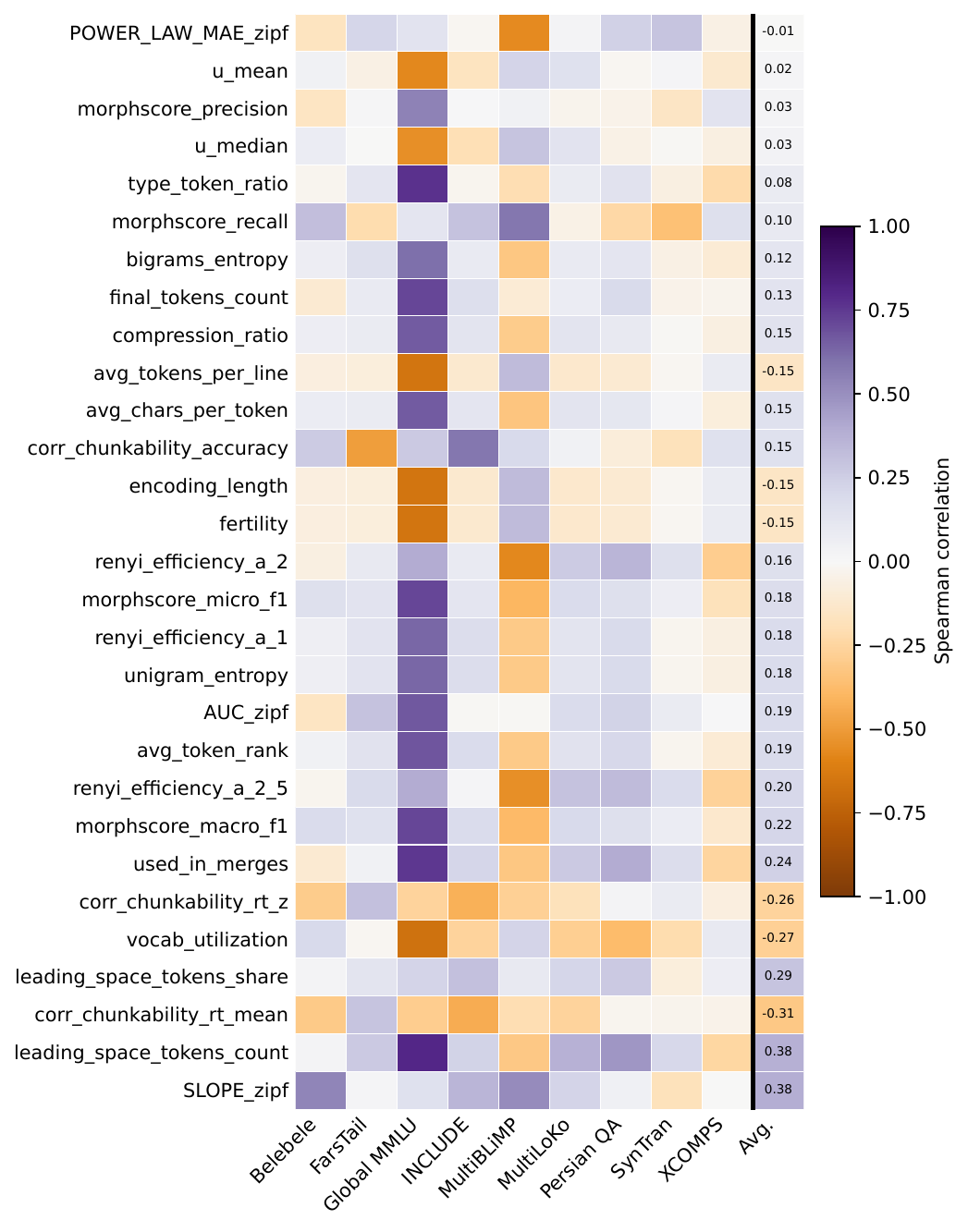}
        \caption{Farsi}
    \end{subfigure}

    \vspace{0.8em}

    \begin{subfigure}{0.6\textwidth}
        \centering
        \includegraphics[height=5.5cm, keepaspectratio]{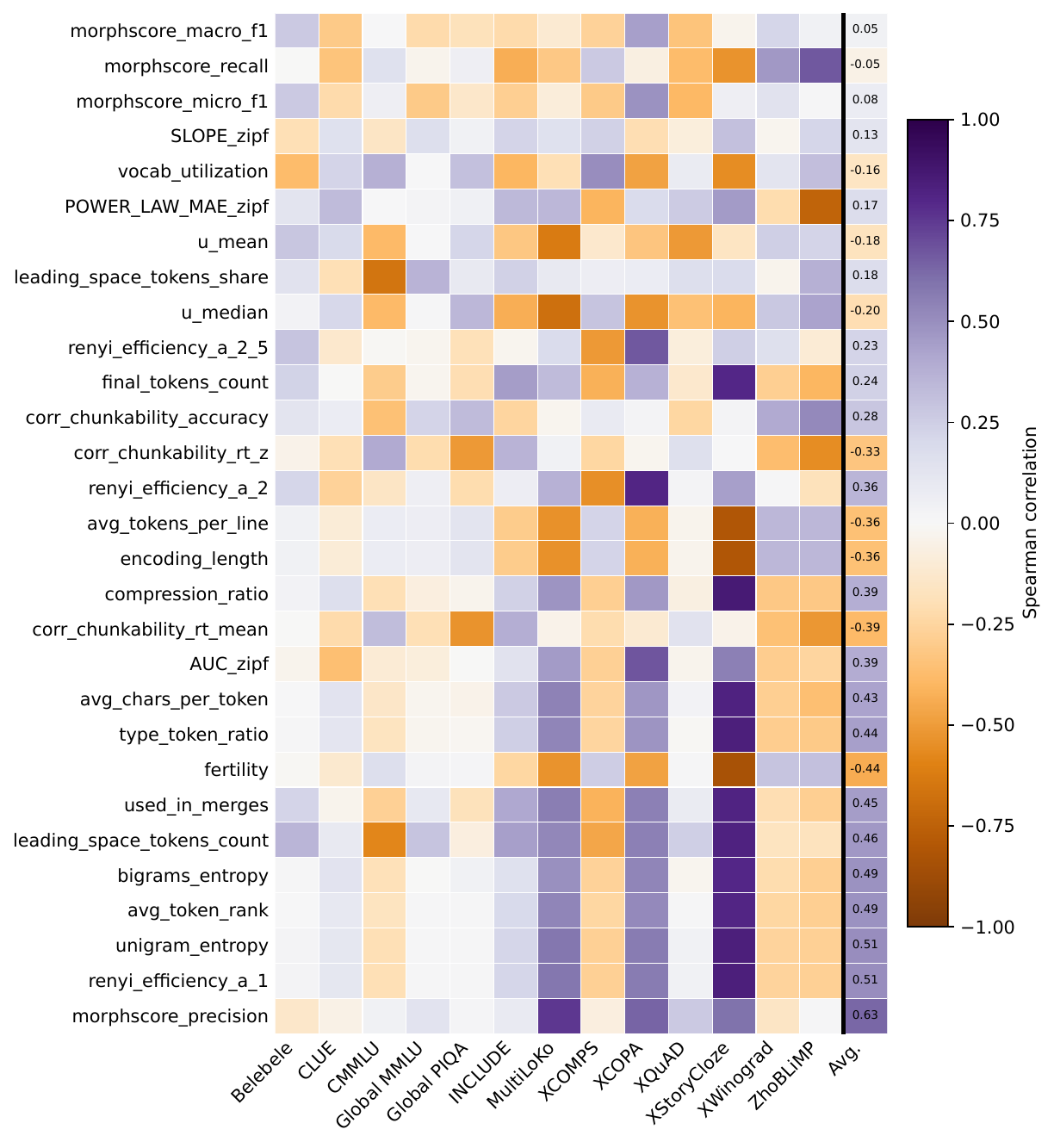}
        \caption{Chinese}
    \end{subfigure}

    \caption{Correlations between intrinsic metric and model performance for TokSuite models for each of the five languages covered. Each column represents a benchmark (the mean over its tasks), with the final column representing aggregated performance. The color of each cell corresponds to direction of correlation and the shade corresponds to the strength.}
    \label{fig:toksuite_correlations_full}
\end{figure*}

\begin{figure}[t]
    \centering
    \includegraphics[width=0.32\textwidth]{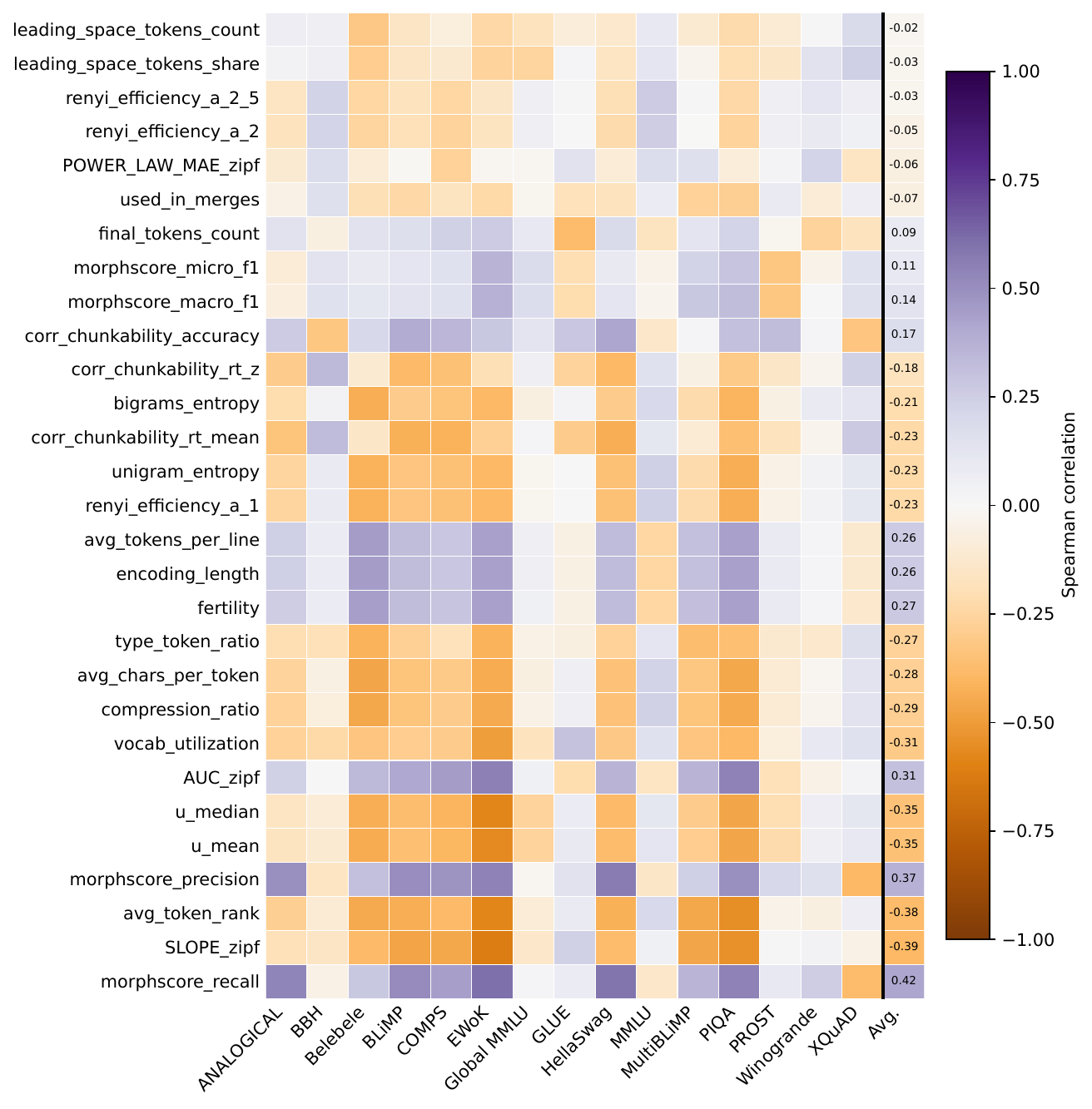}
    \caption{Correlations between intrinsic metric and model performance for the TIMTC models for English. Each column represents a benchmark (the mean over its tasks), with the final column representing aggregated performance. The color of each cell corresponds to direction of correlation and the shade corresponds to the strength.}
    \label{fig:timtc_correlations_full}
\end{figure}

\begin{figure*}[t]
    \centering

    \begin{subfigure}{0.48\textwidth}
        \centering
        \includegraphics[height=6.5cm, keepaspectratio]{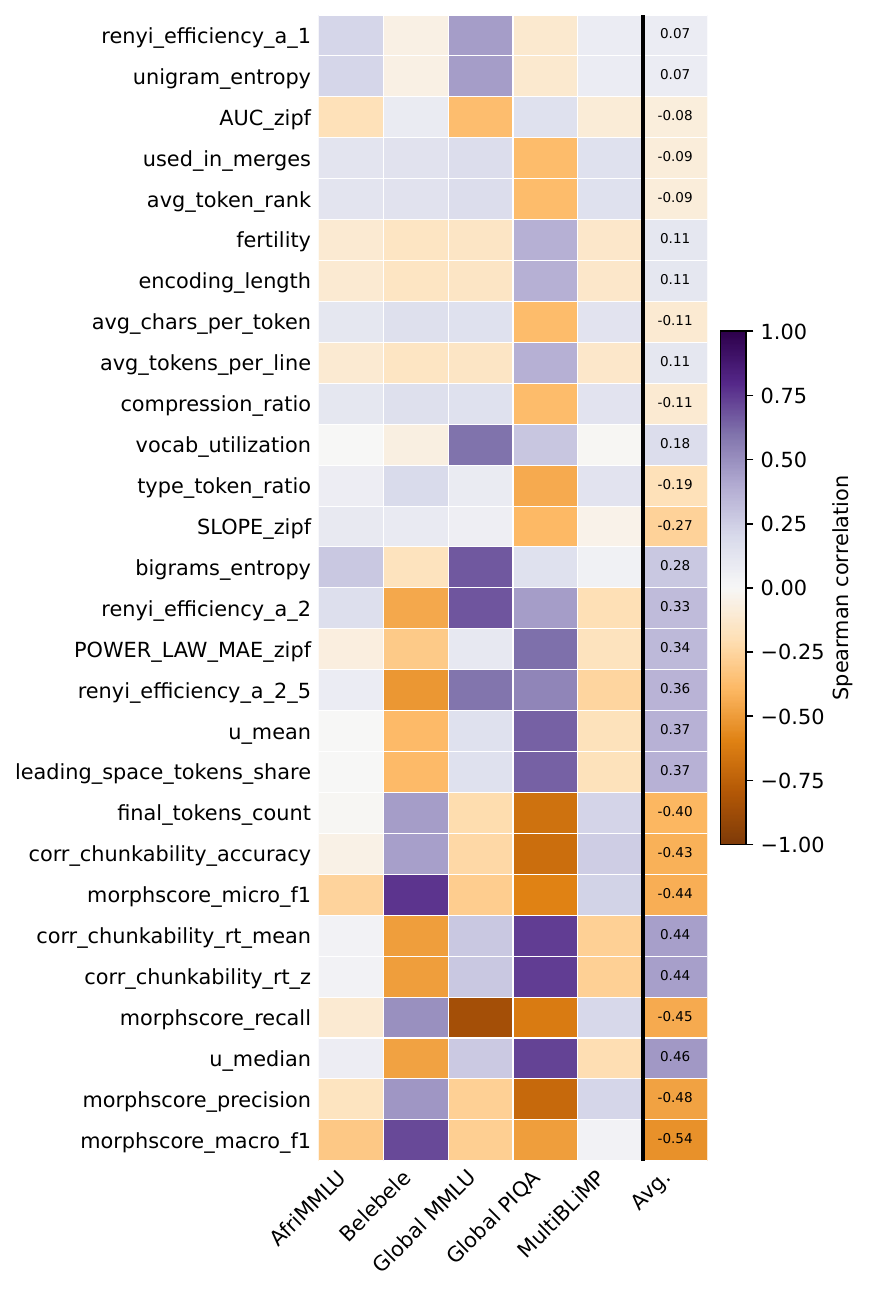}
        \caption{Amharic}
    \end{subfigure}
    \hfill
    \begin{subfigure}{0.48\textwidth}
        \centering
        \includegraphics[height=6.5cm, keepaspectratio]{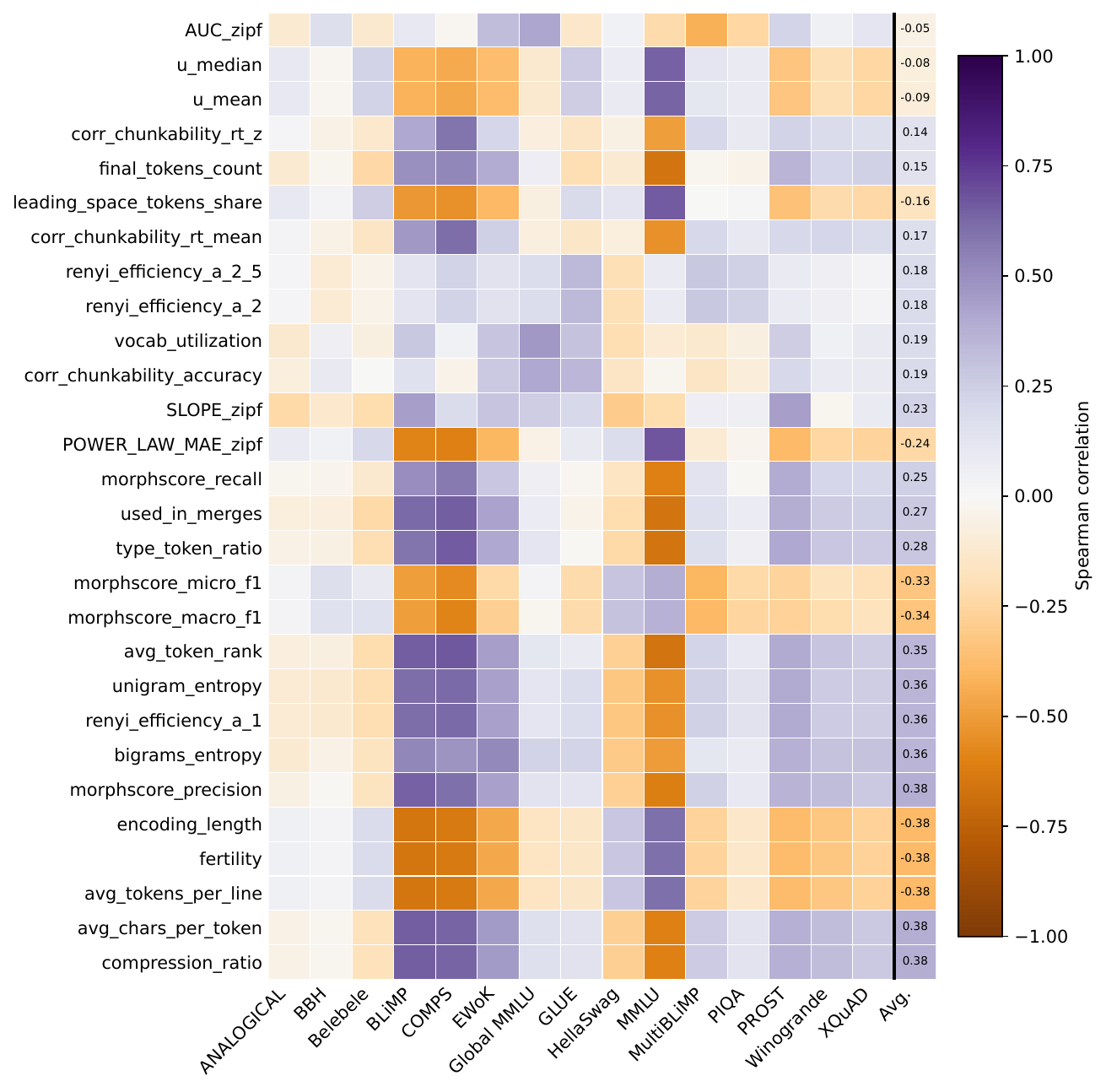}
        \caption{English}
    \end{subfigure}

    \vspace{0.5em}

    \begin{subfigure}{0.48\textwidth}
        \centering
        \includegraphics[height=6.5cm, keepaspectratio]{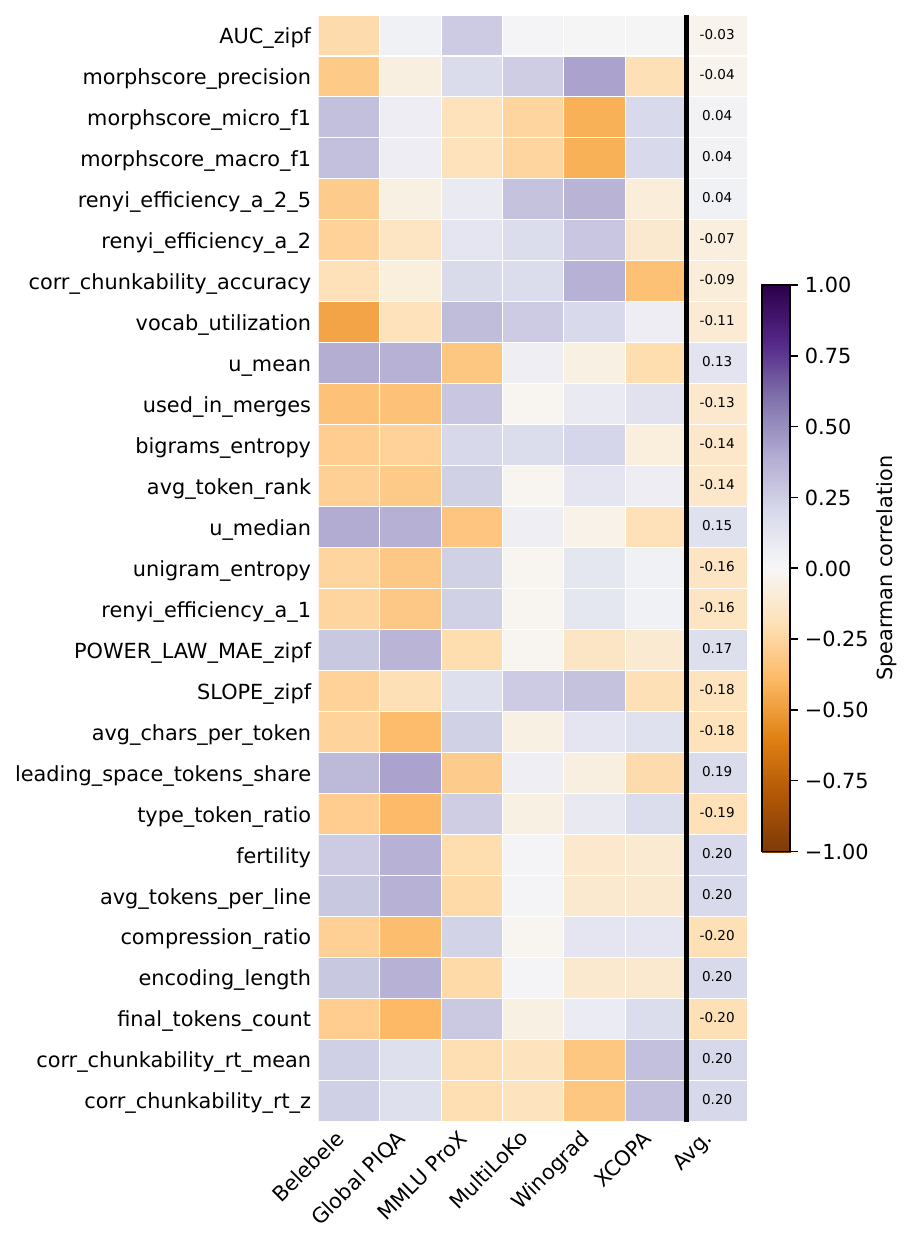}
        \caption{Thai}
    \end{subfigure}
    \hfill
    \begin{subfigure}{0.48\textwidth}
        \centering
        \includegraphics[height=6.5cm, keepaspectratio]{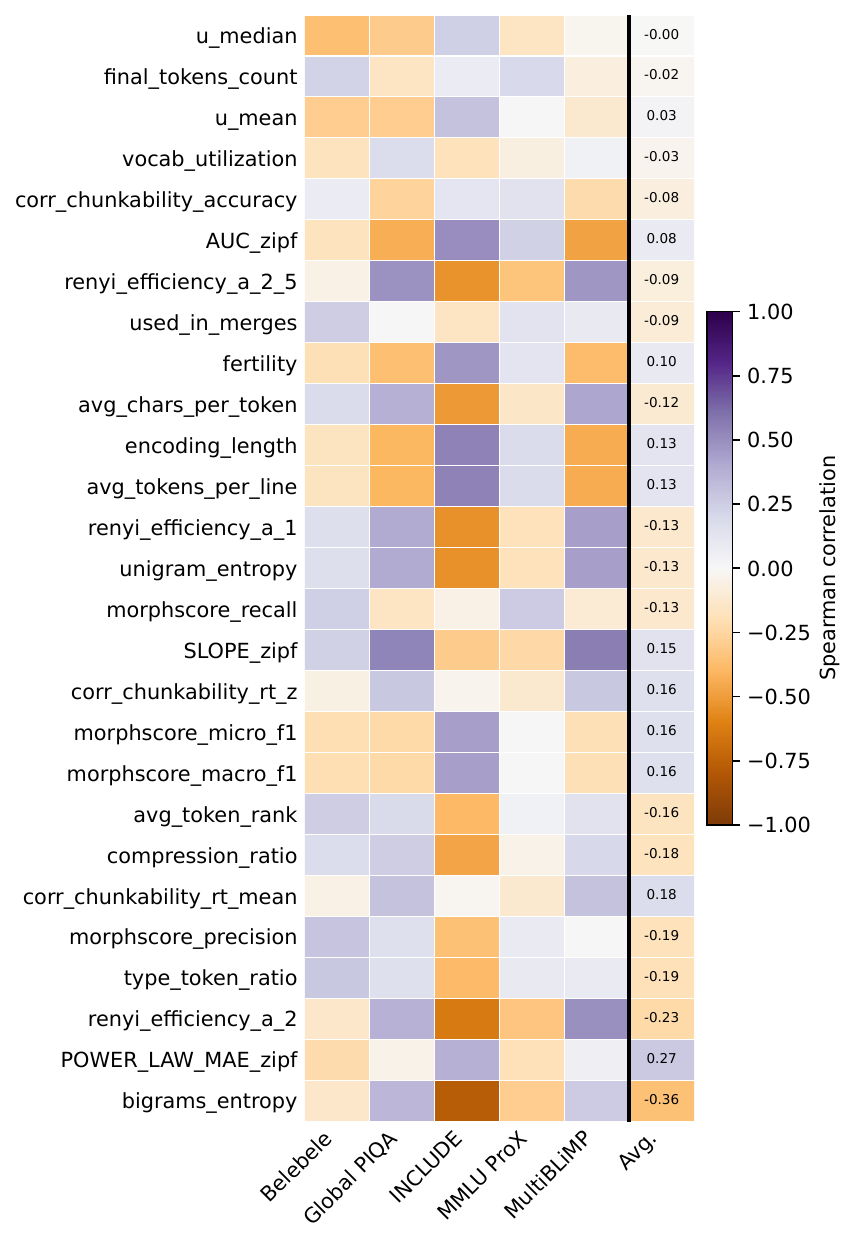}
        \caption{Urdu}
    \end{subfigure}

    \vspace{0.5em}

    \begin{subfigure}{0.48\textwidth}
        \centering
        \includegraphics[height=6.5cm, keepaspectratio]{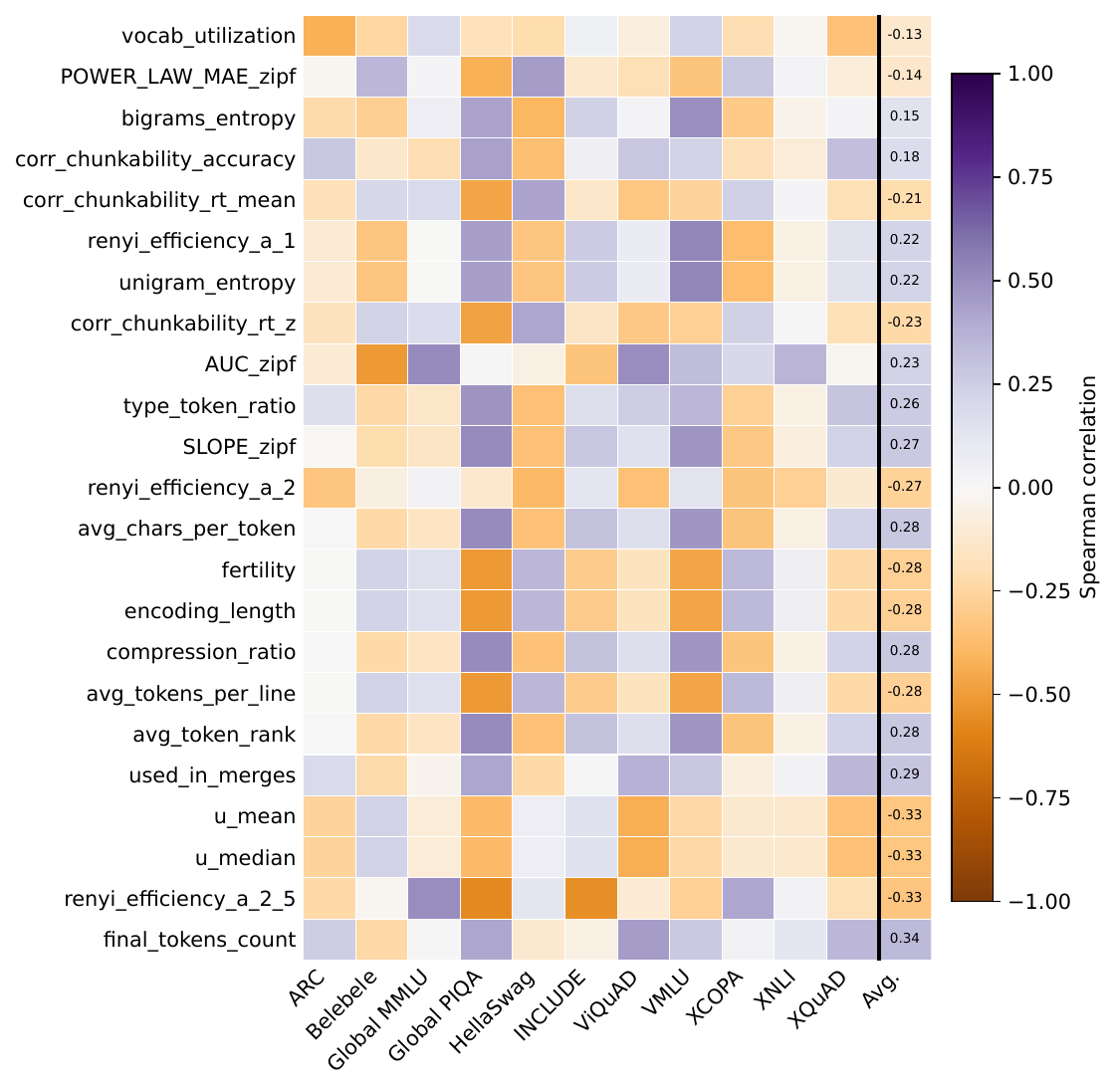}
        \caption{Vietnamese}
    \end{subfigure}

    \caption{Correlations between intrinsic metric and model performance for VISTA models for each of the five languages. Each column represents a benchmark (the mean over its tasks), with the final column representing aggregated performance. The color of each cell corresponds to direction of correlation and the shade corresponds to the strength.}
    \label{fig:vista_correlations_full}
\end{figure*}

Here we show the correlation results for all of the model suites and all of the languages. Figure \ref{fig:toksuite_correlations_full} shows the correlations for the TokSuite models. Figure \ref{fig:timtc_correlations_full} shows the correlations for the TIMTC models. Figure \ref{fig:vista_correlations_full} shows the correlations for the VISTA models.

In Table \ref{tab:correlations_comparison_exclusions}, we report the by-metric correlation reports over all tasks (left) and tasks for which the 130M parameter models scored above chance (right; see Section \ref{app:exclusion_analysis}). 

\begin{table*}[ht]
\tiny
\centering
\caption{Correlation between tokenizer metrics and model performance, with and without task exclusions (for tasks where performance was near chance). Significance markers (*) indicate $p_\text{corrected} < 0.05$.} 
\label{tab:correlations_comparison_exclusions}
\resizebox{\textwidth}{!}{%
\begin{tabular}{l rrrr rrrr}
\toprule
 & \multicolumn{4}{c}{\textbf{All Tasks}} & \multicolumn{4}{c}{\textbf{With Exclusions}} \\
\cmidrule(lr){2-5} \cmidrule(lr){6-9}
\textbf{Metric} & $r$ & $p$ & $p_\text{corr}$ & sig & $r$ & $p$ & $p_\text{corr}$ & sig \\
\midrule
AUC\_zipf                    & $-0.04$ & 0.00 & 0.01 & * & $-0.00$ & 0.78 & 1.00 &   \\
avg\_chars\_per\_token       & $-0.02$ & 0.06 & 1.00 &   & $-0.01$ & 0.60 & 1.00 &   \\
avg\_token\_rank             &  0.08 & 0.00 & 0.00 & * &  0.06 & 0.00 & 0.02 & * \\
avg\_tokens\_per\_line       &  0.05 & 0.00 & 0.00 & * &  0.01 & 0.46 & 1.00 &   \\
bigrams\_entropy             & $-0.03$ & 0.03 & 0.74 &   &  0.00 & 0.91 & 1.00 &   \\
compression\_ratio           & $-0.01$ & 0.27 & 1.00 &   & $-0.01$ & 0.60 & 1.00 &   \\
corr\_chunkability\_accuracy & $-0.16$ & 0.00 & 0.00 & * & $-0.11$ & 0.00 & 0.00 & * \\
corr\_chunkability\_rt\_mean &  0.04 & 0.00 & 0.00 & * &  0.01 & 0.61 & 1.00 &   \\
corr\_chunkability\_rt\_z    &  0.06 & 0.00 & 0.00 & * &  0.02 & 0.25 & 1.00 &   \\
encoding\_length             &  0.05 & 0.00 & 0.00 & * &  0.01 & 0.46 & 1.00 &   \\
fertility                    &  0.07 & 0.00 & 0.00 & * &  0.02 & 0.28 & 1.00 &   \\
final\_tokens\_count         &  0.15 & 0.00 & 0.00 & * &  0.12 & 0.00 & 0.00 & * \\
leading\_space\_tokens\_count &  0.27 & 0.00 & 0.00 & * &  0.23 & 0.00 & 0.00 & * \\
leading\_space\_tokens\_share &  0.25 & 0.00 & 0.00 & * &  0.23 & 0.00 & 0.00 & * \\
morphscore\_macro\_f1        & $-0.05$ & 0.00 & 0.00 & * & $-0.04$ & 0.01 & 0.30 &   \\
morphscore\_micro\_f1        & $-0.05$ & 0.00 & 0.00 & * & $-0.05$ & 0.01 & 0.19 &   \\
morphscore\_precision        & $-0.16$ & 0.00 & 0.00 & * & $-0.10$ & 0.00 & 0.00 & * \\
morphscore\_recall           & $-0.14$ & 0.00 & 0.00 & * & $-0.10$ & 0.00 & 0.00 & * \\
POWER\_LAW\_MAE\_zipf        & $-0.02$ & 0.18 & 1.00 &   &  0.00 & 1.00 & 1.00 &   \\
renyi\_efficiency\_a\_1      &  0.02 & 0.17 & 1.00 &   &  0.03 & 0.09 & 1.00 &   \\
renyi\_efficiency\_a\_2      &  0.00 & 0.78 & 1.00 &   &  0.02 & 0.30 & 1.00 &   \\
renyi\_efficiency\_a\_2\_5   & $-0.00$ & 0.84 & 1.00 &   &  0.01 & 0.42 & 1.00 &   \\
SLOPE\_zipf                  & $-0.06$ & 0.00 & 0.00 & * & $-0.02$ & 0.24 & 1.00 &   \\
type\_token\_ratio           &  0.04 & 0.00 & 0.04 & * &  0.02 & 0.31 & 1.00 &   \\
u\_mean                      & $-0.00$ & 0.88 & 1.00 &   & $-0.00$ & 0.93 & 1.00 &   \\
u\_median                    &  0.00 & 0.83 & 1.00 &   &  0.00 & 1.00 & 1.00 &   \\
unigram\_entropy             &  0.02 & 0.17 & 1.00 &   &  0.03 & 0.09 & 1.00 &   \\
used\_in\_merges             &  0.22 & 0.00 & 0.00 & * &  0.18 & 0.00 & 0.00 & * \\
vocab\_utilization           & $-0.12$ & 0.00 & 0.00 & * & $-0.09$ & 0.00 & 0.00 & * \\
\bottomrule
\end{tabular}%
}
\end{table*}

\clearpage

\clearpage

\clearpage
\subsection{Correlation with Orthographic Awareness Tasks} \label{app:orthographic_corr}

\begin{figure}[ht!]
    \centering
    \includegraphics[width=0.95\linewidth]{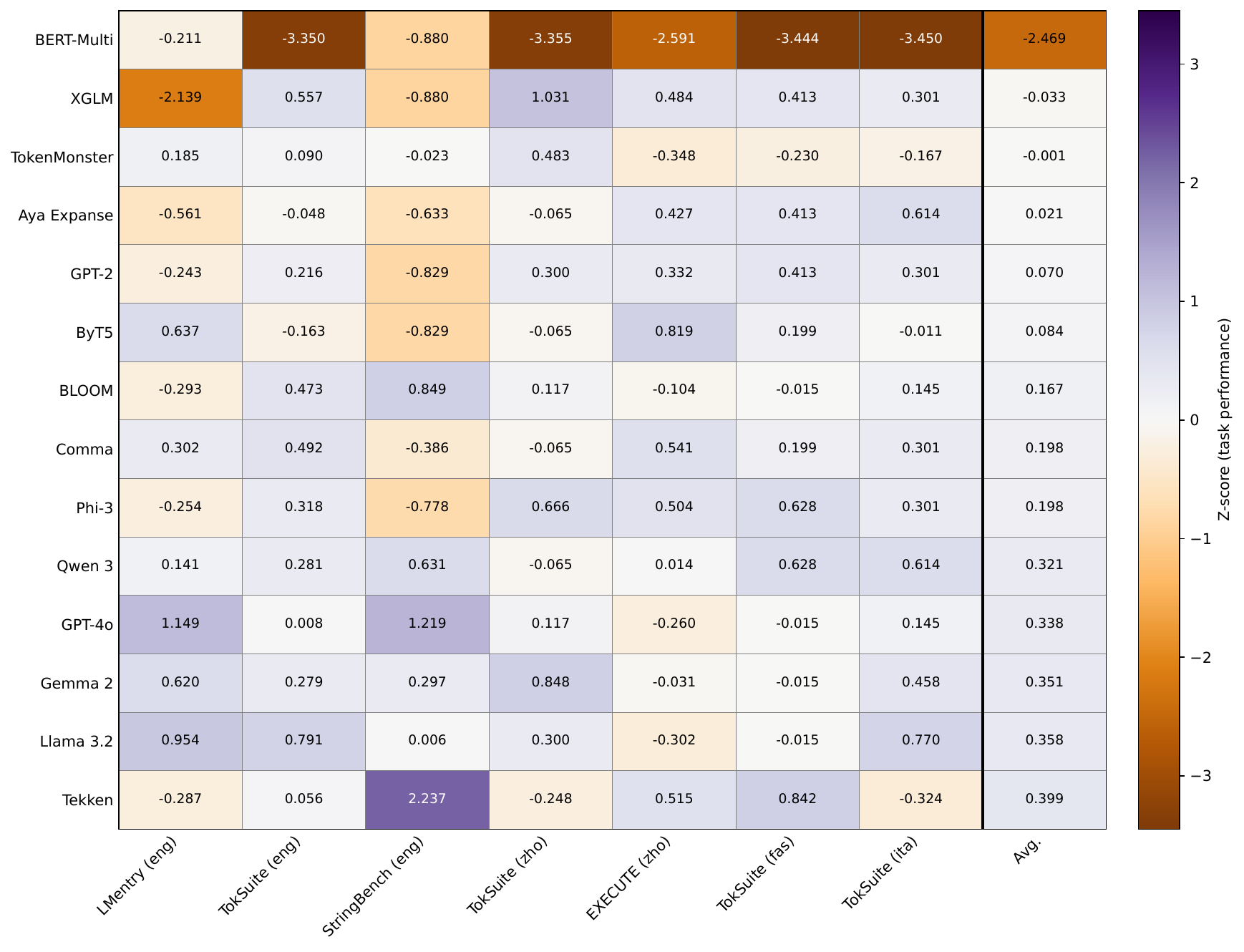}
    \caption{Performance for each of the TokSuite models on the orthographic task groups of Table~\ref{tab:orth_corr} (per model, the mean z-scored performance on the group's tasks, after the exclusion filter), with their mean (Avg.). Color corresponds to the z-scored performance, where darker purple indicates better performance.}
    \label{fig:toksuite_token_task_perf}
\end{figure}

Results in Figure~\ref{fig:toksuite_token_task_perf} show performance on the orthographic tasks.  
The ByT5 tokenizer model performs best on a subset of the orthographic tasks (namely, CUTE and EXECUTE), though, in aggregate, it does not outperform other models. As all the orthographic tasks are similar and are composed of heterogeneous subtasks, it is not clear why the byte-level model outperforms the other models on some tasks but not others. 

Table \ref{tab:orth_corr} presents the correlations between performance on each group of orthographic tasks and aggregate downstream performance in the corresponding language.

\begin{table}[ht!]
\centering
\begin{tabular}{@{}llll@{}}
\toprule
\textbf{Task Group} & \textbf{Language} & \textbf{$\rho$} & \textbf{p-value} \\ \midrule
LMentry & English & 0.07 & 0.728 \\
TokSuite & English & 0.19 & 0.219 \\
StringBench & English & 0.23 & 0.310 \\
TokSuite & Chinese & 0.17 & 0.553 \\
EXECUTE & Chinese & 0.04 & 0.810 \\
TokSuite & Farsi & 0.86 & $<0.001^*$ \\
TokSuite & Italian & 0.38 & 0.185 \\
\bottomrule
\end{tabular}
\caption{Spearman correlation between orthographic task performance and aggregate downstream performance in the same language, over the TokSuite model and task pairs of each group whose task passes the exclusion filter of Appendix~\ref{app:exclusion_analysis} and whose score is above chance. p-values are uncorrected; $^*$ marks significance after Bonferroni correction over the seven groups ($p<0.05/7$).}
\label{tab:orth_corr}
\end{table}

\subsection{Multiple Predictor Correlations} \label{app:multiple_predictors}

\begin{figure}[ht!]
    \centering
    \includegraphics[width=0.95\linewidth]{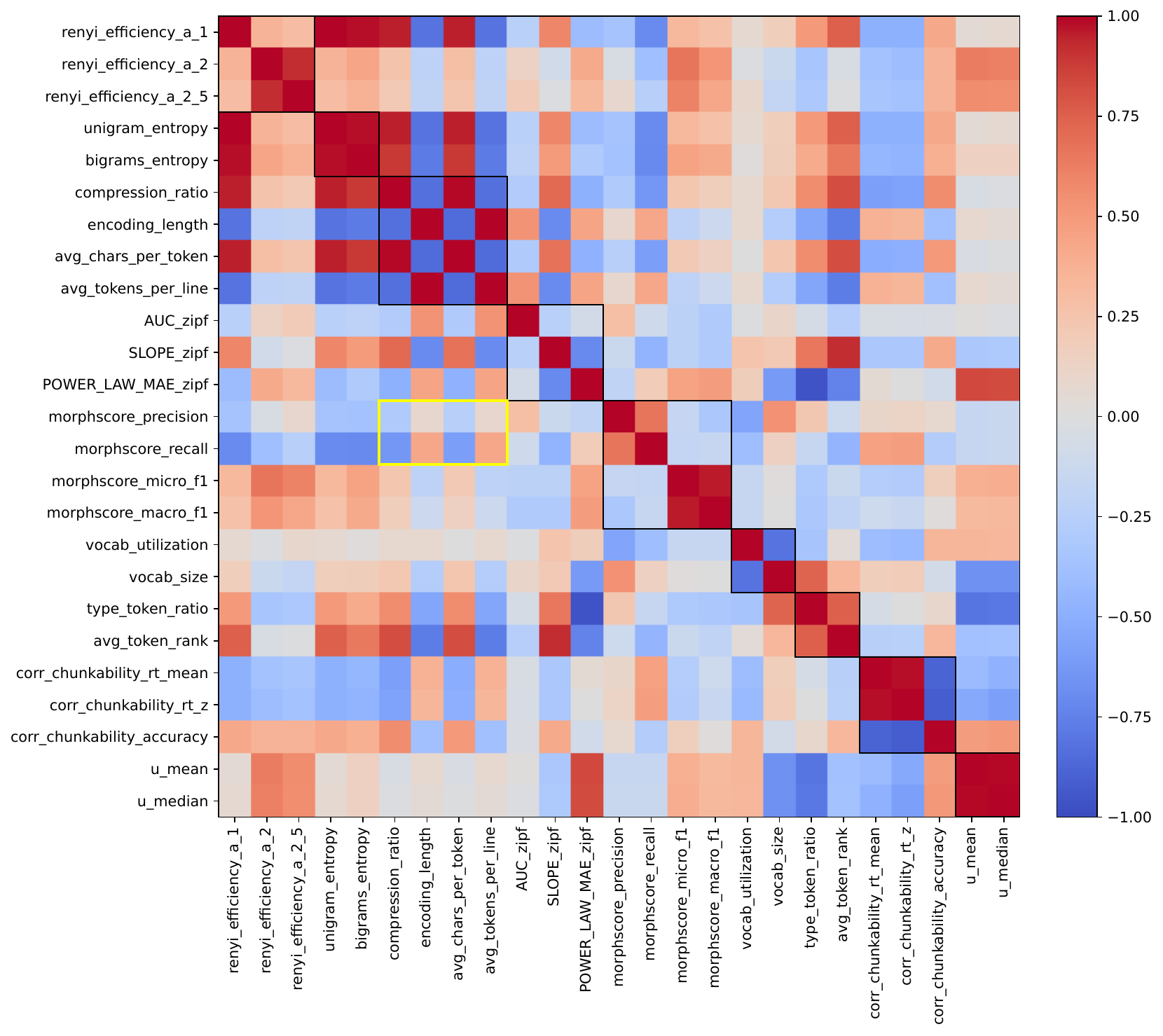}
    \caption{The boxes highlight comparisons between metrics that index similar things. Yellow highlights correlation between MorphScore and compression metrics. } 
    \label{fig:toksuite_english_metrics_correlation}
\end{figure}

Figure \ref{fig:toksuite_english_metrics_correlation} shows the correlations between each of the intrinsic tokenizer metrics. Within each type of tokenizer evaluation (indicated with black boxes), many metrics are strongly correlated with each other. Outside of those types, there are several other strong correlations. For example both unigram and bigram entropy are strongly correlated with compression metrics, and MorphScore is somewhat correlated with worse compression, as predicted. Many potential relationships emerge in this analysis. So, in the next section, we test whether combining any of these metrics can better predict downstream performance.

We test whether a combination of two intrinsic tokenizer metrics better predicts model performance.
Table \ref{tab:two_metrics} shows the variance explained (adjusted $R^2$) by linear regressions with individual metrics (\texttt{adj\_r2\_m1} and \texttt{adj\_r2\_m2}) as predictors and both metrics as predictors (\texttt{adj\_r2\_joint}) for the top ten combinations. \texttt{r2\_gain} shows the increased explained variance with the two predictors together.  The p-value was corrected for multiple comparisons.

\begin{table*}[ht]
\centering
\caption{Variance explained by combinations of two metrics, shown for the ten best combinations.}
\label{tab:two_metrics}
\resizebox{\textwidth}{!}{%
\begin{tabular}{rllrrrrr}
\toprule
 & \textbf{Metric 1} & \textbf{Metric 2} & \textbf{adj\_r2\_m1} & \textbf{adj\_r2\_m2} & \textbf{adj\_r2\_joint} & \textbf{r2\_gain} & \textbf{p-value} \\
\midrule
1  & leading\_space\_tokens\_count & morphscore\_precision & 0.073 & 0.026 & 0.090 & 0.017 & $<0.001$ \\
2  & leading\_space\_tokens\_count & u\_mean & 0.084 & 0.000 & 0.086 & 0.002 & 0.010 \\
3  & leading\_space\_tokens\_count & u\_median & 0.084 & 0.000 & 0.085 & 0.001 & 1.000 \\
4  & final\_tokens\_count & leading\_space\_tokens\_count & 0.021 & 0.073 & 0.085 & 0.012 & $<0.001$ \\
5  & leading\_space\_tokens\_count & leading\_space\_tokens\_share & 0.073 & 0.060 & 0.084 & 0.012 & $<0.001$ \\
6  & leading\_space\_tokens\_count & AUC\_zipf & 0.073 & 0.002 & 0.084 & 0.011 & $<0.001$ \\
7  & leading\_space\_tokens\_share & u\_median & 0.075 & 0.000 & 0.083 & 0.009 & $<0.001$ \\
8  & leading\_space\_tokens\_count & morphscore\_recall & 0.073 & 0.018 & 0.083 & 0.010 & $<0.001$ \\
9  & leading\_space\_tokens\_count & corr\_chunkability\_accuracy & 0.073 & 0.024 & 0.083 & 0.010 & $<0.001$ \\
10 & final\_tokens\_count & used\_in\_merges & 0.021 & 0.047 & 0.083 & 0.036 & $<0.001$ \\
\bottomrule
\end{tabular}%
}
\end{table*}

\subsection{Model Size} \label{app:model_size}

Table~\ref{tab:model_size} gives, for each metric, the variance in task performance it explains beyond model size, the analysis summarised in Section~\ref{sec:individual_metric_corr}.

\begin{table}[ht]
\centering
\small
\setlength{\tabcolsep}{4pt}
\caption{Variance explained by each metric beyond model size, as the gain in $R^2$ when the metric is added to a linear regression of z-scored task performance on log total parameters (model size alone explains $R^2=0.109$). p-values are from a likelihood-ratio test against model size alone, Bonferroni-corrected over the 29 metrics.}
\label{tab:model_size}
\begin{tabular}{lrr}
\toprule
\textbf{Metric} & \textbf{$R^2$ gain} & \textbf{p-value} \\
\midrule
vocab\_utilization & 0.015 & $<0.001$ \\
morphscore\_recall & 0.010 & $<0.001$ \\
morphscore\_precision & 0.009 & $<0.001$ \\
AUC\_zipf & 0.009 & $<0.001$ \\
avg\_token\_rank & 0.009 & $<0.001$ \\
leading\_space\_tokens\_share & 0.007 & $<0.001$ \\
avg\_chars\_per\_token & 0.006 & $<0.001$ \\
compression\_ratio & 0.006 & $<0.001$ \\
final\_tokens\_count & 0.006 & $<0.001$ \\
SLOPE\_zipf & 0.004 & $<0.001$ \\
corr\_chunkability\_accuracy & 0.004 & $<0.001$ \\
u\_mean & 0.004 & $<0.001$ \\
avg\_tokens\_per\_line & 0.004 & $<0.001$ \\
encoding\_length & 0.004 & $<0.001$ \\
type\_token\_ratio & 0.002 & $<0.001$ \\
u\_median & 0.002 & 0.002 \\
leading\_space\_tokens\_count & 0.002 & 0.005 \\
unigram\_entropy & 0.002 & 0.012 \\
renyi\_efficiency\_a\_1 & 0.002 & 0.012 \\
corr\_chunkability\_rt\_mean & 0.001 & 0.663 \\
fertility & 0.001 & 0.778 \\
used\_in\_merges & 0.000 & 1.000 \\
morphscore\_micro\_f1 & 0.000 & 1.000 \\
morphscore\_macro\_f1 & 0.000 & 1.000 \\
corr\_chunkability\_rt\_z & 0.000 & 1.000 \\
bigrams\_entropy & 0.000 & 1.000 \\
POWER\_LAW\_MAE\_zipf & 0.000 & 1.000 \\
renyi\_efficiency\_a\_2 & 0.000 & 1.000 \\
renyi\_efficiency\_a\_2\_5 & 0.000 & 1.000 \\
\bottomrule
\end{tabular}
\end{table}

\section{Exclusion Analysis} \label{app:exclusion_analysis}

In this section, we compare the results with and without the exclusion of tasks where most models did not perform above chance. We excluded tasks for which there was not more than one model from each of the model families that achieved above-chance performance. 
Table \ref{tab:correlations_comparison_exclusions} compares the by-metric correlations with task performance with and without the exclusions. 
Table \ref{tab:two_metrics_exclusions} shows the results for the correlations with two predictors (cf. Table \ref{tab:two_metrics}).

\begin{table*}[ht]
\centering
\caption{Variance explained by combinations of two metrics, shown for the ten best combinations. These results exclude any of the tasks where the small models performed near chance.}
\label{tab:two_metrics_exclusions}
\resizebox{\textwidth}{!}{%
\begin{tabular}{rllrrrrr}
\toprule
 & \textbf{Metric 1} & \textbf{Metric 2} & \textbf{adj\_r2\_m1} & \textbf{adj\_r2\_m2} & \textbf{adj\_r2\_joint} & \textbf{r2\_gain} & \textbf{p-value} \\
\midrule
1  & final\_tokens\_count & used\_in\_merges & 0.013 & 0.033 & 0.071 & 0.038 & $<0.001$ \\
2  & leading\_space\_tokens\_share & u\_median & 0.063 & 0.000 & 0.071 & 0.007 & $<0.001$ \\
3  & leading\_space\_tokens\_share & u\_mean & 0.063 & 0.000 & 0.070 & 0.007 & $<0.001$ \\
4  & leading\_space\_tokens\_share & AUC\_zipf & 0.054 & 0.000 & 0.069 & 0.015 & $<0.001$ \\
5  & final\_tokens\_count & leading\_space\_tokens\_count & 0.013 & 0.052 & 0.068 & 0.016 & $<0.001$ \\
6  & leading\_space\_tokens\_count & leading\_space\_tokens\_share & 0.052 & 0.054 & 0.066 & 0.012 & $<0.001$ \\
7  & leading\_space\_tokens\_share & corr\_chunkability\_accuracy & 0.054 & 0.011 & 0.065 & 0.011 & $<0.001$ \\
8  & used\_in\_merges & leading\_space\_tokens\_share & 0.033 & 0.054 & 0.064 & 0.010 & $<0.001$ \\
9  & leading\_space\_tokens\_count & u\_mean & 0.058 & 0.000 & 0.060 & 0.003 & 0.657 \\
10 & leading\_space\_tokens\_count & morphscore\_precision & 0.052 & 0.010 & 0.059 & 0.008 & $<0.001$ \\
\bottomrule
\end{tabular}%
}
\end{table*}

\section{AI Use} 

The authors used AI to aid in code generation for figure and table creation and to edit the text of this paper. The authors take full responsibility for the content of the paper and any associated artifacts.

\end{document}